\documentclass[11pt]{article}

\usepackage[margin=1in]{geometry}

\usepackage[utf8]{inputenc}
\usepackage[T1]{fontenc}
\usepackage{graphicx}
\usepackage{hyperref}
\usepackage{url}
\usepackage{booktabs}
\usepackage{amsfonts}
\usepackage{amsmath}
\usepackage{amssymb}
\usepackage{amsthm}
\usepackage{float}
\usepackage{mathtools}
\usepackage{nicefrac}
\usepackage{microtype}
\usepackage[capitalize,noabbrev]{cleveref}
\usepackage{enumitem}
\usepackage{bm}
\usepackage{authblk}
\usepackage{natbib}
\theoremstyle{plain}

\theoremstyle{definition}

\theoremstyle{remark}

\newcommand{\R}{\mathbb{R}}

\newcommand{\Det}{\operatorname{Det}}
\newcommand{\Tr}{\operatorname{Tr}}

\newcommand{\wbar}{\bar{w}}
\newcommand{\mbar}{\bar{m}}
\newcommand{\nubar}{\bar{\nu}}
\newcommand{\zbar}{\bar{z}}
\newcommand{\gbar}{\bar{g}}

\renewcommand{\epsilon}{\varepsilon}

\title{A Rod Flow Model for Adam at the Edge of Stability}

\author[1]{Eric Regis}
\author[1]{Sinho Chewi}
\affil[1]{Yale University}
\affil[ ]{\texttt{\{eric.regis, sinho.chewi\}@yale.edu}}

\date{}

\begin{document}

\maketitle

\begin{abstract}
\citet{cohen2022adaptive} observed that adaptive gradient methods such as Adam operate at the edge of stability. While there has been significant work on continuous-time modeling of gradient descent at the edge of stability, extending these models to momentum methods remains underdeveloped. In the gradient descent setting, \citet{regis2026rod} introduced \emph{rod flow}, which models consecutive iterates as an extended one-dimensional object---a ``rod.'' Here we extend rod flow to Adam by working in the joint phase space of parameters and first moment $(w, m)$ and treating the second moment $\nu$ as a smooth auxiliary variable. We also develop rod flows for heavy ball momentum, Nesterov momentum, and scalar and per-component versions of RMSProp, Adam, and NAdam. For all eight optimizers, we empirically evaluate rod flow on representative machine learning architectures, where it tracks the discrete iterates through the edge-of-stability regime significantly more accurately than the corresponding stable flow.
\end{abstract}

%%%%%%%%%%%%%%%%%%%%%%%%%%%%%%%%%%%%%%%%%%%%%%%%%%%%%%%%%%%%%%%%%%%%%%%%%%%%%%%
% INTRODUCTION
%%%%%%%%%%%%%%%%%%%%%%%%%%%%%%%%%%%%%%%%%%%%%%%%%%%%%%%%%%%%%%%%%%%%%%%%%%%%%%%

\section{Introduction}
\label{sec:introduction}

Neural networks are trained by minimizing loss functions with gradient-based
optimizers. \citet{cohen2021gradient} observed that full-batch gradient descent
operates at the \emph{edge of stability} (EoS): the largest eigenvalue of the Hessian, called the \emph{sharpness}, first rises (a phase called
\emph{progressive sharpening}) and then hovers at the stability threshold
$2/\eta$ where $\eta$ is the learning rate. \citet{cohen2022adaptive} extended
this picture to momentum methods and adaptive gradient methods, showing that
each optimizer exhibits its own edge of stability. Rather than hovering at
$2/\eta$, the relevant quantity---the \emph{preconditioned sharpness}---hovers
at a hyperparameter-dependent threshold that depends on the optimizer (\cref{tab:optimizers}).

In practice, the dominant optimizer in machine learning is Adam~\citep{kingma2015adam}, which
differs from gradient descent in two respects. First, it is a momentum method:
rather than updating parameters with the gradient at the current point, it
maintains an exponential moving average $m$ of past gradients and updates based on
this momentum. Second, it is an adaptive method: when updating the parameters, Adam divides each component of
the momentum by $\sqrt{\nu}$, where $\nu$ is an exponential moving average of
the squared gradients. The effect is to normalize each parameter's update by
the typical magnitude of its gradient.

Continuous-time models have long served as a valuable theoretical tool for analyzing discrete-time optimizers~\citep{li2017stochastic, barrett2021implicit, shi2021continuous}, clarifying their implicit biases and asymptotic behavior. Such a model would be especially valuable for Adam, the workhorse of modern deep learning.

However, developing such a model that remains valid at the edge of stability is non-trivial. At EoS, the discrete iterates oscillate along the sharpest directions of the Hessian. But the na\"{\i}ve continuous-time limit---which is called the ``stable flow''---admits no oscillations.

While there is a rich and
growing body of continuous-time theory for gradient descent at the edge of
stability~\citep{arora2022understanding, DamNicLee23SelfStab,
Cohen+25CentralFlow, regis2026rod}, the corresponding theory for Adam and its
close relatives---heavy ball momentum, Nesterov momentum,
NAdam---remains underdeveloped. Existing continuous-time analyses of Adam either take a vanishing-step-size
limit that discards oscillatory dynamics by
construction~\citep{belotto2020general, barakat2021convergence} or study
finite-step-size effects that stop short of the oscillatory EoS
regime~\citep{malladi2022sde, CatKluShi24ImplicitAdam}. The regime that
practitioners actually utilize---finite-step-size Adam operating at the edge of stability---currently lacks a principled continuous-time model.

\subsection{Our Approach}

Recently, \citet{regis2026rod} introduced \emph{rod flow}: a
continuous-time model of gradient descent at the edge of stability. Rod flow tracks a smoothly-evolving
center~$\wbar$, the average of consecutive iterates, and an extent
tensor~$\Sigma$, the outer product of the half-displacement between consecutive iterates. The physical
picture is intuitive: rather than following a point that oscillates back and
forth along the sharp direction of the Hessian, one follows an extended
one-dimensional object---a ``rod''---whose length and orientation encode the
oscillation.

By extending the rod flow framework to momentum and adaptive gradient methods, we develop a continuous-time model for Adam at the edge of stability.

Extending rod flow to Adam requires two modifications. First, the rod must be
lifted from weight space~$w \in \mathbb{R}^d$ to phase
space~$z=(w,m)\in\R^{2d}$. For momentum methods operating at the edge of
stability, the first moment~$m$ oscillates in lockstep with~$w$, so the rod's center and
half-difference must be defined over both coordinates. This extension
handles both heavy-ball and Nesterov momentum.

Second, the preconditioner must be tracked as a smooth auxiliary variable. The
second moment~$\nu$ in RMSProp and Adam is an exponential moving average of the
\emph{squared} gradient; the squaring kills the sign flips, so $\nu$ varies
smoothly even when~$w$ oscillates at the edge of stability. It therefore
needs no rod structure of its own.

Adam is thus modeled as
phase-space rod flow in the adaptive metric defined by~$\nu$.

\subsection{Contributions}
\begin{itemize}[leftmargin=*,itemsep=0pt,topsep=2pt]
    \item A derivation of rod flows for heavy-ball and Nesterov momentum, extending the original weight-space model to phase-space~(\cref{sec:momentum}).
    \item A smooth-auxiliary treatment of the preconditioner, yielding rod
    flows for scalar and per-component RMSProp (\cref{sec:preconditioned}).
    \item A rod flow for Adam and NAdam that combines the phase-space
    and preconditioner extensions~(\cref{sec:adam}).
    \item An empirical evaluation for Adam showing that the rod flow tracks
    the discrete iterates accurately and exhibits self-stabilization of
    the preconditioned sharpness at the theoretically predicted
    thresholds~(\cref{sec:experiments}).
\end{itemize}

%%%%%%%%%%%%%%%%%%%%%%%%%%%%%%%%%%%%%%%%%%%%%%%%%%%%%%%%%%%%%%%%%%%%%%%%%%%%%%%
% RELATED WORK
%%%%%%%%%%%%%%%%%%%%%%%%%%%%%%%%%%%%%%%%%%%%%%%%%%%%%%%%%%%%%%%%%%%%%%%%%%%%%%%
\section{Related Work}
\label{sec:related}

\textbf{Edge of stability.} For full-batch gradient descent, \citet{cohen2021gradient} documented progressive sharpening and edge of stability across architectures, with sharpness rising to $2/\eta$ and then hovering. Theoretical explanations include self-stabilization \citep{DamNicLee23SelfStab}, two-step analysis \citep{ahn2023learning, chen2023beyond}, convergence in the unstable regime \citep{ahn2022understanding, arora2022understanding, zhu2023understanding}, progressive sharpening in quadratic regression \citep{AgaPedPen23SecondOrder}, the catapult mechanism \citep{lewkowycz2020large, ghosh2023implicit} and its momentum-induced amplification \citep{Phu+24MomentumCatapults}, loss landscape analysis \citep{MaKunYin22BeyondQuadratic}, bifurcation theory \citep{song2023trajectory}, chaotic GD dynamics \citep{KonTao20StochasticityGD, Che+24StabToChaos}, regularity-induced implicit biases \citep{Wang+25EoS}, and NTK evolution at EoS \citep{JiaCohLi25NTKEoS}. \citet{cohen2022adaptive} empirically identify edge of stability for adaptive gradient methods.

\textbf{Adam.} The original Adam method is due to \citet{kingma2015adam}; variants include NAdam \citep{dozat2016nadam}, AMSGrad \citep{reddi2018convergence}, and AdamW \citep{loshchilov2019decoupled}. We analyze Adam and NAdam in their standard EMA form. Continuous-time analyses of Adam have all operated in small-step or near-minimum regimes rather than through EoS: foundational ODE limits \citep{barakat2021convergence, dereich2025ode}, general adaptive-ODE frameworks \citep{belotto2020general}, SDE approximations with scaling rules \citep{malladi2022sde}, IMEX time-stepping \citep{bhattacharjee2024imex} and integro-differential \citep{heredia2024modeling} views, control-theoretic framings \citep{chakrabarti2024control}, stability in the hyperparameter-stable regime \citep{gould2024continuous}, backward error analysis at small LR \citep{CatKluShi24ImplicitAdam}, slow-SDE sharpness-reduction near minimizers \citep{li2025adam}, Adam-flow implicit bias on homogeneous networks \citep{wang2021implicit}, and mechanistic accounts of EoS-induced loss spikes \citep{bai2025adaptive}. None of these captures the finite-step-size Adam during EoS.

\textbf{Continuous-time models.} Gradient flow is the classical continuous
limit of gradient descent but fails at the edge of stability since it cannot
capture oscillations. \citet{rosca2023continuous} study instabilities in
gradient flow models. \citet{shi2021continuous} derive high-resolution ODEs for
momentum methods, while \citet{li2017stochastic} develop stochastic modified
equations for SGD. Modified equations from backward error analysis are standard
in numerical analysis~\citep{wilkinson1963rounding, wilkinson1965algebraic,
hairer2006geometric}; \citet{barrett2021implicit} and \citet{smith2021origin}
use this framework to understand implicit regularization in gradient descent,
and \citet{digiovacchino2024backward} provides methodological validation. The
closest related work is Central Flow~\citep{Cohen+25CentralFlow}, which derives
continuous-time models for gradient descent, scalar RMSProp, and
per-component RMSProp at the edge of stability. The direct predecessor of this
work is \citet{regis2026rod}, which introduces a rod flow for gradient descent at
EoS. We extend the rod flow framework to phase space and to adaptive methods.

\section{Rod Flow for Gradient Descent}
\label{sec:rod_flow_GD}

We briefly recap the rod flow for gradient descent, introduced in \citet{regis2026rod}. A full rederivation appears in \cref{app:gradient_descent}.

Consider gradient descent with learning rate $\eta$:
\begin{equation}
  w_{t+1} = w_t - \eta\,\nabla L(w_t).
\end{equation}
Define the center and half-difference of consecutive iterates:
\begin{align}
  \wbar_t &= \tfrac{1}{2}(w_{t+1} + w_t), \\
  \delta_t &= \tfrac{1}{2}(w_{t+1} - w_t).
\end{align}
 The rod flow framework exploits the fact that, even when individual iterates $w_t$ oscillate rapidly at the edge of stability, the center $\wbar_t$ and extent tensor $\delta_t\otimes\delta_t$ vary smoothly---making them amenable to continuous-time ODE approximation.
 
 For brevity, we write $L_\pm \coloneqq L(\wbar\pm\delta)$ with $\nabla L_\pm$ and $\nabla^2 L_\pm$ denoting the corresponding gradients and Hessians. The discrete difference equations for $\bar{w}$ and $\delta_t \otimes \delta_t$ are given as:
\begin{align}
  \bar{w}_{t+1} - \bar{w}_t &= -\frac{\eta}{2}\bigl(\nabla L_+ + \nabla L_-\bigr), \label{eq:diff_gd_center}\\
  \delta_{t+1} \otimes \delta_{t+1} - \delta_t \otimes \delta_t &= \frac{\eta^2}{4}\bigl(\nabla L_+\otimes\nabla L_+ + \nabla L_-\otimes\nabla L_-\bigr) - 2\delta_t \otimes \delta_t. \label{eq:diff_gd_sigma}
\end{align}
Let $\Sigma(t)$ denote the continuous-time analogue of the extent tensor $\delta_t\otimes\delta_t$. In continuous time, $\delta$ is to be identified with the principal eigenvector of $\Sigma$ scaled by the square root of the principal eigenvalue. Promoting \cref{eq:diff_gd_center} and \cref{eq:diff_gd_sigma} to ODEs yields the rod flow ODEs for gradient descent:
\begin{align}
  \frac{d\wbar}{dt} &= -\eta\,\bar{g}, \label{eq:rod_gd_center}\\
  \frac{d\Sigma}{dt} &= \frac{\eta^2}{4}\bigl(\nabla L_+\otimes\nabla L_+ + \nabla L_-\otimes\nabla L_-\bigr) - 2\Sigma. \label{eq:rod_gd_sigma}
\end{align}
where $\gbar = (\nabla L_+ + \nabla L_-)/2$ is the average of the gradients at the endpoints of the rod.

Note that the discrete difference equations for $\wbar_t$ and $\delta_t \otimes \delta_t$ are \textit{exact}. The only approximation in the rod flow ODEs is replacing the discrete differences with continuous-time derivatives. Traditionally, such interpolation would require backward error analysis. The original derivation in \citet{regis2026rod} includes an $O(\eta^2)$ backward-error-analysis correction in \cref{eq:rod_gd_center}, which we drop throughout for simplicity as it does not qualitatively affect the dynamics. See \cref{app:bea} for further discussion of backward error analysis.

%%%%%%%%%%%%%%%%%%%%%%%%%%%%%%%%%%%%%%%%%%%%%%%%%%%%%%%%%%%%%%%%%%%%%%%%%%%%%%%
% MOMENTUM ROD FLOW
%%%%%%%%%%%%%%%%%%%%%%%%%%%%%%%%%%%%%%%%%%%%%%%%%%%%%%%%%%%%%%%%%%%%%%%%%%%%%%%

\section{Rod Flow for Momentum Methods}
\label{sec:momentum}

\begin{figure}
  \centering
  \includegraphics[width=0.37\linewidth]{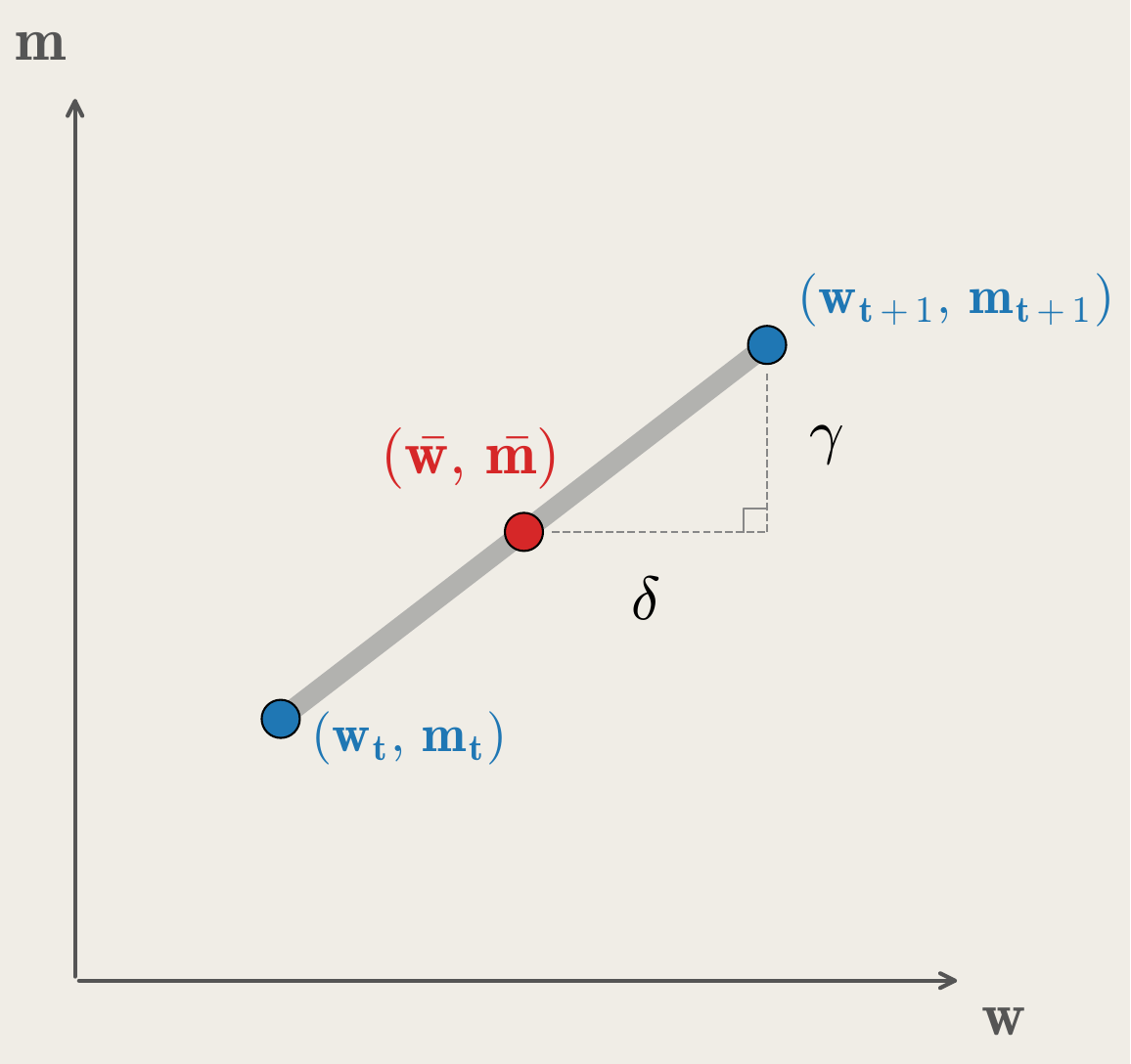}%
  \hfill
  \includegraphics[width=0.60\linewidth]{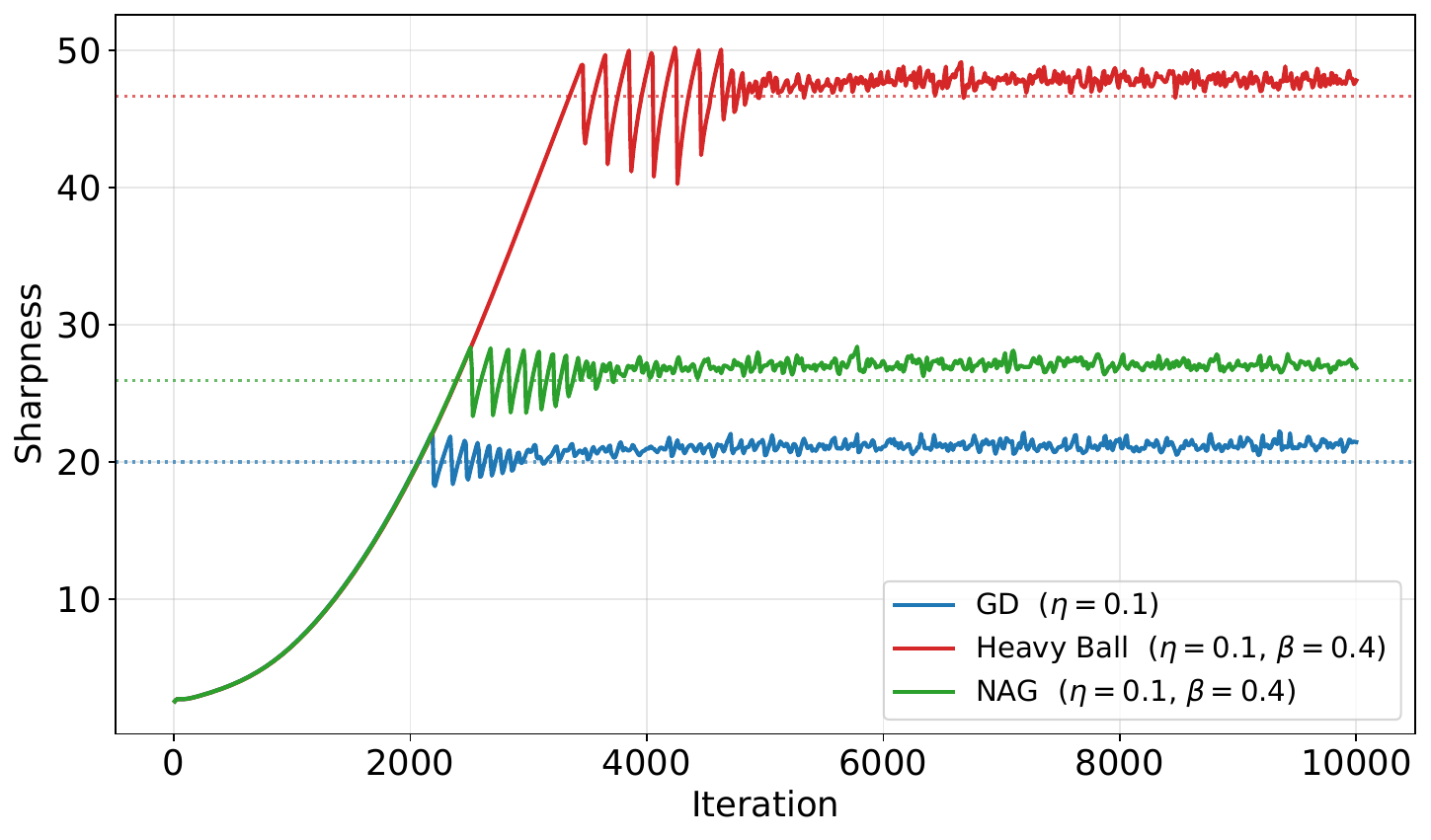}
\caption{\textbf{Phase-Space Rod Flow. Left:} Depiction of a phase-space rod. The endpoints represent consecutive phase-space iterates. \textbf{Right:} Sharpness trajectories from training a 3-layer MLP on CIFAR, showing the differing sharpness thresholds for gradient descent, heavy ball, and Nesterov momentum with the same step size.}
  \label{fig:rod_dynamics}
\end{figure}

We will now extend rod flow to heavy ball and Nesterov momentum. To do so, we will lift the rod from weight space to phase space.

\subsection{Setup}

We will work with heavy ball momentum, using the exponential moving average (EMA) parameterization throughout. The heavy ball momentum update equations are given as:
\begin{align}
m_{t+1} &= \beta\, m_t + (1-\beta)\,\nabla L(w_t), \label{eq:hb_update_momentum_1}\\
w_{t+1} &= w_t - \eta\, m_{t+1}, \label{eq:hb_update_position}
\end{align}
where $\beta \in [0,1)$ is the momentum coefficient. This differs from the form in \citet{polyak1964some}, but we adopt this convention as it is the form in which Adam is typically written.

Just like gradient descent, heavy ball momentum on a quadratic has a sharpness threshold beyond which the iterates will no longer converge. However, it is no longer the $2/\eta$ threshold: it is modified by the momentum. For heavy ball momentum with step size $\eta$ and momentum coefficient $\beta$, the sharpness threshold $S^*$ is:
\begin{equation}
S^* = \frac{2}{\eta} \cdot \frac{1+\beta}{1-\beta}
\end{equation}
 
Formally, one can write heavy ball momentum as a two-step recurrence relation. For the quadratic loss, this recurrence relation is linear, and $S^*$ corresponds to the threshold at which the eigenvalues of the transition operator leave the unit circle. More details are provided in \cref{app:momentum}.

When heavy ball crosses into EoS, $w$ and $m$ oscillate in phase: both flip sign about the center each iteration. Because the position and momentum oscillate together, it is natural to concatenate them into a single phase-space vector $z$:
\begin{equation}
z_t = \begin{pmatrix} w_t \\ m_t \end{pmatrix} \in \R^{2d}.
\end{equation}
In analogy with the rod flow for gradient descent, we can define the average of two consecutive phase-space iterates $\bar{z}$ and the half-displacement between consecutive phase-space iterates $\Delta$:
\begin{align}
\zbar_t &= \tfrac{1}{2}(z_{t+1} + z_t) = \begin{pmatrix} \wbar_t \\ \mbar_t \end{pmatrix}, \\
\Delta_t &= \tfrac{1}{2}(z_{t+1} - z_t) = \begin{pmatrix} \delta_t \\ \gamma_t \end{pmatrix},
\end{align}
where $\bar{m}_t = (m_{t+1} + m_t)/2$ and $\gamma_t = (m_{t+1} - m_t)/2$. We emphasize that $(\delta, \gamma)$ forms a \emph{single} rod in phase space, not two separate rods in position and momentum space: the formalism is symmetric under the simultaneous flip $(\delta, \gamma) \mapsto (-\delta, -\gamma)$ but not under flipping just one of the coordinates.

\subsection{Exact Difference Equations}

Write the phase-space update as: 
\begin{equation}
z_{t+1} = z_t + \Phi(z_t) 
\end{equation}
where we have that:
\begin{equation}
\Phi(z) = \begin{pmatrix} -\eta\bigl[\beta m + (1-\beta)\nabla L(w)\bigr] \\ (1-\beta)\bigl[\nabla L(w) - m\bigr] \end{pmatrix}. \label{eq:Phi_hb}
\end{equation}

Following the analogous derivation for gradient descent, the exact difference equations for the phase-space center and extent are:
\begin{align}
\zbar_{t+1} - \zbar_t &= \tfrac{1}{2}\bigl[\Phi_+ + \Phi_-\bigr], \label{eq:zbar_diff}\\
\Delta_{t+1} \otimes \Delta_{t+1} - \Delta_t \otimes \Delta_t &= \tfrac{1}{4}\bigl[\Phi_+ \otimes \Phi_+ + \Phi_- \otimes \Phi_-\bigr] - 2\,\Delta_t \otimes \Delta_t, \label{eq:ext_diff}
\end{align}
where $\Phi_\pm \coloneqq \Phi(\zbar_t \pm \Delta_t)$. As in the GD case, both discrete difference equations are exact.

\subsection{Phase-Space Rod Flow}

Promoting the discrete difference equations to ODEs, we obtain the rod flow ODEs for heavy ball momentum:
\begin{align}
\frac{d\wbar}{dt} &= -\eta\bigl[\beta \mbar + (1-\beta)\gbar\bigr], \label{eq:ode_wbar_hb}\\
\frac{d\mbar}{dt} &= (1-\beta)\bigl[\gbar - \mbar\bigr], \label{eq:ode_mbar_hb}\\
\frac{d\Sigma}{dt} &= \tfrac{1}{4}\bigl[\Phi_+ \otimes \Phi_+ + \Phi_- \otimes \Phi_-\bigr] - 2\,\Sigma, \label{eq:ode_sigma_hb}
\end{align}
 Note the extent tensor $\Sigma \in \R^{2d \times 2d}$ decomposes into four $d \times d$ blocks:
\begin{equation}
\Sigma = \begin{pmatrix} \Sigma_{\delta\delta} & \Sigma_{\delta\gamma} \\ \Sigma_{\gamma\delta} & \Sigma_{\gamma\gamma} \end{pmatrix}.
\end{equation}
We again identify $\Delta = (\delta, \gamma)$ with the principal eigenvector of $\Sigma$ scaled by the square root of the principal eigenvalue.

\subsection{Nesterov Momentum}
\label{sec:nesterov}

Nesterov momentum is very similar to heavy ball momentum. The only difference is that the gradient in the momentum update (\cref{eq:hb_update_momentum_1}) is replaced by $\nabla L(w_t -\eta \beta m_t)$: a look-ahead along the current momentum. For the Nesterov phase-space update, we have that:
\begin{equation}
\Phi(z) = \begin{pmatrix} -\eta\bigl[\beta m + (1-\beta)\nabla L(w - \eta \beta m)\bigr] \\ (1-\beta)\bigl[\nabla L(w - \eta \beta m) - m\bigr] \end{pmatrix}. \label{eq:Phi_nag}
\end{equation}
The corresponding change in the rod flow is that $\nabla L_\pm$ are now evaluated at the look-ahead points:
\begin{equation}
\nabla L_\pm = \nabla L(\wbar -\eta \beta \mbar \pm [\delta - \eta \beta \gamma]).
\end{equation}

Otherwise, the Nesterov rod flow has the same functional form as the heavy ball rod flow.

%%%%%%%%%%%%%%%%%%%%%%%%%%%%%%%%%%%%%%%%%%%%%%%%%%%%%%%%%%%%%%%%%%%%%%%%%%%%%%%
% PRECONDITIONED ROD FLOW
%%%%%%%%%%%%%%%%%%%%%%%%%%%%%%%%%%%%%%%%%%%%%%%%%%%%%%%%%%%%%%%%%%%%%%%%%%%%%%%

\section{Rod Flow for Preconditioned Methods}
\label{sec:preconditioned}

We now turn to preconditioned methods. For simplicity, we will consider Scalar RMSProp and RMSProp. For these methods, $\nu$ acts as an estimator of the second moment of the gradient along the trajectory. The key observation is that, even at the edge of stability, $\nu$ does not oscillate because squaring kills sign flips.

\subsection{Scalar RMSProp}

Scalar RMSProp maintains a single scalar $\nu \in \R_+$ which follows the update rule:
\begin{equation}
\nu_{t+1} = \beta_2\, \nu_t + (1-\beta_2)\,\|\nabla L(w_t)\|^2
\end{equation}
where $\beta_2 \in [0,1)$, and $\nu$ is an exponential moving average of the gradient norm squared. Define the preconditioner:
\begin{equation}
P(\nu) = (\sqrt{\nu} + \varepsilon)I_d
\end{equation}
where $\varepsilon$ is a small constant for numerical stability. The position update becomes:
\begin{equation}
w_{t+1} = w_t - \eta P^{-1}_{t+1} \nabla L(w_t)
\end{equation}
where $P_{t} = P(\nu_{t})$.

Even when $w$ oscillates at EoS, $\|\nabla L\|^2$ is non-negative and so does not oscillate---$\nu$ evolves smoothly. Because $\nu$ does not oscillate, we will only obtain an ODE for its midpoint $\nubar_t = (\nu_{t+1} + \nu_t)/2$, but not for the half-displacement.

For the discrete difference equation for $\bar{\nu}$, we have:
\begin{equation}
\bar{\nu}_{t+1} - \bar{\nu}_{t} = (1-\beta_2)\left[\frac{\|\nabla L_+\|^2 + \|\nabla L_-\|^2}{2} - \nubar\right],
\end{equation}
which we can then promote to an ODE:
\begin{equation}
\frac{d\nubar}{dt} = (1-\beta_2)\left[\frac{\|\nabla L_+\|^2 + \|\nabla L_-\|^2}{2} - \nubar\right]. \label{eq:ode_nubar_scalar}
\end{equation}
We then obtain the following rod flow ODEs:
\begin{align}
\frac{d\wbar}{dt} &= -\eta P^{-1}(\bar{\nu})\bar{g}, \\
\frac{d \Sigma}{dt} &= \Bigl (\frac{\eta}{2}P^{-1}(\bar{\nu}) \Bigr )^2\,  \bigl(\nabla L_+ \otimes \nabla L_+ + \nabla L_- \otimes \nabla L_- \bigr) - 2 \Sigma
\end{align}
Note that if we define the preconditioned step size
\begin{equation}
\tilde{\eta} = \eta P^{-1}(\bar{\nu}),
\end{equation}
then the rod flow ODEs for $\bar{w}$ and $\Sigma$ are identical to those for gradient descent, except that $\tilde{\eta}$ replaces $\eta$. More details are provided in \cref{app:rmsprop}.

\subsection{RMSProp}

\begin{figure}
\centering
\includegraphics[width=\linewidth]{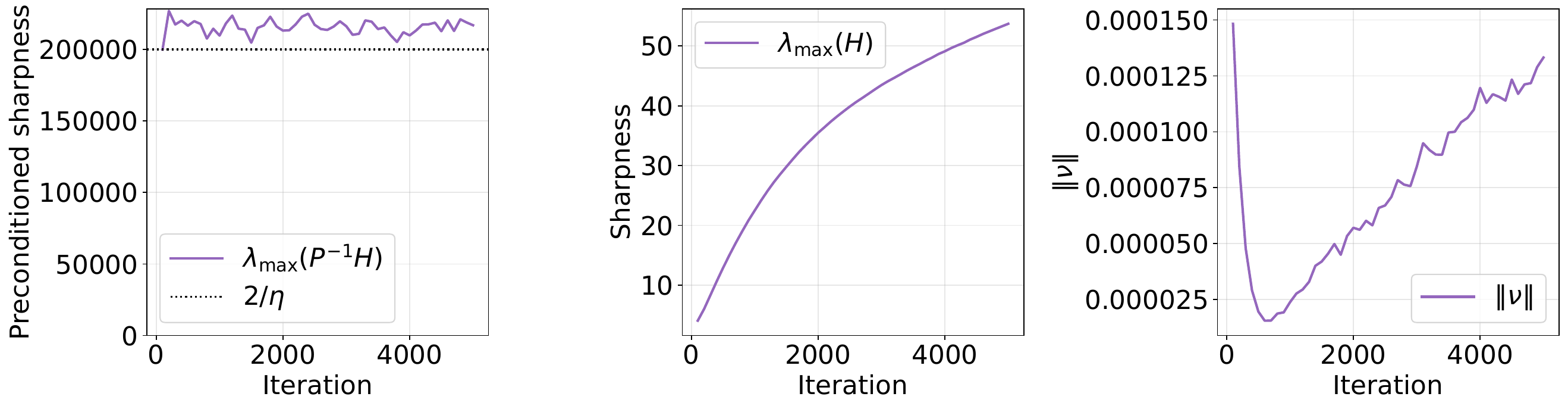}
\caption{\textbf{Edge of Stability for RMSProp.} \textbf{Left:} Preconditioned sharpness $\lambda_{\max}(P^{-1/2} H P^{-1/2})$ hovers around the stability threshold $2/\eta$. \textbf{Middle:} The raw Hessian sharpness $\lambda_{\max}(H)$ continues to rise steadily. \textbf{Right:} After an initial transient, the norm of the second moment $\|\nu\|$ rises smoothly.}
\label{fig:eos_rmsprop}
\end{figure}

RMSProp differs from Scalar RMSProp in that instead of tracking a single scalar, it tracks a vector $\nu \in \R^d$ of per-component second-moment estimates:
\begin{equation}
\nu_{t+1} = \beta_2\, \nu_t + (1-\beta_2)\, \nabla L(w_t)^{\odot 2},
\end{equation}
where $v^{\odot 2}$ denotes the elementwise square of a vector. The ODE for $\bar{\nu}$ is given as:
\begin{equation}
\frac{d\nubar}{dt} = (1-\beta_2)\,\biggl[\frac{\nabla L_+^{\odot 2} + \nabla L_-^{\odot 2}}{2} - \nubar\biggr]\,. \label{eq:ode_nubar_vec}
\end{equation}
The preconditioner is now defined as:
\begin{equation}
P(\nu) = \text{diag}(\sqrt{\nu}) + \epsilon I_d,
\end{equation}
where the square root is applied component-wise. Otherwise, the functional form for RMSProp is identical to that of Scalar RMSProp.

\subsection{Acceleration via Regularization}
\label{sec:acc_via_reg}

A major difference between preconditioned methods and gradient descent is the precise definition of the edge-of-stability condition. The relevant sharpness measure is no longer $\lambda_{\max}(H)$ but $\lambda_{\max}(P^{-1/2}HP^{-1/2})$---it is the \textit{preconditioned} sharpness that hovers at $2/\eta$.

This leads to a feedback loop that gradient descent does not exhibit. In gradient descent, the only mechanism for relieving excess sharpness (sharpness above $2/\eta$) is for the oscillation amplitude $\Sigma$ to grow until it self-stabilizes due to higher-order terms in the loss. In RMSProp, there is a second channel: as oscillations grow, so do the gradients along the sharp direction, which drives $\nu$ upward. Since the preconditioned sharpness threshold remains at $2/\eta$ while the corresponding actual sharpness threshold is $2\sqrt{\nu}/\eta$, growing $\nu$ absorbs excess sharpness without requiring the oscillation amplitude to grow as large. Central Flow~\citep{Cohen+25CentralFlow} calls this ``acceleration via regularization'': by inflating $\nu$, the preconditioner raises the sharpness it can tolerate at the cost of shrinking the effective step size $\eta/\sqrt{\nu}$, a trade-off that in practice leads to faster progress in the long run.

%%%%%%%%%%%%%%%%%%%%%%%%%%%%%%%%%%%%%%%%%%%%%%%%%%%%%%%%%%%%%%%%%%%%%%%%%%%%%%%
% ROD FLOW FOR ADAM
%%%%%%%%%%%%%%%%%%%%%%%%%%%%%%%%%%%%%%%%%%%%%%%%%%%%%%%%%%%%%%%%%%%%%%%%%%%%%%%

\section{Rod Flow for Adam}
\label{sec:adam}

We are now ready to obtain the rod flow for Adam. It is a straightforward combination of the phase-space rod of \cref{sec:momentum} and the smooth auxiliary $\nubar$ of \cref{sec:preconditioned}.

\subsection{Setup}

The update equations for Adam are:
\begin{align}
m_{t+1} &= \beta_1\, m_t + (1-\beta_1)\,\nabla L(w_t), \label{eq:adam_m}\\
\nu_{t+1} &= \beta_2\, \nu_t + (1-\beta_2)\,\nabla L(w_t)^{\odot 2}, \label{eq:adam_nu}\\
w_{t+1} &= w_t - \eta \, P_{t+1}^{-1} \,m_{t+1}. \label{eq:adam_w}
\end{align}
Adam combines momentum with a preconditioner. The momentum $m$ is an exponential moving average of the gradient, the second moment $\nu$ is an exponential moving average of the component-wise squared gradient, and the position is updated using the momentum rescaled by the preconditioner.

One technical detail is bias correction. The equations above define the raw moments, but because $m$ and $\nu$ are initialized at zero, these estimates are biased downward during early training. To compensate, Adam defines bias-corrected first and second moments:
\begin{align}
\hat{m}_t &= \frac{m_t}{1 - \beta_1^t}, \\
\hat{\nu}_t &= \frac{\nu_t}{1 - \beta_2^t},
\end{align}
and the position update uses the corrected versions:
\begin{equation}
w_{t+1} = w_t - \eta \,P^{-1}(\hat{\nu}_{t+1})\, \hat{m}_{t+1}.
\end{equation}
The bias correction matters more when the EMA coefficient is close to 1, as it takes longer for the estimates to reach their steady-state values. RMSProp typically does not use a bias correction because $\beta_2$ is usually smaller (a typical value is 0.9, compared to 0.999 for Adam). More details on the computational implementation of the bias correction can be found in \cref{app:computation}.

For notational simplicity, we omit the bias correction below.

\subsection{Adam Rod Flow ODEs}

As with momentum methods, we define the phase-space update:
\begin{equation}
z_{t+1} = z_t + \Phi_t,
\end{equation}
where
\begin{equation}
\Phi_t = \begin{pmatrix} -\eta P_{t+1}^{-1}\bigl[\beta_1 m_t + (1-\beta_1)\nabla L(w_t)\bigr] \\ (1-\beta_1)\bigl[\nabla L(w_t) - m_t\bigr] \end{pmatrix}.
\end{equation}
The rod flow ODEs for Adam are:
\begin{align}
\frac{d\wbar}{dt} &= -\eta P^{-1}(\bar{\nu})\bigl[\beta_1 \mbar + (1-\beta_1)\gbar\bigr], \label{eq:ode_wbar_adam}\\
\frac{d\mbar}{dt} &= (1-\beta_1)\bigl[\gbar - \mbar\bigr], \label{eq:ode_mbar_adam}\\
\frac{d\nubar}{dt} &= (1-\beta_2)\,\biggl[\frac{\bigl(\nabla L_+^{\odot 2} + \nabla L_-^{\odot 2}\bigr)}{2} - \nubar\biggr]\,, \label{eq:ode_nubar_adam}\\
\frac{d\Sigma}{dt} &= \tfrac{1}{4}\bigl[\Phi_+ \otimes \Phi_+ + \Phi_- \otimes \Phi_-\bigr] - 2\,\Sigma. \label{eq:ode_sigma_adam}
\end{align}
Analogous to the relationship between gradient descent and RMSProp, the rod flow ODEs for $\wbar$, $\mbar$, and $\Sigma$ are identical to their heavy-ball counterparts, except that $\eta$ is replaced by the preconditioned step size $\tilde{\eta} = \eta P^{-1}(\bar{\nu})$. The full derivation is given in \cref{app:adam}.

\subsection{Adam at the Edge of Stability}
\label{sec:adam_eos}

\citet{cohen2022adaptive} showed that Adam at EoS equilibrates at the preconditioned sharpness threshold
\begin{equation}
\lambda_{\max}(P^{-1/2}HP^{-1/2}) = \frac{2}{\eta}\cdot\frac{1+\beta_1}{1-\beta_1}, \label{eq:adam_threshold}
\end{equation}
which is the heavy-ball threshold in the preconditioned metric set by $\nu$.

One possible explanation for why Adam is so ``forgiving'' as an optimizer is that momentum and preconditioning combine to give it higher stability thresholds than gradient descent. The momentum raises the effective sharpness threshold by a factor of $(1+\beta_1)/(1-\beta_1)$ relative to the non-momentum case. The preconditioner, meanwhile, means that it is the \emph{preconditioned} sharpness that must hover at the edge of stability. Through the acceleration-via-regularization mechanism, the actual (unpreconditioned) sharpness is free to continue rising during training.

\subsection{NAdam}

NAdam is a variant of Adam that mimics the effect of Nesterov momentum without an explicit look-ahead gradient. The momentum update is identical to Adam's, but the position update uses a modified momentum
\begin{equation} \widetilde{m}_{t+1} = \beta_1 m_{t+1} + (1-\beta_1) g_t
\end{equation} in place of $m_{t+1}$. This biases the position update towards the most recent gradient, reproducing the effect of a look-ahead step while evaluating the gradient only at the current iterate. Further details on NAdam are given in \cref{app:adam}.

%%%%%%%%%%%%%%%%%%%%%%%%%%%%%%%%%%%%%%%%%%%%%%%%%%%%%%%%%%%%%%%%%%%%%%%%%%%%%%%
% EXPERIMENTS
%%%%%%%%%%%%%%%%%%%%%%%%%%%%%%%%%%%%%%%%%%%%%%%%%%%%%%%%%%%%%%%%%%%%%%%%%%%%%%%

\section{Experiments}
\label{sec:experiments}

\begin{figure}
  \centering
  \includegraphics[width=0.32\linewidth]{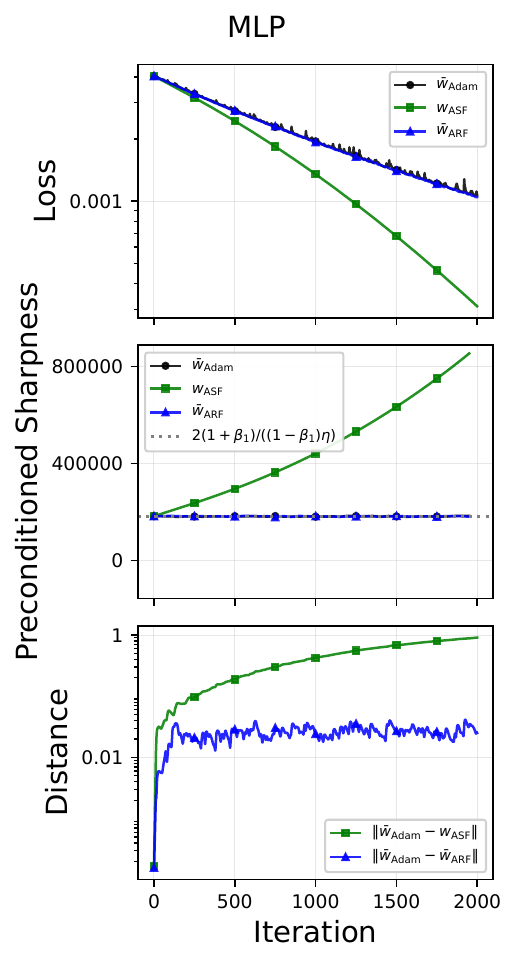}%
  \hfill
  \includegraphics[width=0.32\linewidth]{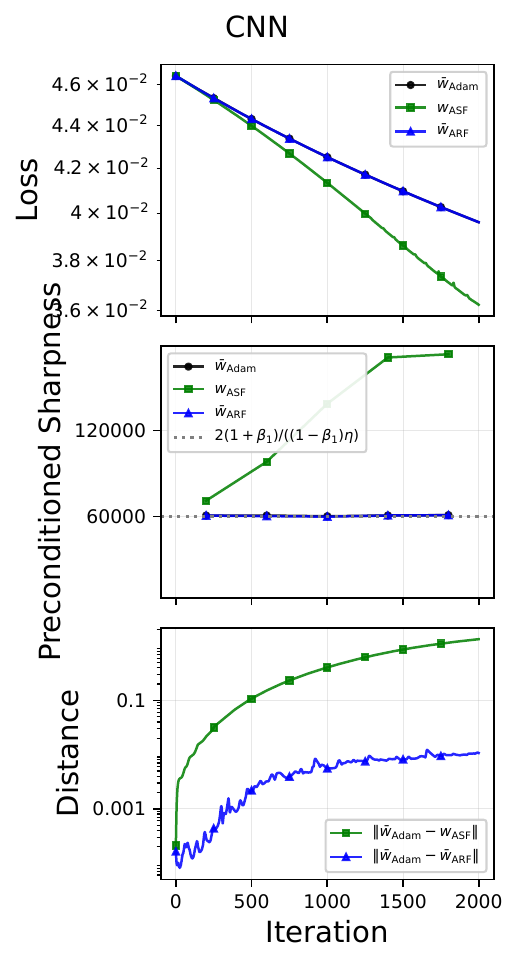}%
  \hfill
  \includegraphics[width=0.32\linewidth]{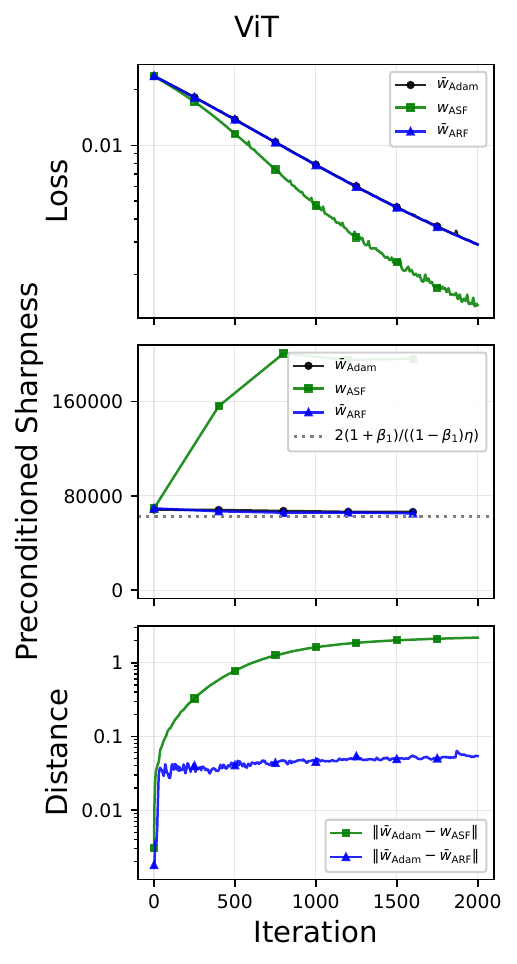}
\caption{\textbf{Experimental Results.} \textbf{Left:} MLP ($\eta = 10^{-4}$, $\beta_1 = 0.8$, $\beta_2 = 0.999$). \textbf{Center:} CNN ($\eta = 10^{-4}$, $\beta_1 = 0.5$, $\beta_2 = 0.999$). \textbf{Right:} ViT ($\eta = 7.5 \times 10^{-5}$, $\beta_1 = 0.4$, $\beta_2 = 0.999$). Within each column, the top panel shows the loss over time, the middle panel shows the effective sharpness over time, and the bottom panel shows the distance between the centers of the stable flow and rod flow relative to the discrete Adam trajectory.}
  \label{fig:centers_comparison}
\end{figure}

We evaluate the Adam rod flow on three representative neural network architectures (MLP, CNN, and ViT~\citep{Dos+21ViT}). We compare the rod flow to the discrete update and the ``stable flow''---the na\"{\i}ve continuous-time limit of the discrete optimizer.

\subsection{Setup}

All three models are trained full-batch on the same subset of $5{,}000$ examples from CIFAR-10~\citep{Kri09CIFAR}. The MLP is a 3-layer fully-connected network, the CNN is a 2-layer convolutional network, and the ViT is a 3-layer vision transformer.

For each experiment, we tracked the loss, sharpness, distance between centers, and other relevant comparison quantities. As in \citet{regis2026rod}, we used a warm-up phase so that each flow was initialized in the steady-state edge-of-stability regime. Full architectural and hyperparameter details, including the warm-up procedure, can be found in \cref{app:experiments}.

\subsection{Results}

As shown in \cref{fig:centers_comparison}, the Adam rod flow tracks the center of the discrete iterates through the edge-of-stability regime on all three architectures, matching it several orders of magnitude more closely than the stable flow does. It also stabilizes at the correct value of the preconditioned sharpness. The rod flows for the other optimizers perform similarly (\cref{app:results}).

One limitation is that rod flow struggles to accurately track the iterates for larger values of the momentum coefficient $\beta$. This may be due to the prevalence of period-doubling bifurcations: if the iterates oscillate in $4$-cycles rather than $2$-cycles, the assumptions underlying rod flow break down. 

%%%%%%%%%%%%%%%%%%%%%%%%%%%%%%%%%%%%%%%%%%%%%%%%%%%%%%%%%%%%%%%%%%%%%%%%%%%%%%%
% DISCUSSION
%%%%%%%%%%%%%%%%%%%%%%%%%%%%%%%%%%%%%%%%%%%%%%%%%%%%%%%%%%%%%%%%%%%%%%%%%%%%%%%

\section{Conclusion}
\label{sec:discussion}

\paragraph{Limitations.} Rod flow is a full-batch model and does not apply to the mini-batch setting. It also assumes the steady-state edge-of-stability regime, and does not model the transient early-training phase before oscillations equilibrate. Furthermore, it remains too costly to serve as a practical substitute for running the discrete optimizer. Finally, rod flow says nothing about \emph{progressive sharpening}---the empirical phenomenon by which sharpness drifts upward before reaching the EoS threshold---which remains not fully explained by existing theory.

\paragraph{Future work.} The most natural extensions are to the mini-batch setting, where stochastic noise may require augmenting the rod flow equations with diffusion terms, and to optimizers beyond the Adam family---such as Muon~\citep{Jor+24Muon}, Lion~\citep{Chen+23Lion}, and Shampoo~\citep{GupKorSin18Shampoo}---that share a momentum-plus-preconditioner structure. Finally, a formal characterization of when the rod flow approximation breaks down remains open.

%%%%%%%%%%%%%%%%%%%%%%%%%%%%%%%%%%%%%%%%%%%%%%%%%%%%%%%%%%%%%%%%%%%%%%%%%%%%%%%
% ACKNOWLEDGEMENTS
%%%%%%%%%%%%%%%%%%%%%%%%%%%%%%%%%%%%%%%%%%%%%%%%%%%%%%%%%%%%%%%%%%%%%%%%%%%%%%%

\section*{Acknowledgements}

We thank the authors of the Central Flow paper for making their code publicly available. Our implementation builds on code from \url{https://github.com/alex-damian/EOS} and \url{https://github.com/centralflows/centralflows}.

\newpage
\bibliography{references}

%%%%%%%%%%%%%%%%%%%%%%%%%%%%%%%%%%%%%%%%%%%%%%%%%%%%%%%%%%%%%%%%%%%%%%%%%%%%%%%
\newpage
\appendix

\section{Notation}\label{app:notation}

\begin{table}[H]
\centering
\caption{Unified notation used throughout the paper.}
\label{tab:notation}
\begin{tabular}{cl}
\toprule
Symbol & Meaning \\
\midrule
$\eta$ & Learning rate \\
$\beta_1$, $\beta_2$ & First-moment and second-moment decay rates \\
$\varepsilon$ & Numerical stability constant \\
\midrule
$w_t$ & Parameters (position) at iterate $t$ \\
$m_t$ & First moment (EMA of gradient) at iterate $t$ \\
$\nu_t$ & Second moment (EMA of squared gradient) at iterate $t$ \\
$\hat{m}_t, \hat{\nu}_t$ & Bias-corrected first and second moments \\
$g_t = \nabla L(w_t)$ & Gradient at iterate $t$ \\
$v^{\odot 2}$ & Elementwise square of a vector $v$ \\
$P$ & Preconditioner \\
\midrule
$\wbar_t, \mbar_t, \nubar_t$ & Midpoint of consecutive iterates \\
$\delta_t$ & Position half-difference \\
$\gamma_t$ & First-moment half-difference \\
$\nabla L_\pm$  & Gradients at the two rod endpoints \\
$\gbar$  & Average of the endpoint gradients \\
\midrule
$z_t = (w_t, m_t)$ & Phase-space vector \\
$\Delta_t = (\delta_t, \gamma_t)$ & Phase-space half-difference \\
$\Phi(z)$ & Phase-space displacement \\
$\Sigma$ & Extent tensor \\
$\Sigma_{\delta\delta}, \Sigma_{\delta\gamma}, \Sigma_{\gamma\delta}, \Sigma_{\gamma\gamma}$ & $d \times d$ blocks of the phase-space extent tensor \\
\bottomrule
\end{tabular}
\end{table}

\section{Rod Flow for Gradient Descent}
\label{app:gradient_descent}

We briefly rederive the gradient descent rod flow, first introduced in \cite{regis2026rod}.

Recall the update equation for gradient descent:
\begin{equation}
w_{t+1} = w_t - \eta \nabla L(w_t)    
\end{equation}
Define the midpoint and half-difference of consecutive iterates:
\begin{align}
\wbar_t &= \tfrac{1}{2}(w_{t+1} + w_t) \\
\delta_t &= \tfrac{1}{2}(w_{t+1} - w_t)
\end{align}
At the edge of stability, the GD iterates oscillate every step about their running center. However, even while the individual iterates oscillate, the midpoint $\bar{w}_t$ and the outer product $\delta_t \otimes \delta_t$ vary smoothly. So we can hope to model these two quantities in continuous time.

We seek recurrence equations for $\bar{w}_t$ and $\delta_t \otimes \delta_t$ satisfying two conditions, each motivated by the eventual promotion to an ODE.

\begin{enumerate}

\item The right-hand sides should be expressible purely in terms of $\bar{w}_t$ and $\delta_t \otimes \delta_t$, without reference to the raw iterates $w_t$. Otherwise, the system is not closed in the variables we want to pass to continuous time.

\item The recurrences should be invariant under $\delta \mapsto -\delta$. In continuous time, the sign of $\delta$ will no longer be a meaningful distinction.

\end{enumerate}

Note that the original iterates can be recovered from the transformed variables via:
\begin{align}
w_t &= \bar{w}_t - \delta_t, \\
w_{t+1} &= \bar{w}_t + \delta_t,
\end{align}
and that the half-displacement can be expressed in terms of the gradient as:
\begin{equation}
\delta_t = -\frac{\eta}{2} \nabla L(w_t).
\end{equation}
The difference equation for the midpoint is:
\begin{align*}
\wbar_{t+1} - \wbar_t &= \frac{w_{t+2} + w_{t+1}}{2} - \frac{w_{t+1} + w_t}{2} \\
&= \frac{w_{t+2} - w_{t+1}}{2} + \frac{w_{t+1} - w_t}{2} \\
&= -\frac{\eta}{2}\bigl[\nabla L(w_{t+1}) + \nabla L(w_t)\bigr] \\
&= -\frac{\eta}{2}\bigl[\nabla L(\wbar_t + \delta_t) + \nabla L(\wbar_t - \delta_t)\bigr].
\end{align*}
And the difference equation for the outer product of the half-difference is:
\begin{align*}
\delta_{t+1} \otimes \delta_{t+1} - \delta_t \otimes \delta_t &= \delta_{t+1}\otimes\delta_{t+1} - \delta_t\otimes\delta_t + (\delta_t \otimes \delta_t - \delta_t \otimes \delta_t) \\
&= \delta_{t+1} \otimes \delta_{t+1} + \delta_t \otimes \delta_t - 2 \delta_t \otimes \delta_t \\
&= \frac{\eta^2}{4}\bigl[\nabla L_+ \otimes \nabla L_+ + \nabla L_- \otimes \nabla L_-\bigr] - 2\,\delta_t \otimes \delta_t.
\end{align*}
It should be emphasized that both difference equations are \textit{exact}. The only approximation in the rod flow framework is the step where we promote these difference equations to ODEs.

Let $\Sigma(t)$ denote the continuous-time analog of $\delta \otimes \delta$. We obtain the following rod flow ODEs for gradient descent:
\begin{align}
\frac{d\bar{w}}{dt} &= -\eta \,\bar{g} \\
\frac{d \Sigma}{dt} &= \frac{\eta^2}{4}\, \bigl (\nabla L_+ \otimes \nabla L_+ + \nabla L_- \otimes \nabla L_- \bigr ) - 2 \Sigma
\end{align}
where we identify $\delta$ with the principal eigenvector of $\Sigma$ scaled by the square root of the principal eigenvalue.

We introduce the $\Sigma$ notation for two reasons. First, in continuous time, $\Sigma$ will no longer be exactly rank-one, so denoting the extent as $\delta \otimes \delta$ would be misleading. One should note though that $\Sigma$ does remain \textit{approximately} rank-one: during experiments, its largest eigenvalue is consistently several orders of magnitude larger than the second-largest eigenvalue. Second, this notation aligns with Central Flow \citep{Cohen+25CentralFlow}, where $\Sigma$ plays the role of the covariance matrix of the oscillations.

While the extent tensor of rod flow and the covariance matrix of Central Flow play similar roles, there are important conceptual differences. Rod flow's extent tensor is always approximately rank-one---even when there are multiple sharp directions, or when there are no oscillations at all (as is the case when the iterates are \textit{not} at the edge of stability). The covariance matrix of Central Flow, by contrast, has rank equal to the number of sharp directions.

\section{Rod Flow for Heavy Ball and Nesterov Momentum}
\label{app:momentum}

\subsection{Derivation of Rod Flows}

\subsubsection{Heavy Ball}
The heavy ball momentum update equations are:
\begin{align}
m_{t+1} &= \beta\, m_t + (1-\beta)\,\nabla L(w_t), \label{eq:hb_update_momentum_app}\\
w_{t+1} &= w_t - \eta\, m_{t+1}, \label{eq:hb_update_position_app}
\end{align}
where $\beta \in [0,1)$ is the momentum coefficient.

 When heavy ball crosses into EoS, $w$ and $m$ oscillate in phase: both flip sign about the center each iteration. Because the position and momentum oscillate together, it is natural to concatenate them into a single phase-space vector $z$:
\begin{equation}
z_t = \begin{pmatrix} w_t \\ m_t \end{pmatrix} \in \R^{2d}\,.
\end{equation}
In analogy with the rod flow for gradient descent, define the average of two consecutive phase-space iterates $\bar{z}$ and the half-displacement between them $\Delta$:
\begin{align}
\bar{z}_t &= \tfrac{1}{2}(z_{t+1} + z_t) = \begin{pmatrix} \bar{w}_t \\ \bar{m}_t \end{pmatrix}, \\
\Delta_t &= \tfrac{1}{2}(z_{t+1} - z_t) = \begin{pmatrix} \delta_t \\ \gamma_t \end{pmatrix},
\end{align}
where 
\begin{align}
\bar{m}_t &= \frac{m_{t+1} + m_t}{2} \\
\gamma_t &= \frac{m_{t+1} - m_t}{2}
\end{align} 
are the momentum midpoint and the momentum half-difference, respectively.

We emphasize that $(\delta, \gamma)$ forms a \emph{single} rod in phase space, not two separate rods in position and momentum space: the formalism is symmetric under the simultaneous flip $(\delta, \gamma) \mapsto (-\delta, -\gamma)$, but not under flipping just one of the coordinates. This is why the phase-space structure is necessary: when we pass to continuous time, we will extract $\delta$ and $\gamma$ jointly from the principal eigenvector of $\Sigma$, and the phase-space structure ensures their signs are consistent.

Write the phase-space update as:
\begin{equation}
z_{t+1} = z_t + \Phi(z_t), \qquad \Phi(z) = \begin{pmatrix} -\eta\bigl[\beta m + (1-\beta)\nabla L(w)\bigr] \\ (1-\beta)\bigl[\nabla L(w) - m\bigr] \end{pmatrix}. \label{eq:Phi_hb_app}
\end{equation}
As in the gradient descent case, the phase-space midpoint $\bar{z}_t$ and the outer product $\Delta_t \otimes \Delta_t$ both evolve smoothly at EoS, so these are the quantities we model in continuous time. Writing $\Phi_t \coloneqq \Phi(z_t)$, the difference equation for $\bar{z}$ is
\begin{align*}
\bar{z}_{t+1} - \bar{z}_t &= \tfrac{1}{2}(z_{t+2} + z_{t+1}) - \tfrac{1}{2}(z_{t+1} + z_t) \\
&= \tfrac{1}{2}(\Phi_{t+1} + \Phi_t) \\
&= \tfrac{1}{2}\bigl[\Phi(\bar{z}_t + \Delta_t) + \Phi(\bar{z}_t - \Delta_t)\bigr],
\end{align*}
where the last line uses $z_{t+1} = \bar{z}_t + \Delta_t$ and $z_t = \bar{z}_t - \Delta_t$. 

For the difference equation for $\Delta\otimes \Delta$:
\begin{align*}
\Delta_{t+1} \otimes \Delta_{t+1} - \Delta_t \otimes \Delta_t &= \Delta_{t+1} \otimes \Delta_{t+1} - \Delta_t \otimes \Delta_t + (\Delta_t\otimes \Delta_t - \Delta_t \otimes \Delta_t) \\
&= \Delta_{t+1} \otimes \Delta_{t+1} + \Delta_t \otimes \Delta_t - 2 \Delta_t \otimes \Delta_t \\
&= \frac{1}{4}  ( \Phi_+ \otimes \Phi_+ + \Phi_- \otimes \Phi_- ) - 2 \Delta_t \otimes \Delta_t
\end{align*}
As in the gradient descent case, both difference equations are exact.

Promoting the difference equations to ODEs gives the rod flow for heavy ball momentum:
\begin{align}
\frac{d\bar{w}}{dt} &= -\eta\bigl[\beta\, \bar{m} + (1-\beta)\,\bar{g}\bigr], \label{eq:ode_wbar_hb_app}\\
\frac{d\bar{m}}{dt} &= (1-\beta)\bigl[\bar{g} - \bar{m}\bigr], \label{eq:ode_mbar_hb_app}\\
\frac{d\Sigma}{dt} &= \tfrac{1}{4}\bigl[\Phi_+ \otimes \Phi_+ + \Phi_- \otimes \Phi_-\bigr] - 2\,\Sigma, \label{eq:ode_sigma_hb_app}
\end{align}
with $\bar{g} = (\nabla L_+ + \nabla L_-)/2$. Note that the extent tensor $\Sigma \in \R^{2d \times 2d}$ decomposes into four $d \times d$ blocks:
\begin{equation}
\Sigma = \begin{pmatrix} \Sigma_{\delta\delta} & \Sigma_{\delta\gamma} \\ \Sigma_{\gamma\delta} & \Sigma_{\gamma\gamma} \end{pmatrix}.
\end{equation}
As before, we identify $\Delta = (\delta, \gamma)$ with the principal eigenvector of $\Sigma$ scaled by the square root of the principal eigenvalue.

\subsubsection{Nesterov}

Nesterov momentum is very similar to heavy ball momentum. The only difference between the two momentum schemes is where the gradient is evaluated. Rather than evaluating the gradient at $w_t$, Nesterov momentum evaluates the gradient at a \textit{look-ahead point} $\theta_t$, obtained by extrapolating $w_t$ along the current momentum:
\begin{align}
\theta_t &= w_t - \eta\beta\, m_t, \label{eq:nest_theta}\\
m_{t+1} &= \beta\, m_t + (1-\beta)\,\nabla L(\theta_t), \label{eq:nest_update_m}\\
w_{t+1} &= w_t - \eta\, m_{t+1}. \label{eq:nest_update_w}
\end{align}
Nesterov originally introduced this look-ahead to obtain an accelerated convergence rate of $O(1/T^2)$ on smooth convex objectives~\citep{nesterov1983method}, improving on the $O(1/T)$ rate of vanilla gradient descent. Heuristically, the look-ahead acts as a one-step prediction--correction: we use the current value of the momentum to predict where the iterate is heading, and then query the gradient there for a course correction.

For the Nesterov rod flow, the only change relative to heavy ball is that $\nabla L(w)$ is replaced by $\nabla L(w - \eta\beta\, m)$ inside $\Phi$. The phase-space update becomes:
\begin{equation}
\Phi(z) = \begin{pmatrix} -\eta\bigl[\beta m + (1-\beta)\nabla L(w - \eta\beta m)\bigr] \\ (1-\beta)\bigl[\nabla L(w - \eta\beta m) - m\bigr] \end{pmatrix}. \label{eq:Phi_nest}
\end{equation}
The phase-space midpoint $\bar{z}_t$ and phase-space half-displacement $\Delta_t = (\delta_t, \gamma_t)$ are defined exactly as in the heavy ball case. And the difference equations for $\bar{z}$ and $\Delta\otimes\Delta$ remain identical in form:
\begin{align*}
\bar{z}_{t+1} - \bar{z}_t &= \tfrac{1}{2}\bigl[\Phi(\bar{z}_t + \Delta_t) + \Phi(\bar{z}_t - \Delta_t)\bigr],\\
\Delta_{t+1} \otimes \Delta_{t+1} - \Delta_t \otimes \Delta_t &= \tfrac{1}{4}\bigl[\Phi_+ \otimes \Phi_+ + \Phi_- \otimes \Phi_-\bigr] - 2\,\Delta_t \otimes \Delta_t.
\end{align*}
Expanding $\Phi(\bar{z}_t \pm \Delta_t)$, the gradient evaluation point becomes:
\[
(\bar{w} \pm \delta) - \eta\beta\,(\bar{m} \pm \gamma) \;=\; \bigl(\bar{w} - \eta\beta\,\bar{m}\bigr) \;\pm\; \bigl(\delta - \eta\beta\,\gamma\bigr) \;=\; \bar{\theta} \;\pm\; \varphi,
\]
where $\bar{\theta} = \bar{w} - \eta\beta\,\bar{m}$ is the look-ahead midpoint and $\varphi = \delta - \eta\beta\,\gamma$ is the look-ahead-shifted half-displacement.

Promoting the discrete difference equations to ODEs yields the rod flow for Nesterov momentum:
\begin{align}
\frac{d\bar{w}}{dt} &= -\eta\bigl[\beta\, \bar{m} + (1-\beta)\,\bar{g}\bigr], \label{eq:ode_wbar_nest}\\
\frac{d\bar{m}}{dt} &= (1-\beta)\bigl[\bar{g} - \bar{m}\bigr], \label{eq:ode_mbar_nest}\\
\frac{d\Sigma}{dt} &= \tfrac{1}{4}\bigl[\Phi_+ \otimes \Phi_+ + \Phi_- \otimes \Phi_-\bigr] - 2\,\Sigma, \label{eq:ode_sigma_nest}
\end{align}
identical in form to the heavy ball rod flow, with the sole modification that:
\[
\bar{g} = \tfrac{1}{2}\bigl[\nabla L(\bar{\theta} + \varphi) + \nabla L(\bar{\theta} - \varphi)\bigr]
\]

\subsection{Derivation of the Sharpness Threshold on Quadratic Loss}

\subsubsection{Heavy Ball}
Just as with gradient descent, heavy ball momentum on a quadratic loss has a sharpness threshold beyond which the iterates no longer converge. However, this threshold is no longer $2/\eta$: it is modified by the momentum.

To find this critical value of the sharpness, we first express the momentum update as a two-step recurrence in $w$ alone. Recall the heavy ball update equations:
\begin{align*}
m_{t+1} &= \beta\, m_t + (1-\beta)\,\nabla L(w_t),\\
w_{t+1} &= w_t - \eta\, m_{t+1}.
\end{align*}
From the position update, we can isolate $m_{t+1}$:
\begin{equation}
m_{t+1} = -\frac{1}{\eta}(w_{t+1} - w_t)\,.
\end{equation}
Shifting the index back by one likewise gives $m_t = -\tfrac{1}{\eta}(w_t - w_{t-1})$. Substituting both into the momentum update yields:
\begin{equation}
-\frac{1}{\eta}(w_{t+1} - w_t) = -\frac{\beta}{\eta}(w_t - w_{t-1}) + (1-\beta)\,\nabla L(w_t)\,.
\end{equation}
Now consider the quadratic loss $L(w) = \tfrac{1}{2} S w^2$. We have that $\nabla L(w) = S w$. Substituting and rearranging gives the two-step recurrence:
\begin{equation}
w_{t+1} = \bigl(1 + \beta - \eta(1-\beta)S\bigr)\, w_t - \beta\, w_{t-1}\,.
\end{equation}
We can express this recurrence as a matrix equation. Let $x_t = \begin{bmatrix} w_t \\ w_{t-1} \end{bmatrix}$ be the state vector. Then we have that:
\begin{equation}
x_{t+1} = M x_t, \qquad M = \begin{bmatrix} 1 + \beta - \eta(1-\beta)S & -\beta \\ 1 & 0 \end{bmatrix}.
\end{equation}
The dynamics of the system are governed by the eigenvalues $\lambda$ of the transition matrix $M$. For the iterates to converge, both eigenvalues must lie within the unit circle ($|\lambda| \leq 1$). We can solve for the eigenvalues using the characteristic equation $\Det(M - \lambda I) = 0$:
\begin{equation}
\lambda^2 - \bigl(1 + \beta - \eta(1-\beta)S\bigr)\lambda + \beta = 0\,.
\end{equation}
The coefficients of this quadratic are real, so its roots are either both real or a complex-conjugate pair. In the complex-conjugate case, the two roots have equal magnitude. And because their product satisfies $\lambda_1 \lambda_2 = \beta$, that common magnitude is $\sqrt{\beta} < 1$ (using $\beta \in [0,1)$). So complex roots cannot leave the unit circle.

The eigenvalues can only cross the boundary along the real axis: at $\lambda = 1$ or $\lambda = -1$. The $\lambda = 1$ case is degenerate, corresponding to stationary dynamics. The $\lambda = -1$ case corresponds to a 2-period cycle---which is precisely what we are looking for. Substituting $\lambda = -1$ into the characteristic equation gives the condition for the sharpness threshold $S^\ast$:
\begin{equation}
2 + 2\beta - \eta(1-\beta)S^\ast = 0 \implies S^\ast = \frac{2}{\eta} \cdot \frac{1+\beta}{1-\beta}\,.
\end{equation}
At the critical sharpness $S^\ast$, we know that $\lambda_1 = -1$ is an eigenvalue. As the product of the eigenvalues equals $\beta$, it immediately follows that the second eigenvalue is:
\begin{equation}
\lambda_2 = -\beta\,.
\end{equation}
Because the momentum coefficient $\beta \in [0,1)$, this second eigenvalue has magnitude strictly less than 1. This implies that, at the sharpness threshold, the system does not globally diverge. Rather, there is a one-dimensional family of 2-period orbits, and the iterates decay toward this family at a rate of $\beta$. 

To see this explicitly, we can decompose the system's initial conditions into the eigenvectors of $M$. At $S = S^\ast$, the transition matrix becomes:
\begin{equation}
M = \begin{bmatrix} -(1+\beta) & -\beta \\ 1 & 0 \end{bmatrix}\,.
\end{equation}
Solving $(M - \lambda I)v = 0$ for each eigenvalue yields the corresponding eigenvectors:
\begin{equation}
v_1 = \begin{bmatrix} 1 \\ -1 \end{bmatrix} \quad (\text{for } \lambda_1 = -1), \qquad v_2 = \begin{bmatrix} -\beta \\ 1 \end{bmatrix} \quad (\text{for } \lambda_2 = -\beta)\,.
\end{equation}
The vector $v_1$ represents the persistent 2-period oscillation, while $v_2$ represents the decaying transient. 

Let $w_0$ and $m_0$ be our initial position and momentum, respectively. We can express the initial state $x_0 = \begin{bmatrix} w_0 & w_{-1} \end{bmatrix}^\top$ in terms of $w_0$ and $m_0$. Rearranging the position update equation, one can see that $w_{-1} = w_0 + \eta m_0$. We then have that:
\begin{equation}
x_0 = \begin{bmatrix} w_0 \\ w_0 + \eta m_0 \end{bmatrix}\,.
\end{equation}
We want to decompose this initial state into the eigenbasis of $M$:
\begin{equation}
x_0 = c_1 v_1 + c_2 v_2
\end{equation}
Which yields the following linear system of equations for the coefficients:
\begin{align}
w_0 &=  c_1 - \beta c_2 \,, \\
w_0 + \eta m_0 &= -c_1 + c_2 \,.
\end{align}
Solving for the coefficients gives:
\begin{equation}
c_1 = \frac{(1+\beta)w_0 + \beta \eta m_0}{1-\beta}, \qquad c_2 = \frac{2w_0 + \eta m_0}{1-\beta}\,.
\end{equation}
The dynamics over time are given by:
\begin{equation} 
x_t = M^t x_0 = c_1(-1)^t v_1 + c_2(-\beta)^t v_2
\end{equation}
And we have that as $t \to \infty$:
\begin{equation}
x_\infty = c_1 v_1 \implies |w_\infty| = |c_1|
\end{equation}
Specifically, in the case that $m_0 = 0$, we have that:
\begin{equation}
|w_\infty| = \frac{1+\beta}{1- \beta} |w_0|
\end{equation}

\begin{figure}[htbp]
  \centering
  \includegraphics[width=\linewidth]{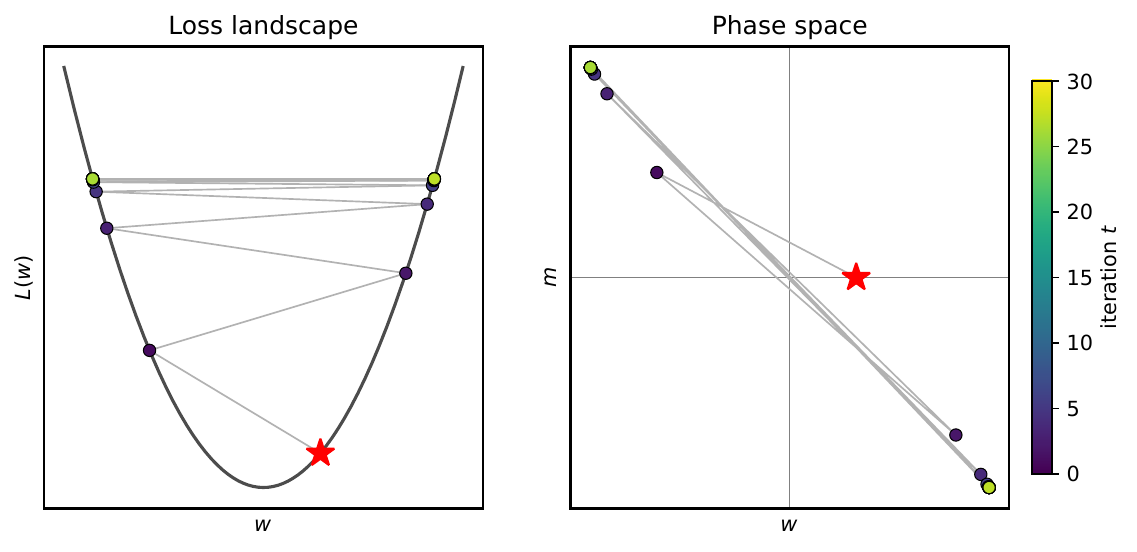}
  \caption{\textbf{Heavy Ball at the EoS Threshold.} Dynamics of Heavy Ball momentum on a 1D quadratic loss, illustrating convergence to a persistent 2-period orbit at the sharpness threshold.}
  \label{fig:hb_threshold_quadratic}
\end{figure}

\subsubsection{Nesterov}

The same approach yields the sharpness threshold for Nesterov momentum on a quadratic.

Recall the update equations for Nesterov momentum:
\begin{align*}
\theta_t &= w_t - \eta\beta\, m_t,\\
m_{t+1} &= \beta\, m_t + (1-\beta)\,\nabla L(\theta_t),\\
w_{t+1} &= w_t - \eta\, m_{t+1}.
\end{align*}
Once again, we aim to eliminate $m$ to obtain a two-step recurrence in $w$ alone. From the position update, we have:
\begin{equation*}
m_{t+1} = -\frac{1}{\eta}(w_{t+1} - w_t),
\end{equation*}
from which shifting the index back by one gives $m_t = -\tfrac{1}{\eta}(w_t - w_{t-1})$. Substituting this into the look-ahead point yields
\begin{align}
\theta_t &= w_t - \eta\beta\, m_t \nonumber \\ 
&= w_t + \beta(w_t - w_{t-1}) \nonumber\\
&= (1+\beta)\, w_t - \beta\, w_{t-1}\,. \label{eq:nest_theta_recurrence}
\end{align}
Consider the quadratic loss $L(w) = \tfrac{1}{2} S w^2$. One can see that $\nabla L(w) = S w$. Substituting the above expressions for the momentum, the look-ahead, and the gradient into the update equation for the momentum yields:
\begin{equation}
-\frac{1}{\eta}(w_{t+1} - w_t) = -\frac{\beta}{\eta}(w_t - w_{t-1}) + (1-\beta)\,S\bigl[(1+\beta) w_t - \beta\, w_{t-1}\bigr]\,.
\end{equation}
Rearranging to isolate $w_{t+1}$ yields:
\begin{equation}
w_{t+1} = \bigl(1 + \beta - \eta(1-\beta^2)\,S\bigr)\, w_t \;-\; \beta\bigl(1 - \eta(1-\beta)\,S\bigr)\, w_{t-1}\,.
\end{equation}
Let $x_t = \begin{bmatrix} w_t \\ w_{t-1}\end{bmatrix}$. We have the following two-step recurrence:
\begin{equation}
x_{t+1} = M x_t, \qquad M = \begin{bmatrix} 1+\beta - \eta(1-\beta^2)\,S & -\beta\bigl(1 - \eta(1-\beta) S\bigr) \\ 1 & 0 \end{bmatrix}.
\end{equation}
The characteristic equation for the transition matrix $M$ is:
\begin{equation}
\lambda^2 - \bigl(1 + \beta - \eta(1-\beta^2) S\bigr)\,\lambda + \beta\bigl(1 - \eta(1-\beta) S\bigr) = 0\,.
\end{equation}
The characteristic equation for Nesterov momentum is somewhat more complicated than the one for heavy ball. For example, in heavy ball the constant term was simply $\beta$, whereas here it involves both the step size and the sharpness of the quadratic.

As before, we find the sharpness threshold $S^\ast$ by substituting $\lambda = -1$ into the characteristic equation:
\begin{equation}
2(1+\beta) - (1-\beta)(1+2\beta)\,\eta S^\ast = 0\,.
\end{equation}
Isolating $S^\ast$ gives:
\begin{equation}
S^\ast = \frac{2}{\eta} \cdot \frac{1+\beta}{(1-\beta)(1+2\beta)}\,.
\end{equation}
Compared to the heavy ball threshold $S^\ast_{\text{HB}} = \tfrac{2}{\eta} \cdot \tfrac{1+\beta}{1-\beta}$, the look-ahead tightens the threshold by an extra factor of $1/(1+2\beta)$: at the same $\eta$ and $\beta$, Nesterov destabilizes at a strictly smaller sharpness threshold than heavy ball.

We want to find the other eigenvalue of $M$ at $S = S^\ast$. To do so, we can use the fact that the product of the eigenvalues equals the constant term of the characteristic equation. Substituting $S^\ast$ into our expression for the constant term yields:
\begin{align*}
\lambda_1 \lambda_2 &= \beta\bigl(1 - \eta(1-\beta)\, S^\ast\bigr) \\
&= \beta\left(1 - \eta(1-\beta) \cdot \frac{2}{\eta} \cdot \frac{1+\beta}{(1-\beta)(1+2\beta)}\right) \\
&= \beta\left(1 - \frac{2(1+\beta)}{1+2\beta}\right) \\
&= \beta \cdot \frac{(1+2\beta) - 2(1+\beta)}{1+2\beta} \\
&= -\frac{\beta}{1+2\beta}\,.
\end{align*}
Recall that, because we are at the sharpness threshold, $\lambda_1 = -1$. It then follows that:
\begin{equation}
\lambda_2 = \frac{\beta}{1+2\beta}\,.
\end{equation}
Since $\beta \in [0,1)$, we have that $\lambda_2 \in [0, \tfrac{1}{3})$. So once again, the iterates decay onto a one-dimensional family of 2-period orbits. But notably: the transient decays more quickly than was the case for heavy ball, where the analogous decay rate was $\beta$ rather than $\beta/(1+2\beta)$.

At $S = S^\ast$, the transition matrix simplifies to:
\begin{equation}
M = \begin{bmatrix} -(1+\beta)/(1+2\beta) & \beta/(1+2\beta) \\ 1 & 0 \end{bmatrix}\,.
\end{equation}
Solving $(M - \lambda I)\, v = 0$ for each eigenvalue gives the corresponding eigenvector:
\begin{equation}
v_1 = \begin{bmatrix} 1 \\ -1 \end{bmatrix} \;\;(\text{for } \lambda_1 = -1), \qquad v_2 = \begin{bmatrix} \beta \\ 1+2\beta \end{bmatrix} \;\;\bigl(\text{for } \lambda_2 = \tfrac{\beta}{1+2\beta}\bigr)\,.
\end{equation}
The persistent oscillation direction $v_1$ is identical to the heavy-ball case and again corresponds to symmetric bouncing about the minimum. However, the transient direction $v_2$ has changed.

Let $w_0$ and $m_0$ be our initial position and momentum, respectively. Using the position update equation, we have that our initial state is:
\begin{equation}
x_0 = \begin{bmatrix} w_0 \\ w_0 + \eta m_0 \end{bmatrix}\,.
\end{equation}
We can express our initial state as a linear combination of the eigenvectors of $M$: 
\begin{equation}
x_0 = c_1 v_1 + c_2 v_2
\end{equation}
We can then obtain a linear system of equations for the coefficients $c_1$ and $c_2$:
\begin{align}
w_0 &= c_1 + \frac{\beta}{1+2\beta}\,c_2\,, \\
w_0 + \eta m_0 &= -c_1 + c_2\,,
\end{align}
Solving for the coefficients:
\begin{equation}
c_1 = \frac{(1+\beta)\, w_0 - \beta\,\eta\, m_0}{1+3\beta}\,, \qquad c_2 = \frac{(2 w_0 + \eta m_0)(1+2\beta)}{1+3\beta}\,.
\end{equation}
So our state evolves over time according to:
\begin{equation}
x_t = M^t x_0 = c_1 (-1)^t v_1 + c_2 \Bigl(\tfrac{\beta}{1+2\beta}\Bigr)^t v_2\,.
\end{equation}
The asymptotic 2-period oscillation amplitude is $|w_\infty| = |c_1|$. In particular, when $m_0 = 0$, we have:
\begin{equation}
|w_\infty| = \frac{1+\beta}{1+3\beta}\, |w_0|\,.
\end{equation}
Note that
\begin{equation}
\frac{1+\beta}{1+3\beta} < \frac{1+\beta}{1-\beta}
\end{equation}
for any $\beta \in (0,1)$. Hence, for the same initial position and momentum coefficient, Nesterov momentum settles into a strictly smaller-amplitude 2-period orbit than heavy ball.

\subsection{Theoretical Analysis}

As was the case with the gradient descent rod flow, it's instructive to analyze the behavior of the momentum rod flows on three one-dimensional loss functions: the linear loss, the quadratic loss, and the quartic loss.

Let $\Phi^w$ and $\Phi^m$ denote the position and momentum components of the phase-space update, respectively. For heavy ball:
\begin{align}
\Phi^w_{\text{HB}} &= -\eta\bigl[\beta\, m + (1-\beta)\,\nabla L(w)\bigr], \\
\Phi^m_{\text{HB}} &= (1-\beta)\bigl[\nabla L(w) - m\bigr].
\end{align}
And for Nesterov:
\begin{align}
\Phi^w_{\text{NAG}} &= -\eta\bigl[\beta\, m + (1-\beta)\,\nabla L(w - \eta\beta\, m)\bigr], \\
\Phi^m_{\text{NAG}} &= (1-\beta)\bigl[\nabla L(w - \eta\beta\, m) - m\bigr].
\end{align}
Throughout, we drop the subscripts from $\Phi$ when it is clear from context whether we are working with heavy ball or Nesterov momentum.

When our loss landscape is one-dimensional, momentum rod flow reduces down to just six scalar ODEs:
\begin{align}
\frac{d \bar{w}}{dt} &= - \eta [\beta \bar{m} + (1-\beta)\bar{g}] \\
\frac{d \bar{m}}{dt} &= (1-\beta)(\bar{g} - \bar{m}) \\
\frac{d \Sigma_{\delta \delta}}{dt} &= \frac{1}{4}\bigl((\Phi^w_+)^2 + (\Phi^w_-)^2 \bigr) - 2 \Sigma_{\delta \delta}\\
\frac{d \Sigma_{\gamma \gamma}}{dt} &= \frac{1}{4}\bigl((\Phi^m_+)^2 + (\Phi^m_-)^2 \bigr) - 2 \Sigma_{\gamma \gamma}\\
\frac{d \Sigma_{\delta \gamma}}{dt} &= \frac{1}{4}\bigl((\Phi^w_+)(\Phi^m_+) + (\Phi^w_-)(\Phi^m_-) \bigr) - 2 \Sigma_{\delta \gamma}\\
\frac{d \Sigma_{\gamma \delta}}{dt} &=  \frac{1}{4}\bigl((\Phi^m_+)(\Phi^w_+) + (\Phi^m_-)(\Phi^w_-) \bigr) - 2 \Sigma_{\gamma \delta}
\end{align}
Because we constrain $\Sigma$ to be symmetric, $\Sigma_{\delta\gamma} = \Sigma_{\gamma\delta}$. Without loss of generality, we will only consider the dynamics of $\Sigma_{\delta\gamma}$.

\subsubsection{Linear Loss}

Consider the linear loss $L(w) = bw$. The gradient is constant: $\nabla L = b$. Because the gradient is independent of the evaluation point, heavy ball and Nesterov momentum behave identically on the linear loss.

The $\bar{m}$ ODE is:
\begin{equation}
\frac{d\bar{m}}{dt} = (1-\beta)(b - \bar{m})\,.
\end{equation}
The dynamics of $\bar{m}$ are independent of the other variables. At steady state, $\bar{m}^\ast = b$. This makes sense: the momentum is an exponential moving average of the gradient, and the gradient is constant. The $\bar{w}$ ODE is then:
\begin{equation}
\frac{d\bar{w}}{dt} = -\eta b\,.
\end{equation}
So the midpoint of the rod in position space moves at constant speed. At steady state, we have that:
\begin{align*}
\Phi^w_\pm &= -\eta b, \\
\Phi^m_\pm &= 0.
\end{align*}
The ODEs for the components of $\Sigma$ are then:
\begin{align}
\frac{d\Sigma_{\delta\delta}}{dt} &= \tfrac{1}{2}(\eta b)^2 - 2\,\Sigma_{\delta\delta}, \\
\frac{d\Sigma_{\gamma\gamma}}{dt} &= -2\,\Sigma_{\gamma\gamma}, \\
\frac{d\Sigma_{\delta\gamma}}{dt} &= -2\,\Sigma_{\delta\gamma}.
\end{align}
By inspection, the steady-state value of $\Sigma$ is:
\begin{equation}
\Sigma^\ast = \begin{pmatrix} \left(\tfrac{\eta b}{2}\right)^2 & 0 \\ 0 & 0 \end{pmatrix},
\end{equation}
and the corresponding steady-state value of $\Delta$ is:
\begin{equation}
\Delta^\ast = \begin{pmatrix} -\eta b / 2 \\ 0 \end{pmatrix}.
\end{equation}

\subsubsection{Quadratic Loss}

Consider the quadratic loss $L(w) = \frac{1}{2}Sw^2$ with $S > 0$. For the gradient, we have that $\nabla L(w) = Sw$. 

We will first consider heavy ball momentum. For the center ODEs, we have that:
\begin{align*}
\frac{d\bar{w}}{dt} &= -\eta\bigl[\beta\, \bar{m} + (1-\beta)\, S\bar{w}\bigr], \\
\frac{d\bar{m}}{dt} &= (1-\beta)\bigl[S\bar{w} - \bar{m}\bigr].
\end{align*}
Unlike the linear-loss case, the dynamics of $\bar{m}$ are now coupled to those of $\bar{w}$. However, both $\bar{w}$ and $\bar{m}$ still evolve independently of $\Sigma$. And because the gradient is linear in $w$, they form a linear system:
\begin{equation}
\frac{d}{dt}\begin{pmatrix}\bar{w} \\ \bar{m}\end{pmatrix} = A\begin{pmatrix}\bar{w} \\ \bar{m}\end{pmatrix}, \qquad A = \begin{pmatrix} -\eta(1-\beta)S & -\eta\beta \\ (1-\beta)S & -(1-\beta) \end{pmatrix}\,.
\end{equation}
The eigenvalues of the transition matrix $A$ govern the dynamics of our system. The eigenvalues are the roots of the characteristic equation:
\begin{equation} \lambda^2 - \Tr(A)\lambda + \Det(A) = 0
\end{equation}
Evaluating these terms for $A$:
\begin{align*}
\Tr(A) &=  -(1-\beta)(1+\eta S), \\
\Det(A) &= (1-\beta)\eta S.
\end{align*}
To determine the stability of the system, we examine the real parts of the eigenvalues of the transition matrix $A$. The fixed point $(0,0)$ is asymptotically stable if and only if both eigenvalues satisfy $\operatorname{Re}(\lambda) < 0$.

Under the assumptions $\eta > 0$, $S > 0$, and $\beta \in [0,1)$, we observe that $\Tr(A) < 0$ and $\Det(A) > 0$. Since the trace is the sum of the eigenvalues and the determinant is their product, it follows that:
\begin{equation}
\lambda_1 + \lambda_2 < 0 \quad \text{and} \quad \lambda_1 \lambda_2 > 0\,.
\end{equation}
By the Routh--Hurwitz stability criterion, the system is stable and decays to zero. Hence the steady state is $(\bar{w}^\ast, \bar{m}^\ast) = (0, 0)$.

Now we will consider the dynamics for $\Sigma$. Let $(\bar{w}, \bar{m}) = (0, 0)$. Assume that $\Sigma$ is rank-one and positive semi-definite, so that it can be written as the outer product of $\Delta = (\delta, \gamma)$ with itself:
\begin{align*}
\Sigma &= \begin{pmatrix} \Sigma_{\delta\delta} & \Sigma_{\delta\gamma} \\ \Sigma_{\gamma\delta} & \Sigma_{\gamma\gamma} \end{pmatrix} \\
&= \begin{pmatrix} \delta \\ \gamma \end{pmatrix} \otimes \begin{pmatrix} \delta \\ \gamma \end{pmatrix} \\
&= \begin{pmatrix} \delta^2 & \delta\gamma \\ \gamma\delta & \gamma^2 \end{pmatrix}.
\end{align*}
For our phase-space update, we have:
\begin{align*}
\Phi_\pm &= \begin{pmatrix} \Phi_\pm^w \\ \Phi_\pm^m \end{pmatrix} \\
&= \pm \begin{pmatrix} -\eta\bigl[\beta\, \gamma + (1-\beta)\, S\delta\bigr] \\ (1-\beta)\bigl[S\delta - \gamma\bigr] \end{pmatrix} \\
&= \pm A \begin{pmatrix} \delta \\ \gamma \end{pmatrix},
\end{align*}
where $A$ is the transition matrix introduced above.

For the $\Sigma$ ODE:
\begin{align*}
\frac{d\Sigma}{dt} &= \tfrac{1}{4}\bigl(\Phi_+ \otimes \Phi_+ + \Phi_- \otimes \Phi_-\bigr) - 2\Sigma \\
&= \frac{1}{2}\,\biggl[A \begin{pmatrix} \delta \\ \gamma \end{pmatrix}\biggr] \otimes \biggl[A \begin{pmatrix} \delta \\ \gamma \end{pmatrix}\biggr] - 2\Sigma \\
&= \tfrac{1}{2}\, A\, \Sigma\, A^\top - 2\Sigma.
\end{align*}
Consider the fixed point $\Sigma^\ast = 0$. We want to determine for which values of the sharpness $S$ is the fixed point stable. The right-hand side of this ODE is linear in $\Sigma$, so the stability of $\Sigma^\ast = 0$ is controlled by the eigenvalues of the linear operator:
\begin{equation}
\mathcal{L}\colon \Sigma \;\longmapsto\; \tfrac{1}{2}\, A\, \Sigma\, A^\top - 2\Sigma,
\end{equation}
acting on the three-dimensional space of symmetric $2 \times 2$ matrices. The fixed point $\Sigma^\ast = 0$ is asymptotically stable if and only if all three eigenvalues of $\mathcal{L}$ have negative real parts.

Suppose that $A$ has eigenvalues $\alpha_1, \alpha_2$ with corresponding eigenvectors $v_1, v_2$. Acting on the rank-one matrix $v_i v_j^\top$, the operator $\Sigma \mapsto A\,\Sigma\, A^\top$ multiplies by $\alpha_i \alpha_j$:
\begin{equation*}
A\bigl(v_i v_j^\top\bigr) A^\top = (A v_i)(A v_j)^\top = \alpha_i \alpha_j\, v_i v_j^\top\,.
\end{equation*}
Restricting to symmetric matrices, the three eigendirections of $\mathcal{L}$ are:
\begin{align*}
v_1 v_1^\top \quad &\text{with eigenvalue} \quad \lambda_{11} = \tfrac{1}{2}\alpha_1^2 - 2\,, \\
v_2 v_2^\top \quad &\text{with eigenvalue} \quad \lambda_{22} = \tfrac{1}{2}\alpha_2^2 - 2\,, \\
v_1 v_2^\top + v_2 v_1^\top \quad &\text{with eigenvalue} \quad \lambda_{12} = \tfrac{1}{2}\alpha_1 \alpha_2 - 2\,.
\end{align*}
We start with the cross eigenvalue $\lambda_{12}$. Using $\alpha_1 \alpha_2 = \Det(A)$, we have that:
\begin{align*}
\lambda_{12} &= \tfrac{1}{2}\alpha_1 \alpha_2 - 2\\
&= \tfrac{1}{2} \Det(A) - 2 \\
&= \tfrac{1}{2}(1-\beta) \eta S - 2
\end{align*}
In order for $\lambda_{12}$ to be negative, we require that:
\begin{equation}
\frac{(1-\beta)}{2} \eta S - 2 < 0 \implies S < \frac{4}{\eta(1-\beta)}
\end{equation}
Anticipating our final result, it's helpful to notice that because $\beta \in [0,1)$, we have that:
\begin{equation}
S > \frac{4}{\eta(1-\beta)} \implies S > \frac{2}{\eta} \cdot\frac{1+\beta}{1-\beta}
\end{equation}
Let $\lambda_{ii}$ generically denote either of the two diagonal eigenvalues of $\mathcal{L}$. For $\lambda_{ii}$ to have a negative real part, we require:
\begin{equation}
\operatorname{Re}\{\lambda_{ii}\} = \tfrac{1}{2}\operatorname{Re}\{\alpha_i^2\} - 2 < 0 \;\;\implies\;\; \operatorname{Re}\{\alpha_i^2\} < 4\,.
\end{equation}
Recall from the analysis of the centers that $\Tr(A) < 0$ and $\Det(A) > 0$---from which it followed that the eigenvalues of $A$ are either a complex-conjugate pair or both real and negative. We can handle the two cases separately.

Consider the case that $\alpha_{1}$ and $\alpha_2$ form a complex-conjugate pair. For any complex number $\alpha$, we have $\operatorname{Re}\{\alpha^2\} \leq |\alpha|^2$, with equality only if $\alpha$ is real. Combined with $|\alpha_1|^2 = |\alpha_2|^2 = \Det(A)$, this gives $\operatorname{Re}\{\alpha_i^2\} \leq \Det(A)$. One can then see that:
\begin{align*}
\text{Re}\{\lambda_{ii} \} &= \tfrac{1}{2}\operatorname{Re}\{\alpha_i^2\} - 2 \\
&\le \tfrac{1}{2} |\alpha_i|^2 - 2 \\
&= \tfrac{1}{2} \alpha_1\alpha_2 - 2 \\
&= \lambda_{12}
\end{align*}
In that case, $\lambda_{12} < 0$ would imply that $\text{Re} \{\lambda_{ii}\} < 0$.

Now consider the case that both $\alpha_i$ are real and negative. We have that $\operatorname{Re}\{\alpha_i^2\} = \alpha_i^2$, so our stability condition becomes $|\alpha_i| < 2$. Since $\alpha_i < 0$, the stability threshold corresponds to when the more negative eigenvalue of $A$ reaches $-2$. Substituting $\lambda = -2$ into the characteristic equation gives:
\begin{equation*}
4 + 2\Tr(A) + \Det(A) = 0\,.
\end{equation*}
Plugging in the expressions for $\Tr(A)$ and $\Det(A)$ and simplifying:
\begin{equation*}
2 + 2\beta - \eta(1-\beta)\, S^\ast = 0\,.
\end{equation*}
Solving for $S^\ast$ yields the heavy ball edge-of-stability threshold:
\begin{equation}
S^\ast = \frac{2}{\eta} \cdot \frac{1+\beta}{1-\beta}\,.
\end{equation}
Now consider Nesterov momentum. For the quadratic loss $L(w) = \tfrac{1}{2}Sw^2$, we have $\nabla L(\theta_t) = S(w_t - \eta\beta\, m_t)$. As was the case with heavy ball momentum, the phase-space update is linear in $z$:
\begin{equation*}
\Phi(z) = A\, z\,, \qquad A = \begin{pmatrix} -\eta(1-\beta)S & -\eta\beta\bigl[1 - \eta(1-\beta)S\bigr] \\ (1-\beta)S & -(1-\beta)\bigl[1 + \eta\beta S\bigr] \end{pmatrix}\,.
\end{equation*}
For the trace and determinant, we have that:
\begin{align*}
\Tr(A) &= -(1-\beta)\bigl[1 + \eta(1+\beta)S\bigr]\,, \\
\Det(A) &= \eta(1-\beta)S\,.
\end{align*}
Once again, we have that $\Tr(A) < 0$ and $\Det(A) > 0$. So the fixed point $(\bar{w}^\ast, \bar{m}^\ast) = (0, 0)$ is stable. 

Though the transition matrix $A$ differs between heavy ball and Nesterov momentum, the $\Sigma$ ODE retains the same functional form:
\begin{equation*}
\frac{d\Sigma}{dt} = \tfrac{1}{2}\, A\, \Sigma\, A^\top - 2\Sigma\,.
\end{equation*}
To find the sharpness threshold, we again substitute $\lambda = -2$ into the characteristic equation:
\begin{equation}
2(1+\beta) - \eta(1-\beta)(1+2\beta)\, S^\ast= 0\,.
\end{equation}
Solving for $S^\ast$ yields the sharpness threshold for Nesterov momentum:
\begin{equation}
S^\ast = \frac{2}{\eta}\cdot\frac{1+\beta}{(1-\beta)(1+2\beta)}\,.
\end{equation}

\subsubsection{Quartic Loss}

Consider the quartic potential
\begin{equation}
L(w) = \frac{S}{2}w^2 - \frac{Q}{4}w^4,
\end{equation}
where $S, Q > 0$. The gradient is $\nabla L(w) = Sw - Qw^3$. As was the case with gradient descent, this potential is intended as a minimal model of a loss landscape that is sharpest near the origin and flattens out farther away. We will see that the momentum algorithms settle into finite-amplitude oscillations precisely when the sharpness at the minimum exceeds the respective sharpness thresholds derived in the quadratic analysis.

Consider heavy ball momentum. Because the loss function is even ($L(-w) = L(w)$), the gradient is odd ($\nabla L(-w) = -\nabla L(w)$). This means that at the origin $(\bar{w}^\ast, \bar{m}^\ast) = (0, 0)$, we have:
\begin{equation}
\bar{g} = \tfrac{1}{2} (\nabla L(\delta) + \nabla L(-\delta)) = 0
\end{equation}
from which it follows that:
\begin{equation}
\frac{d\bar{w}}{dt} = -\eta [\beta \bar{m} + (1-\beta)\bar{g}] = 0
\end{equation}
and:
\begin{equation}
\frac{d \bar{m}}{dt} = (1-\beta)(\bar{g} - \bar{m}) = 0.
\end{equation}
Thus $(\bar{w}, \bar{m}) = (0,0)$ is a fixed point of the center ODEs. 

We restrict our analysis to this case throughout. Note that because:
$$\frac{d\bar{z}}{dt} = \tfrac{1}{2}(\Phi_+ + \Phi_-) = 0$$
It immediately follows that $\Phi_- = - \Phi_+$.
For our $\Sigma$ ODE, we then have that: 
\begin{equation}
\frac{d\Sigma}{dt} = \tfrac{1}{2}\,\Phi_+\otimes \Phi_+ - 2\Sigma\,.
\label{eq:hb_quartic_sigma}
\end{equation}
We want to analyze the fixed points of the $\Sigma$ ODE. Setting $\frac{d \Sigma}{dt} = 0$, we have that fixed points $\Sigma^\ast$ satisfy:
\begin{equation}
\Sigma^\ast = \tfrac{1}{4} \Phi_+ (\Delta^\ast) \otimes \Phi_+(\Delta^\ast) \label{eq:sigma_fixed_point}
\end{equation}
In terms of the components of $\Sigma^\ast$, we have that:
\begin{align}
\Sigma_{\delta \delta}^\ast &= \tfrac{1}{4}\eta^2 [\beta \gamma^\ast + (1-\beta) g(\delta^\ast)]^2 \\
\Sigma^\ast_{\gamma \gamma} &= \tfrac{1}{4}(1 -\beta)^2[g(\delta^\ast) - \gamma^\ast]^2 \\
\Sigma^\ast_{\delta \gamma}&= - \tfrac{1}{4}\eta (1-\beta)[\beta \gamma^\ast + (1-\beta) g(\delta^\ast)][g(\delta^\ast) - \gamma^\ast] 
\end{align}
where $g(\delta) = S \delta - Q \delta^3$.

The right-hand side of \cref{eq:sigma_fixed_point} is an outer product, so any non-trivial fixed point $\Sigma^\ast$ must be rank-one. Writing $\Sigma^\ast = \Delta^\ast \otimes \Delta^\ast$, it then follows that:
\begin{equation}
\Phi_+(\Delta^\ast) = \pm 2\,\Delta^\ast\,.
\end{equation}
For the quiescent fixed point $\Delta^\ast = 0$, both sign choices are satisfied trivially. For non-trivial fixed points, we want the negative solution, since the minus sign encodes the period-2 condition $z_{t+1} = -z_t$ of the underlying discrete heavy ball map:
\begin{equation}
z_{t+1} = z_t + \Phi(z_t) \quad \text{and} \quad z_{t+1} = -z_t \quad \implies \quad \Phi(z_t) = -2 z_t\,.
\end{equation}
We therefore take:
\begin{equation}
\Phi_+(\Delta^\ast) = -2\,\Delta^\ast\,.
\label{eq:hb_quartic_period2}
\end{equation}
We can express \cref{eq:hb_quartic_period2} as two scalar equations:
\begin{align}
-2\delta^\ast &= -\eta\bigl[\beta\gamma^\ast + (1-\beta)g(\delta^\ast)\bigr]\,, \\
-2\gamma^\ast &= (1-\beta)\bigl[g(\delta^\ast) - \gamma^\ast\bigr]\,.
\end{align}
The second equation can be solved for $\gamma^\ast$ in terms of $\delta^\ast$:
\begin{equation}
\gamma^\ast = -\frac{1-\beta}{1+\beta}\,g(\delta^\ast)\,.
\label{eq:gamma_equation}
\end{equation}
Substituting this into the first equation, we have that:
\begin{align*}
- 2 \delta^\ast &= -\eta \left[\beta \left ( - \frac{1-\beta}{1+\beta} g(\delta^\ast)\right) + (1-\beta) g(\delta^\ast) \right] \\
&= - \eta g(\delta^\ast) \left [ \frac{-\beta(1-\beta) + (1+\beta)(1-\beta)}{1+\beta} \right] \\
&= -\eta g(\delta^\ast) \left[\frac{1-\beta}{1+\beta} \right]
\end{align*}
Rearranging, we can obtain the following equation:
\begin{equation}
\frac{g(\delta^\ast)}{\delta^\ast} = \frac{2(1+\beta)}{\eta(1-\beta)} = S^\ast.
\label{eq:average_curvature_hb}
\end{equation}
The left-hand side of \cref{eq:average_curvature_hb} is the gradient divided by the distance from the center of the oscillations. Because the change in gradient between the rod endpoints is $2g(\delta)$ while the displacement is $2\delta$, the left-hand side gives the \textit{average} curvature along the line segment connecting the two endpoints of the rod. The right-hand side is the sharpness threshold for heavy ball momentum. This generalizes the result from the parabola: stable two-point cycles correspond to orbits whose average curvature over the rod equals the sharpness threshold. What is special about the parabola is that, because the curvature is constant, there is an entire manifold of two-point orbits. For generic loss functions, only a special amplitude satisfies the average-curvature condition. 

Using the form of the gradient for our quartic loss, we have:
\begin{equation}
\frac{g(\delta^\ast)}{\delta^\ast} = S - Q(\delta^\ast)^2\,.
\end{equation}
Recalling that $\Sigma^\ast_{\delta\delta} = (\delta^\ast)^2$, the average-curvature condition gives:
\begin{equation}
S - Q(\delta^\ast)^2 = S^\ast \quad \implies \quad \Sigma^\ast_{\delta\delta} = \frac{S - S^\ast}{Q}\,.
\end{equation}
We require $\Sigma^\ast_{\delta\delta} > 0$ for the fixed point to be physically meaningful. Since $Q > 0$, we have:
\begin{equation}
\Sigma^\ast_{\delta\delta} > 0 \iff S - S^\ast > 0\,.
\end{equation}
So, the oscillation fixed point becomes physically meaningful precisely when the sharpness at the minimum exceeds the sharpness threshold.

For $\gamma^\ast$, substituting $g(\delta^\ast) = S^\ast \delta^\ast$ into \cref{eq:gamma_equation} gives:
\begin{align*}
\gamma^\ast &= -\frac{1-\beta}{1+\beta}\, g(\delta^\ast) \\
&= -\frac{1-\beta}{1+\beta} \left(\frac{2(1+\beta)}{\eta(1-\beta)}\, \delta^\ast \right) \\
&= -\frac{2}{\eta}\, \delta^\ast\,,
\end{align*}
from which it follows:
\begin{equation}
\frac{\gamma^\ast}{\delta^\ast} = -\frac{2}{\eta}\,.
\end{equation}
Recalling that $\Sigma^\ast_{\gamma\gamma} = (\gamma^\ast)^2$ and $\Sigma^\ast_{\delta\gamma} = \delta^\ast \gamma^\ast$, we then have that:
\begin{equation}
\Sigma^\ast_{\gamma\gamma} = \frac{4(S - S^\ast)}{\eta^2 Q}\,, \qquad \Sigma^\ast_{\delta\gamma} = -\frac{2(S - S^\ast)}{\eta\, Q}\,.
\end{equation}
Let $\Sigma^\ast$ denote the non-zero fixed point. We want to determine when $\Sigma^\ast$ is a stable fixed point. Let $f(\Sigma)$ denote the right-hand side of the ODE for $\Sigma$:
\[
 f(\Sigma) = \frac{1}{2}\,\Phi_+(\Sigma) \otimes \Phi_+(\Sigma) - 2\Sigma.
\]
A fixed point $\Sigma^\ast$ is asymptotically stable if all eigenvalues of the linear operator $\nabla_\Sigma f(\Sigma^\ast)$ have strictly negative real parts. Let $\widetilde{\Sigma}$ be a symmetric perturbation matrix. The action of $\nabla_\Sigma f(\Sigma^\ast)$ on $\widetilde{\Sigma}$ is given by:
\[
\nabla_\Sigma f(\Sigma^\ast)[\widetilde{\Sigma}] = \frac{1}{2}\Bigl( \nabla_\Sigma \Phi_+[\widetilde{\Sigma}] \otimes \Phi_+ + \Phi_+ \otimes \nabla_\Sigma \Phi_+[\widetilde{\Sigma}] \Bigr) - 2\widetilde{\Sigma}.
\]

We now need an expression for $\nabla_\Sigma \Phi_+$. Since $\Phi_+$ depends on $\Sigma$ entirely through its principal eigencomponent $\Delta$, we can apply the chain rule. The derivative of $\Phi_+$ with respect to $\Sigma$ decomposes into the product of the Jacobian of $\Phi_+$ with respect to the principal vector $\Delta$ and the derivative of $\Delta$ with respect to $\Sigma$:
\[
\nabla_\Sigma \Phi_+ = \frac{d \Phi_+}{d \Delta}\,\frac{d \Delta}{d\Sigma}.
\]
We begin by computing the Jacobian of $\Phi_+$ with respect to $\Delta$. Recall that:
\[
\Phi_+ = \begin{pmatrix} \Phi_+^w \\ \Phi_+^m \end{pmatrix} =
\begin{pmatrix} -\eta\,[\beta \gamma + (1-\beta)\,g(\delta)] \\ (1-\beta)\,[g(\delta) - \gamma] \end{pmatrix}.
\]
Let $J = d\Phi_+/d\Delta$. Then we have that:
\begin{equation}
J = \begin{pmatrix} \frac{d\Phi_+^w}{d\delta} & \frac{d\Phi_+^w}{d\gamma} \\
\frac{d\Phi_+^m}{d\delta} & \frac{d\Phi_+^m}{d\gamma} \\
\end{pmatrix}
=\begin{pmatrix} -\eta(1-\beta)H(\delta) & -\eta\beta \\ (1-\beta)H(\delta) & -(1-\beta) \end{pmatrix},
\end{equation}
where $H$ is the Hessian of our loss function:
\begin{equation}
H(\delta) = g'(\delta) = S - 3Q\delta^2.
\end{equation}
We now need to find $d\Delta/d\Sigma$. Although $\Sigma$ is a $2 \times 2$ matrix with four independent entries, we can restrict our attention to the three-dimensional space of symmetric $2 \times 2$ matrices. When considering perturbations $\widetilde{\Sigma}$, it helps to work in a basis aligned with the rod of the fixed point. Define:
\begin{equation} 
e_1 = \frac{\Delta^\ast}{\|\Delta^\ast\|} \qquad \text{and} \qquad e_2 \perp e_1.
\end{equation}
We then have that:
\[
\widetilde{\Sigma} \in \operatorname{span}\left\{ e_1 \otimes e_1,\; e_2 \otimes e_2,\; \tfrac{1}{2}(e_1 \otimes e_2 + e_2 \otimes e_1) \right\}.
\]
Recall that a \textit{cone} is a set that is closed under scalar multiplication. Because $\Sigma^\ast = \Delta^\ast \otimes \Delta^\ast$, it lies on the cone of rank-1 symmetric $2 \times 2$ matrices. The manifold of symmetric $2 \times 2$ matrices is three-dimensional, whereas the cone of rank-1 symmetric matrices is two-dimensional. We can therefore distinguish two types of perturbations. \textit{Off-cone} perturbations $\widetilde{\Sigma}_\perp$ increase the rank. \textit{On-cone} perturbations $\widetilde{\Sigma}_\parallel$ rotate and scale the principal eigenvector but remain on the cone of rank-1 symmetric matrices.

Consider an off-cone perturbation: 
\begin{equation}    
\widetilde{\Sigma}_\perp = c\, e_2 \otimes e_2
\end{equation}
for small $c > 0$. We then have that:
 \begin{equation}
\Sigma^\ast + \widetilde{\Sigma}_\perp = \Delta^\ast \otimes \Delta^\ast + c \,e_2 \otimes e_2
 \end{equation}
 Because $e_2$ is orthogonal to $e_1$ by construction, the eigenvectors of $\Sigma^\ast + \widetilde{\Sigma}_\perp$ remain $e_1$ and $e_2$. And because $c$ is assumed small, we have that $\|\Delta\|^2 > c$, so the principal eigenvalue remains $\|\Delta\|^2$. As the principal eigenvector ($e_1$) and principal eigenvalue ($\|\Delta^\ast\|^2$) remain unchanged under off-cone perturbations, it follows that:
 \begin{equation}
\frac{d\Delta}{d\Sigma}[\widetilde{\Sigma}_\perp] = 0
 \end{equation}
We can then show that for off-cone perturbations:
\begin{align*}
\nabla_\Sigma f(\Sigma^\ast)[\widetilde{\Sigma}_\perp] &= \frac{1}{2}\Bigl( \nabla_\Sigma \Phi_+[\widetilde{\Sigma}_\perp] \otimes \Phi_+ + \Phi_+ \otimes \nabla_\Sigma \Phi_+[\widetilde{\Sigma}_\perp] \Bigr) - 2\widetilde{\Sigma}_\perp\\
&=\frac{1}{2} \Bigl (\frac{d \Phi_+ }{d \Delta} \frac{d\Delta}{d\Sigma}[\widetilde{\Sigma}_\perp] \otimes \Phi_+ + \Phi_+ \otimes \frac{d \Phi_+ }{d \Delta} \frac{d\Delta}{d\Sigma}  [\widetilde{\Sigma}_\perp] \Bigr) - 2\widetilde{\Sigma}_\perp \\
&= -2 \widetilde{\Sigma}_\perp
\end{align*}
Therefore, off-cone perturbations are eigenvectors of $\nabla_\Sigma f$ with eigenvalue $-2$.

Now consider an on-cone perturbation: \begin{equation}
\widetilde{\Sigma}_\parallel = u \otimes \Delta^\ast + \Delta^\ast \otimes u, \qquad u = a e_1 + b e_2 
\end{equation}
One can see that:
\begin{align*}
\Sigma^\ast + \widetilde{\Sigma}_\parallel &= \Delta^\ast \otimes \Delta^\ast + u \otimes \Delta^\ast + \Delta^\ast \otimes u \\
&= (\Delta^\ast + u) \otimes (\Delta^\ast + u) - u \otimes u \\
&= (\Delta^\ast + u) \otimes (\Delta^\ast + u) + O(u^2)
\end{align*}
To first order, $\widetilde{\Sigma}_\parallel$ shifts the principal eigenvector by $u$ (assumed small). We therefore deduce that:
\begin{equation}
\frac{d\Delta}{d\Sigma}\bigl[\widetilde{\Sigma}_\parallel\bigr] = u.
\end{equation}
Recall that $\Phi_+ = - 2 \Delta^\ast$. For on-cone perturbations,
\begin{align*}
\nabla_\Sigma f(\Sigma^\ast)[\widetilde{\Sigma}_\parallel]
&= \tfrac{1}{2}\Bigl( \nabla_\Sigma \Phi_+[\widetilde{\Sigma}_\parallel] \otimes \Phi_+ + \Phi_+ \otimes \nabla_\Sigma \Phi_+[\widetilde{\Sigma}_\parallel] \Bigr) - 2\widetilde{\Sigma}_\parallel \\
&= \tfrac{1}{2} \Bigl( \tfrac{d \Phi_+}{d \Delta}\, \tfrac{d\Delta}{d\Sigma}[\widetilde{\Sigma}_\parallel] \otimes \Phi_+ + \Phi_+ \otimes \tfrac{d \Phi_+}{d \Delta}\, \tfrac{d\Delta}{d\Sigma}[\widetilde{\Sigma}_\parallel] \Bigr) - 2\widetilde{\Sigma}_\parallel \\
&= \tfrac{1}{2} \Bigl( Ju \otimes (-2\Delta^\ast) + (-2 \Delta^\ast) \otimes Ju \Bigr) - 2 \widetilde{\Sigma}_\parallel \\
&= -\bigl( Ju \otimes \Delta^{\ast} + \Delta^\ast \otimes Ju \bigr) - 2 \widetilde{\Sigma}_\parallel \\
&= -\bigl[ (J + 2I)u \otimes \Delta^\ast + \Delta^\ast \otimes (J + 2I)u \bigr].
\end{align*}

Factoring out the basis structure yields a 2-dimensional ODE for the perturbation vector $u$:
\begin{equation}
\frac{du}{dt} = -(J + 2I)u.
\end{equation}
In order for the fixed point $\Sigma^\ast$ to be stable, the real part of both eigenvalues of $-(J + 2I)$ must be negative. Let $\lambda$ denote an eigenvalue of $J$. We have that $\lambda$ satisfies the characteristic polynomial:
\[
\lambda^2 - \Tr(J)\lambda + \Det(J) = 0.
\]
Let $\nu$ denote an eigenvalue of $-(J+2I)$. If $v$ is an eigenvector of $J$, we then have that: 
\begin{equation}
-(J+2I) v = -(\lambda + 2) v \implies \nu = - (\lambda +2)
\end{equation}
Substituting $\lambda = -\nu - 2$ into the characteristic polynomial of $J$, yields the characteristic polynomial of $-(J+2I)$:
\begin{equation}
(-\nu - 2)^2 - \Tr(J)(-\nu - 2) + \Det(J) = 0
\end{equation}
Using $\Tr(J) = -(1-\beta)(1 + \eta H(\delta^\ast))$ and $\Det(J) = \eta(1-\beta)H(\delta^\ast)$, we have that:
\[
\nu^2 + \bigl[3 + \beta - \eta(1-\beta)H(\delta^\ast)\bigr]\,\nu + \bigl[2(1+\beta) - \eta(1-\beta)H (\delta^\ast)\bigr] = 0.
\]
The characteristic polynomial above is expressed in terms of the Hessian $H$ at the fixed point. We want to rewrite it in terms of the sharpness threshold $S^\ast$ and the sharpness $S$ at the minimum.

First, one can show that:
\begin{align*}
H &= S - 3Q \cdot \left (\frac{S - S^\ast}{Q} \right) \\
&= S - 3(S - S^\ast) \\
&= 3S^\ast - 2S.
\end{align*}
One can also see that:
\begin{equation}
S^\ast = \frac{2}{\eta} \cdot \frac{1+\beta}{1-\beta} \quad \Longleftrightarrow \quad \eta(1-\beta)S^\ast = 2(1+\beta).
\end{equation}
Combining the two above equations, one can see that:
\begin{align}
\eta(1-\beta)H &= 3 \eta (1-\beta)S^\ast - 2 \eta(1-\beta)S \nonumber \\
&= 6(1+\beta) - 2\eta(1-\beta)S \label{eq:Hidentity} 
\end{align}
Substituting \eqref{eq:Hidentity} into the linear coefficient:
\begin{align*}
3 + \beta - \eta(1-\beta)H 
&= 3 + \beta - \bigl[6(1+\beta) - 2\eta(1-\beta)S\bigr] \\
&= 2\eta(1-\beta)S - 3 - 5\beta.
\end{align*}
Substituting \eqref{eq:Hidentity} into the constant term:
\begin{align*}
2(1+\beta) - \eta(1-\beta)H 
&= 2(1+\beta) - \bigl[6(1+\beta) - 2\eta(1-\beta)S\bigr] \\
&= 2\eta(1-\beta)S - 4(1+\beta) \\
&= 2\eta (1-\beta)S - 2 \eta(1-\beta)S^\ast\\
&= 2\eta(1-\beta)(S- S^\ast)
\end{align*}
The characteristic polynomial therefore simplifies to:
\[
\nu^2 + \bigl[2\eta(1-\beta)S - 3 - 5\beta\bigr]\,\nu + 2\eta(1-\beta)(S - S^\ast) = 0.
\]
By the Routh--Hurwitz criterion, both eigenvalues have negative real parts if and only if both coefficients are strictly positive:
\begin{itemize}
    \item \textbf{Constant term:} $2\eta(1-\beta)(S - S^\ast) > 0 \iff S > S^\ast$.
    \item \textbf{$\nu$-coefficient:} $2\eta(1-\beta)S - 3 - 5\beta > 0 \iff S > \dfrac{3 + 5\beta}{2\eta(1-\beta)}$.
\end{itemize}
The second condition is automatically implied by the first. Since $S^\ast = \dfrac{4(1+\beta)}{2\eta(1-\beta)}$, comparing the two thresholds:
\[
\frac{3 + 5\beta}{2\eta(1-\beta)} < \frac{4 + 4\beta}{2\eta(1-\beta)} \iff 3 + 5\beta < 4 + 4\beta \iff \beta < 1,
\]
which holds since the momentum parameter satisfies $\beta < 1$. Hence the $\nu$-coefficient threshold is strictly weaker than $S^\ast$, and both on-cone eigenvalues have negative real parts if and only if $S > S^\ast$.

Combined with the off-cone eigenvalue of $-2$, we conclude that the fixed point $\Sigma^\ast$ is stable if and only if $S > S^\ast$.

\section{Rod Flow for RMSProp}
\label{app:rmsprop}

\subsection{Derivation of Rod Flows}
\subsubsection{Scalar RMSProp}

For the Scalar RMSProp update equations, we have:
\begin{align*} 
\nu_{t+1} &= \beta_2 \nu_t + (1-\beta_2) \|\nabla L(w_t)\|^2 \\
w_{t+1}&= w_t - \eta \,P_{t+1}^{-1} \,\nabla L(w_t) 
\end{align*}
where $P$ is the preconditioner:
$$P(\nu) = (\sqrt{\nu} + \varepsilon)I_d$$
As was the case with gradient descent, we can define the midpoint and half-displacement of the position:
\begin{align*}
\bar{w}_t &= \frac{w_{t+1} + w_t}{2} \\
\delta_t &= \frac{w_{t+1} - w_t}{2}
\end{align*}
With Scalar RMSProp, we will only track the midpoint $\bar{\nu}$ of the second moment. There is no need to track the half-difference of $\nu$ because, even at the edge of stability, there are no oscillations in the gradient norm.
$$\bar{\nu}_t = \frac{\nu_{t+1} + \nu_t}{2}$$
Due to not tracking the half-difference in $\nu$, the difference equations are not exact as they were for gradient descent. However, the discrepancy should be negligible as long as the norm of the gradient is smoothly varying---which is a necessary precondition for rod flow to work as an effective theory in the first place.

For the difference equation of the midpoint, we have the following:
\begin{align*}
\bar{w}_{t+1} - \bar{w}_t &= \frac{w_{t+2} + w_{t+1}}{2} - \frac{w_{t+1} + w_t}{2}\\
&= \frac{w_{t+2} - w_{t+1}}{2} + \frac{w_{t+1} - w_t}{2} \\
&= - \frac{\eta}{2 (\sqrt{\nu_{t+2}}+ \varepsilon)}\nabla L(w_{t+1}) - \frac{\eta}{2 (\sqrt{\nu_{t+1}} + \varepsilon)} \nabla L(w_t) \\
&\approx -\frac{\eta}{2(\sqrt{\bar{\nu}_t} + \varepsilon)}\,\bigl (\nabla L_+ + \nabla L_- \bigr)
\end{align*}
And for the difference equation of the extent:
\begin{align*}
\delta_{t+1}\otimes \delta_{t+1} - \delta_t \otimes \delta_t &= (\delta_{t+1} \otimes \delta_{t+1} + \delta_t \otimes \delta_t) - 2 \delta_t \otimes \delta_t \\
&= \left(\frac{\eta}{2 (\sqrt{\nu_{t+2}}+ \varepsilon)}\right)^2\nabla L(w_{t+1}) \otimes \nabla L(w_{t+1}) \\
&\qquad{} + \left (\frac{\eta}{2 (\sqrt{\nu_{t+1}}+ \varepsilon)} \right)^2\nabla L(w_t) \otimes \nabla L(w_t) - 2\delta_t \otimes \delta_t \\
&\approx \left (\frac{\eta}{2 (\sqrt{\bar{\nu}_t} + \varepsilon)} \right)^2\, \bigl (\nabla L_+ \otimes \nabla L_+ + \nabla L_- \otimes \nabla L_- \bigr) - 2\delta_t \otimes \delta_t
\end{align*}
For the difference equation of the second moment midpoint:
\begin{align*}
\bar{\nu}_{t+1} - \bar{\nu}_t &= \frac{\nu_{t+2} + \nu_{t+1}}{2} - \frac{\nu_{t+1} + \nu_t}{2} \\
&= \frac{\nu_{t+2} - \nu_{t+1}}{2} + \frac{\nu_{t+1} - \nu_t}{2} \\
&=\frac{(1-\beta_2) [ \|\nabla L(w_{t+1})\|^2 - \nu_{t+1}]}{2}  + \frac{(1-\beta_2) [ \|\nabla L(w_{t})\|^2 - \nu_t]}{2} \\
&= (1-\beta_2)\left[\frac{\|\nabla L_+\|^2 + \|\nabla L_-\|^2}{2} - \bar{\nu}_t\right]
\end{align*}
We can then promote the above difference equations to obtain the rod flow for Scalar RMSProp:
\begin{align*}
\frac{d\wbar}{dt} &= -\eta \, P^{-1}(\bar{\nu})\,\bar{g}, \\
\frac{d \Sigma}{dt} &= \left (\frac{\eta}{2} P^{-1}(\bar{\nu}) \right )^2 \, \bigl(\nabla L_+ \otimes \nabla L_+ + \nabla L_- \otimes \nabla L_- \bigr) - 2 \Sigma \\
\frac{d\nubar}{dt} &= (1-\beta_2)\left[\frac{\|\nabla L_+\|^2 + \|\nabla L_-\|^2}{2} - \nubar\right].
\end{align*}
Note that if we define the preconditioned step size
\begin{equation}
\tilde{\eta} = \eta \, P^{-1}(\bar{\nu})
\end{equation}
then the rod flow ODEs for $\bar{w}$ and $\Sigma$ are identical to those for gradient descent, except that $\tilde{\eta}$ replaces $\eta$.

\subsubsection{RMSProp}

For standard RMSProp, the second moment estimate $\nu \in \mathbb{R}^d$ is promoted to a vector of the same dimension as $w$, tracking per-component second-moment estimates. The update equations are:
\begin{align*} 
\nu_{t+1} &= \beta_2 \nu_t + (1-\beta_2) \nabla L(w_t)^{\odot 2} \\
w_{t+1}&= w_t - \eta \, P^{-1}_{t+1} \,\nabla L(w_t) 
\end{align*}
The preconditioner $P$ is given as:
$$P(\nu) = \text{diag}(\sqrt{\nu}) + \varepsilon I_d$$
The midpoints $\bar{w}_t$, $\bar{\nu}_t$, and the half-displacement $\delta_t$ are defined exactly as they were in Scalar RMSProp. For the difference equation of the $\nu$ midpoint, we replace the squared norm from the scalar case with the component-wise square:
\begin{align*}
\bar{\nu}_{t+1} - \bar{\nu}_t &= \frac{\nu_{t+2} + \nu_{t+1}}{2} - \frac{\nu_{t+1} + \nu_t}{2} \\
&= \frac{\nu_{t+2} - \nu_{t+1}}{2} + \frac{\nu_{t+1} - \nu_t}{2} \\
&=\frac{(1-\beta_2) [ \nabla L(w_{t+1})^{\odot 2} - \nu_{t+1}]}{2}  + \frac{(1-\beta_2) [ \nabla L(w_{t})^{\odot 2} - \nu_t]}{2} \\
&= (1-\beta_2)\left[\frac{\nabla L_+^{\odot 2} + \nabla L_-^{\odot 2}}{2} - \bar{\nu}_t\right]
\end{align*}
Promoting these difference equations to continuous time, we obtain the rod flow for standard RMSProp:
\begin{align*}
\frac{d\wbar}{dt} &= -\eta \, P^{-1} (\bar{\nu}) \, \bar{g}, \\
\frac{d \Sigma}{dt} &= \left (\frac{\eta}{2} P^{-1}(\bar{\nu}) \right )^2 \bigg(\nabla L_+ \otimes \nabla L_+ + \nabla L_- \otimes \nabla L_- \bigg)  - 2 \Sigma \\
\frac{d\nubar}{dt} &= (1-\beta_2)\left[\frac{\nabla L_+^{\odot 2} + \nabla L_-^{\odot 2}}{2} - \nubar\right].
\end{align*}

\subsection{Theoretical Analysis}
\subsubsection{Linear Loss}

Consider the linear loss $L(w) = -b \cdot w$. Because we are working in one-dimension, there is no difference between Scalar RMSProp and standard RMSProp. For the theoretical analysis section, we will drop the regularization parameter $\varepsilon$.

Since $\nabla L_+ = \nabla L_- = -b$, the rod flow equations become:
\begin{align}
    \frac{d\bar{\nu}}{dt} &= (1-\beta_2)(b^2 - \bar{\nu})\,, \\
    \frac{d\bar{w}}{dt} &=  \frac{\eta b}{\sqrt{\bar{\nu}} }\,, \\
    \frac{d\Sigma}{dt} &= \frac{\eta^2 b^2}{2\bar{\nu}} - 2\Sigma\,.
\end{align}
Setting $\frac{d\bar{\nu}}{dt} = 0$ gives the steady-state second moment:
\begin{equation}
    \bar{\nu}^* = b^2\,.
\end{equation}
Substituting our expression for $\bar{\nu}^*$ into the $\bar{w}$ equation yields the steady-state equation of motion for the midpoint of the position:
\begin{equation}
    \frac{d\bar{w}}{dt}  = \eta \operatorname{sign}(b)\,.
\end{equation}
Setting $\frac{d\Sigma}{dt} = 0$ gives the steady-state extent:
\begin{equation}
    \Sigma^* =  \frac{\eta^2}{4}\,.
\end{equation}
In a flat landscape, RMSProp behaves like Sign-GD: it takes a step down the landscape with fixed step size $\eta$.

\subsubsection{Quadratic Loss}

Consider the quadratic $L(w) = \frac{1}{2}Sw^2$ with $S > 0$. In one dimension, the extent $\Sigma$ is a scalar. So the half-displacement is $\delta = \sqrt{\Sigma}$. The gradient is $\nabla L(w) = Sw$.

Evaluating the gradients at the endpoints of the rod yields:
\begin{align}
    \nabla L_+ &= S(\bar{w} + \delta) \\
    \nabla L_- &= S(\bar{w} - \delta)
\end{align}
Summing these terms and summing their squares gives:
\begin{align}
    \nabla L_+ + \nabla L_- &= 2S\bar{w} \\
    \nabla L_+^2 + \nabla L_-^2 &= S^2(\bar{w} + \delta)^2 + S^2(\bar{w} - \delta)^2 = 2S^2(\bar{w}^2 + \Sigma)
\end{align}
Plugging these into our rod flow equations for RMSProp:
\begin{align}
    \frac{d\bar{w}}{dt} &= -\frac{\eta S}{\sqrt{\bar{\nu}}}\bar{w} \\
    \frac{d\Sigma}{dt} &= \left(\frac{\eta^2 S^2}{2\bar{\nu}}\right)(\bar{w}^2 + \Sigma) - 2\Sigma \\
    \frac{d\bar{\nu}}{dt} &= (1-\beta_2)\left(S^2(\bar{w}^2 + \Sigma) - \bar{\nu}\right)
\end{align}
By inspection, the center $\bar{w}$ experiences a restoring force and decays to the origin. Because of this, it suffices to set $\bar{w}^\ast = 0$. We can then study the coupled dynamics of the extent and the second moment at steady-state:
\begin{align}
    \frac{d\Sigma}{dt} &= \left(\frac{\eta^2 S^2}{2\bar{\nu}} - 2\right)\Sigma \\
    \frac{d\bar{\nu}}{dt} &= (1-\beta_2)(S^2\Sigma - \bar{\nu})
\end{align}
We want to find the fixed point $(\Sigma^*, \bar{\nu}^*)$. From the second moment equation, we have that: 
\begin{equation}
    \bar{\nu}^* = S^2\Sigma^*
\end{equation}
Substituting this relationship into the extent equation and setting $\frac{d\Sigma}{dt} = 0$ yields:
\begin{equation}
    \frac{\eta^2 S^2}{2(S^2\Sigma^*)} - 2 = 0 \implies \frac{\eta^2}{2\Sigma^*} = 2
\end{equation}
Solving for $\Sigma^*$ yields:
\begin{equation}
    \Sigma^* = \frac{\eta^2}{4}
\end{equation}
To determine the stability of the fixed point $(\Sigma^*, \bar{\nu}^*) = \left(\frac{\eta^2}{4}, \frac{S^2 \eta^2}{4}\right)$, we perform a linear stability analysis by evaluating the Jacobian matrix of the system at this point.

Let our system of ODEs be defined as:
\begin{align*}
    f(\Sigma, \bar{\nu}) &= \left(\frac{\eta^2 S^2}{2\bar{\nu}} - 2\right)\Sigma  \\
    g(\Sigma, \bar{\nu}) &= (1-\beta_2)(S^2\Sigma - \bar{\nu})
\end{align*}
$J$ is given by the partial derivatives of $f$ and $g$:
\begin{equation*}
    J = \begin{pmatrix} 
    \frac{\partial f}{\partial \Sigma} & \frac{\partial f}{\partial \bar{\nu}} \\[1ex]
    \frac{\partial g}{\partial \Sigma} & \frac{\partial g}{\partial \bar{\nu}} 
    \end{pmatrix}
\end{equation*}
First, we compute the partial derivatives:
\begin{align*}
    \frac{\partial f}{\partial \Sigma} &= \frac{\eta^2 S^2}{2\bar{\nu}} - 2 \\
    \frac{\partial f}{\partial \bar{\nu}} &= -\frac{\eta^2 S^2 \Sigma}{2\bar{\nu}^2} \\
    \frac{\partial g}{\partial \Sigma} &= (1-\beta_2)S^2 \\
    \frac{\partial g}{\partial \bar{\nu}} &= -(1-\beta_2)
\end{align*}
Next, we evaluate these derivatives at the fixed point $\Sigma^* = \frac{\eta^2}{4}$ and $\bar{\nu}^* = \frac{S^2 \eta^2}{4}$:

For $J_{11}$:
\begin{equation*}
    \frac{\partial f}{\partial \Sigma} \Bigg|_* = \frac{\eta^2 S^2}{2\left(\frac{S^2 \eta^2}{4}\right)} - 2 =  0
\end{equation*}
For $J_{12}$:
\begin{equation*}
    \frac{\partial f}{\partial \bar{\nu}} \Bigg|_* = -\frac{\eta^2 S^2 \left(\frac{\eta^2}{4}\right)}{2\left(\frac{S^2 \eta^2}{4}\right)^2}  = -\frac{2}{S^2}
\end{equation*}
For $J_{21}$ and $J_{22}$ (which are independent of $\Sigma$ and $\bar{\nu}$):
\begin{align*}
    \frac{\partial g}{\partial \Sigma} \Bigg|_* &= (1-\beta_2)S^2 \\
    \frac{\partial g}{\partial \bar{\nu}} \Bigg|_* &= -(1-\beta_2)
\end{align*}
We then have for our Jacobian at the fixed point:
\begin{equation*}
    J^* = \begin{pmatrix} 
    0 & -\frac{2}{S^2} \\ 
    (1-\beta_2)S^2 & -(1-\beta_2) 
    \end{pmatrix}
\end{equation*}
To determine stability, we must compute the trace and the determinant of $J^*$:
\begin{align*}
     \Tr(J^*) &  = -(1-\beta_2) \\
    \Det(J^*) &= 2(1-\beta_2)
\end{align*}
As $\beta_2 < 1$, we have that $\Tr(J^\ast) < 0$ and $\Det(J^\ast) > 0$. So the fixed point is stable.

This highlights an important difference between gradient descent and RMSProp. Gradient descent exhibits a strict sharpness threshold: on a quadratic loss, the iterates diverge to infinity if $S > 2/\eta$. For RMSProp, the preconditioner perfectly adapts to the local sharpness---it never diverges to infinity for any sharpness. Instead of diverging, RMSProp locks into a stable oscillation with a half-displacement $\delta^* = \eta/2$. 

\section{Rod Flow for Adam}
\label{app:adam}

\begin{table}[t]
\centering
\caption{Summary of rod flow models. Each row lists the optimizer, its
preconditioner type, and the predicted self-stabilization threshold for the
preconditioned sharpness.}
\label{tab:optimizers}
\begin{tabular}{ll@{\quad}l}
\toprule
\textbf{Optimizer} & \textbf{Preconditioner} & \textbf{Preconditioned sharpness threshold} \\
\midrule
Gradient descent & $I$ & $\frac{2}{\eta}$ \\[4pt]
Heavy ball & $I$ & $\frac{2}{\eta}\cdot\frac{1+\beta}{1-\beta}$ \\[4pt]
Nesterov & $I$ & $\frac{2}{\eta}\cdot\frac{(1+\beta)}{(1-\beta)(1+2\beta)}$ \\[4pt]
Scalar RMSProp & scalar $\nu$ & $\frac{2}{\eta}$ \\[4pt]
RMSProp & per-component $\nu$ & $\frac{2}{\eta}$ \\[4pt]
Scalar Adam & scalar $\nu$ & $\frac{2}{\eta}\cdot\frac{1+\beta}{1-\beta}$ \\[4pt]
Scalar NAdam & scalar $\nu$ & $\frac{2}{\eta}\cdot\frac{(1+\beta)}{(1-\beta)(1+2\beta)}$ \\[4pt]
Adam & per-component $\nu$ & $\frac{2}{\eta}\cdot\frac{1+\beta}{1-\beta}$ \\[4pt]
NAdam & per-component $\nu$ & $\frac{2}{\eta}\cdot\frac{(1+\beta)}{(1-\beta)(1+2\beta)}$ \\[4pt]
\bottomrule
\end{tabular}
\end{table}

\subsection{Derivation of Rod Flows}

\subsubsection{Adam}

The discrete update equations for Adam are given as:
\begin{align}
\nu_{t+1} &= \beta_2 \nu_t + (1-\beta_2) \nabla L(w_t)^{\odot 2}, \label{eq:adam_update_nu} \\
m_{t+1} &= \beta_1 m_t + (1-\beta_1)\nabla L(w_t), \label{eq:adam_update_m} \\
w_{t+1} &= w_t - \eta \, P_{t+1}^{-1} \, m_{t+1}. \label{eq:adam_update_w}
\end{align}
where $\beta_1$ is the momentum coefficient and $\beta_2$ controls the decay of the second moment estimate. 

To construct the rod flow for Adam, we must carefully treat the distinct behaviors of these three variables at the edge of stability.

As was the case with heavy ball momentum, the position and momentum oscillate in lockstep. It is therefore natural to concatenate them into a single phase-space vector $z_t \in \R^{2d}$:
\begin{equation}
z_t = \begin{pmatrix} w_t \\ m_t \end{pmatrix}.
\end{equation}
We define the phase-space midpoint $\bar{z}_t$ and the phase-space half-displacement $\Delta_t$ exactly as before:
\begin{align}
\bar{z}_t &= \tfrac{1}{2}(z_{t+1} + z_t) = \begin{pmatrix} \bar{w}_t \\ \bar{m}_t \end{pmatrix}, \\
\Delta_t &= \tfrac{1}{2}(z_{t+1} - z_t) = \begin{pmatrix} \delta_t \\ \gamma_t \end{pmatrix}.
\end{align}
By substituting the momentum update (Equation \ref{eq:adam_update_m}) directly into the position update (Equation \ref{eq:adam_update_w}), we can define the phase-space update function $\Phi$:
\begin{equation}
z_{t+1} = z_t + \Phi(z_t; \nu_{t+1}), \qquad \Phi(z; \nu) = \begin{pmatrix} -\eta\, P^{-1}(\nu)\bigl[\beta_1 m + (1-\beta_1)\nabla L(w)\bigr] \\ (1-\beta_1)\bigl[\nabla L(w) - m\bigr] \end{pmatrix}. \label{eq:Phi_adam}
\end{equation}
For the difference equation of $\bar{w}$, we have:
\begin{align*}
\bar{w}_{t+1} - \bar{w}_t &= \frac{w_{t+2} + w_{t+1}}{2} - \frac{w_{t+1} + w_t}{2}\\
&= \frac{w_{t+2} - w_{t+1}}{2} + \frac{w_{t+1} - w_t}{2} \\
&=-\frac{\eta}{2(\sqrt{\nu_{t+2}} + \varepsilon)}\, m_{t+2} -\frac{\eta}{2(\sqrt{\nu_{t+1}} + \varepsilon)} \, m_{t+1} \\
&= -\frac{\eta}{2(\sqrt{\nu_{t+2}} + \varepsilon)} [\beta_1 m_{t+1} + (1-\beta_1) \nabla L(w_{t+1})]\\
&\qquad{} -\frac{\eta}{2(\sqrt{\nu_{t+1}} + \varepsilon)}[ \beta_1 m_{t} + (1-\beta_1) \nabla L(w_t)] \\ 
&\approx -\frac{\eta}{(\sqrt{\bar{\nu}} +\varepsilon)} \left[\beta_1 \frac{(m_{t+1} + m_t)}{2} + (1-\beta_1)\frac{(\nabla L(w_{t+1}) + \nabla L(w_t))}{2} \right] \\
&= - \frac{\eta}{(\sqrt{\bar{\nu}} + \varepsilon)}\, \bigl [\beta_1 \bar{m} + (1-\beta_1) \bar{g} \bigr]
\end{align*}
For the difference equation of $\bar{m}$:
\begin{align*}
\bar{m}_{t+1} - \bar{m}_t &= \frac{m_{t+2} + m_{t+1}}{2} - \frac{m_{t+1} + m_t}{2}\\
&= \frac{m_{t+2} - m_{t+1}}{2} + \frac{m_{t+1} - m_t}{2}\\
&= \frac{(1-\beta_1)}{2} \bigg[\nabla L(w_{t+1}) - m_{t+1} \bigg] + \frac{(1-\beta_1)}{2} \bigg[\nabla L(w_t) - m_t \bigg] \\
&= (1-\beta_1) \left [\frac{\nabla L(w_{t+1}) + \nabla L(w_t)}{2} - \frac{m_{t+1} + m_{t}}{2} \right] \\
&=(1-\beta_1)\, [\bar{g} - \bar{m}]
\end{align*}
As was the case with RMSProp, the difference equation for $\bar{w}$ is not exact (unlike in the gradient descent case). Because we do not track the half-difference of the second-moment $\nu$, an approximation must be made. However, this approximation should be negligible---especially for large values of $\beta_2$.

Let $\Phi_\pm$ denote $\Phi(\bar{z} \pm \Delta)$. Then all together, the phase-space difference equation can be expressed as:
\begin{align*}
\bar{z}_{t+1} - \bar{z}_t &= \tfrac{1}{2}\bigl[\Phi_+ + \Phi_-\bigr].
\end{align*}
For the difference equation of the extent, we have:
\begin{align*}
\Delta_{t+1} \otimes \Delta_{t+1} - \Delta_t \otimes \Delta_t &= \Delta_{t+1} \otimes \Delta_{t+1} - \Delta_t \otimes \Delta_t + (\Delta_t \otimes \Delta_t - \Delta_t \otimes \Delta_t)\\
&= \Delta_{t+1} \otimes \Delta_{t+1} + \Delta_t \otimes \Delta_t - 2 \Delta_t \otimes \Delta_t\\
&=\tfrac{1}{4}  ( \Phi_+ \otimes \Phi_+ + \Phi_- \otimes \Phi_- ) - 2 \Delta_t \otimes \Delta_t.
\end{align*}
And for the difference equation of the midpoint of $\nu$:
\begin{align*}
\bar{\nu}_{t+1} - \bar{\nu}_t &= \frac{\nu_{t+2} + \nu_{t+1}}{2} - \frac{\nu_{t+1} + \nu_t}{2} \\
&= \frac{\nu_{t+2} - \nu_{t+1}}{2} + \frac{\nu_{t+1} - \nu_t}{2} \\
&= \frac{(1-\beta_2)}{2} \bigg [\nabla L(w_{t+1})^{\odot 2} - \nu_{t+1}\bigg] + \frac{(1-\beta_2)}{2} \bigg[\nabla L(w_t)^{\odot 2} - \nu_t \bigg] \\
&= (1-\beta_2) \left [\frac{\nabla L(w_{t+1})^{\odot 2} + \nabla L(w_t)^{\odot 2}}{2} - \frac{\nu_{t+1} + \nu_t}{2} \right] \\
&= (1 -\beta_2) \left [\frac{\nabla L(w_{t+1})^{\odot 2} + \nabla L(w_t)^{\odot 2}}{2} - \bar{\nu} \right]
\end{align*}
Promoting these discrete difference equations to continuous-time ODEs yields the rod flow for Adam:
\begin{align}
\frac{d\bar{w}}{dt} &= -\eta \, P^{-1}(\bar{\nu})\bigl[\beta_1\, \bar{m} + (1-\beta_1)\,\bar{g}\bigr], \label{eq:ode_wbar_adam_app}\\
\frac{d\bar{m}}{dt} &= (1-\beta_1)\bigl[\bar{g} - \bar{m}\bigr], \label{eq:ode_mbar_adam_app}\\
\frac{d\bar{\nu}}{dt} &= (1-\beta_2)\, \biggl (\frac{\nabla L_+^{\odot 2} + \nabla L_-^{\odot 2}}{2} - \bar{\nu} \biggr), \label{eq:ode_nubar_adam_app} \\
\frac{d\Sigma}{dt} &= \tfrac{1}{4}\bigl[\Phi_+ \otimes \Phi_+ + \Phi_- \otimes \Phi_-\bigr] - 2\,\Sigma, \label{eq:ode_sigma_adam_app}
\end{align}

\subsubsection{NAdam}

The update equations for NAdam are given as:
\begin{align}
\nu_{t+1} &= \beta_2 \nu_t + (1-\beta_2) \nabla L(w_t)^{\odot 2}, \label{eq:nadam_update_nu} \\
m_{t+1} &= \beta_1 m_t + (1-\beta_1)\nabla L(w_t), \label{eq:nadam_update_m} \\
w_{t+1} &= w_t - \eta \, P^{-1}_{t+1}\, \widetilde{m}_{t+1}. \label{eq:nadam_update_w}
\end{align}
where $\widetilde{m}$ is the modified momentum:
\begin{equation}
\widetilde{m}_{t+1} = \beta_1 m_{t+1} + (1-\beta_1) \nabla L(w_t)
\end{equation}
Relative to raw momentum, the modified momentum more heavily weights the most recent gradient. We can express the modified momentum in terms of $m_t$ and $\nabla L(w_t)$:
\begin{align*}
\widetilde{m}_{t+1} &= \beta_1 m_{t+1} + (1-\beta_1) \nabla L(w_t) \\
&= \beta_1 [\beta_1 m_t + (1-\beta_1) \nabla L(w_t)] + (1-\beta_1)\nabla L(w_t) \\
&= \beta_1^2 m_t + [\beta_1(1-\beta_1) + (1-\beta_1)] \nabla L(w_t) \\
&= \beta_1^2 m_t + (1-\beta_1^2) \nabla L(w_t)
\end{align*}
By substituting this modified momentum directly into the position update (Equation \ref{eq:nadam_update_w}), we define the phase-space update function $\Phi$ for NAdam:
\begin{equation}
 \Phi(z; \nu) = \begin{pmatrix} -\eta \, P^{-1}(\nu)\bigl[\beta_1^2 m + (1-\beta_1^2)\nabla L(w)\bigr] \\ (1-\beta_1)\bigl[\nabla L(w) - m\bigr] \end{pmatrix}. \label{eq:Phi_nadam}
\end{equation}
For the difference equation of $\bar{w}$, we have:
\begin{align*}
\bar{w}_{t+1} - \bar{w}_t &= \frac{w_{t+2} + w_{t+1}}{2} - \frac{w_{t+1} + w_t}{2}\\
&= \frac{w_{t+2} - w_{t+1}}{2} + \frac{w_{t+1} - w_t}{2} \\
&=-\frac{\eta}{2(\sqrt{\nu_{t+2}} + \varepsilon)}\, \widetilde{m}_{t+2} -\frac{\eta}{2(\sqrt{\nu_{t+1}} + \varepsilon)} \, \widetilde{m}_{t+1} \\
&= -\frac{\eta}{2(\sqrt{\nu_{t+2}} + \varepsilon)} [\beta_1^2 m_{t+1} + (1-\beta_1^2) \nabla L(w_{t+1})]\\
&\qquad{} -\frac{\eta}{2(\sqrt{\nu_{t+1}} + \varepsilon)}[ \beta_1^2 m_{t} + (1-\beta_1^2) \nabla L(w_t)] \\
&\approx -\frac{\eta}{(\sqrt{\bar{\nu}} +\varepsilon)}\, \biggl[\beta_1^2 \frac{(m_{t+1} + m_t)}{2} + (1-\beta_1^2)\frac{(\nabla L(w_{t+1}) + \nabla L(w_t))}{2} \biggr] \\
&= - \frac{\eta}{(\sqrt{\bar{\nu}} + \varepsilon)}\, \bigl [\beta_1^2 \bar{m} + (1-\beta_1^2) \bar{g} \bigr]
\end{align*}
Because the updates for $m_t$ and $\nu_t$ remain structurally identical to standard Adam, the difference equations for $\bar{m}$ and $\bar{\nu}$ are unchanged:
\begin{align*}
\bar{m}_{t+1} - \bar{m}_t &= (1-\beta_1)\, [\bar{g} - \bar{m}] \\
\bar{\nu}_{t+1} - \bar{\nu}_t &= (1 -\beta_2)\, \biggl [\frac{\nabla L(w_{t+1})^{\odot 2} + \nabla L(w_t)^{\odot 2}}{2} - \bar{\nu} \biggr]
\end{align*}
Similarly, the phase-space difference equation and the extent difference equation take the exact same form as Adam, substituting our new $\Phi_\pm$:
\begin{align*}
\bar{z}_{t+1} - \bar{z}_t &= \tfrac{1}{2}\bigl[\Phi_+ + \Phi_-\bigr] \\
\Delta_{t+1} \otimes \Delta_{t+1} - \Delta_t \otimes \Delta_t &= \tfrac{1}{4}  ( \Phi_+ \otimes \Phi_+ + \Phi_- \otimes \Phi_- ) - 2 \Delta_t \otimes \Delta_t
\end{align*}
Promoting these discrete difference equations to continuous-time ODEs yields the rod flow for NAdam:
\begin{align}
\frac{d\bar{w}}{dt} &= -\eta \, P^{-1}(\bar{\nu})\bigl[\beta_1^2\, \bar{m} + (1-\beta_1^2)\,\bar{g}\bigr], \label{eq:ode_wbar_nadam}\\
\frac{d\bar{m}}{dt} &= (1-\beta_1)\bigl[\bar{g} - \bar{m}\bigr], \label{eq:ode_mbar_nadam}\\
\frac{d\bar{\nu}}{dt} &= (1-\beta_2)\, \biggl (\frac{\nabla L_+^{\odot 2} + \nabla L_-^{\odot 2}}{2} - \bar{\nu} \biggr), \label{eq:ode_nubar_nadam} \\
\frac{d\Sigma}{dt} &= \tfrac{1}{4}\bigl[\Phi_+ \otimes \Phi_+ + \Phi_- \otimes \Phi_-\bigr] - 2\,\Sigma, \label{eq:ode_sigma_nadam}
\end{align}

\subsection{NAdam versus Nesterov Momentum}

Despite its name, NAdam is not \textit{quite} Adam with Nesterov momentum replacing heavy-ball momentum. True Nesterov momentum requires a look-ahead gradient---evaluating the gradient at the shifted point $w_t - \eta\beta m_t$ rather than the current iterate---which is awkward to slot into standard training loops. The look-ahead gradient requires temporarily perturbing the parameters before each forward/backward pass, and it does not compose cleanly with Adam's bias correction or a time-varying $\beta$. Instead, NAdam adopts Dozat's formulation (building on Sutskever's change of variables), which evaluates the gradient only at the current position. With constant $\beta$, it is mathematically equivalent to Nesterov momentum---only in a shifted coordinate system.

Recall the Nesterov momentum update equations:
\begin{align*}
m_{t+1} &= \beta m_t + (1-\beta) \nabla L(w_t - \eta\beta m_t), \\
w_{t+1} &= w_t - \eta \bigl[\beta m_t + (1-\beta) \nabla L(w_t - \eta \beta m_t)\bigr].
\end{align*}
Dozat's formulation (equivalent to NAdam without the preconditioner) replaces these with:
\begin{align*}
m_{t+1} &= \beta m_t + (1-\beta) \nabla L(w_t), \\
w_{t+1} &= w_t - \eta \bigl[\beta^2 m_t + (1-\beta^2) \nabla L(w_t)\bigr].
\end{align*}
To see the relationship between the two, introduce the change of variables
$\theta_t \coloneqq w_t - \eta \beta m_t$. One can see that $\theta_t$ is precisely the
``look-ahead'' point at which Nesterov evaluates the gradient. Substituting
into the Nesterov update for $w_{t+1}$ gives:
\begin{align*}
\theta_{t+1} + \eta \beta m_{t+1}
  &= \theta_t + \eta \beta m_t - \eta\bigl[\beta m_t + (1-\beta) \nabla L(\theta_t)\bigr].
\end{align*}
Isolating $\theta_{t+1}$ yields:
\begin{align*}
\theta_{t+1}
  &= \theta_t - \eta\bigl[\beta m_{t+1} + (1-\beta) \nabla L(\theta_t)\bigr].
\end{align*}
Expanding $m_{t+1} = \beta m_t + (1-\beta)\nabla L(\theta_t)$ in the last
expression recovers Dozat's position update rule.

Hence, Nesterov momentum and Dozat's formulation are equivalent up to a change
of variables: given an initial condition $(w_0, m_0)$ for Nesterov, the
shifted iterate $\theta_0 = w_0 - \eta \beta m_0$ run under Dozat's update
produces the same trajectory of look-ahead points, and shifting back recovers
the original Nesterov iterates.

We can show that Dozat's formulation has the same sharpness threshold $S^\ast$ on a quadratic objective as Nesterov momentum. Consider the quadratic loss $L(w) = \tfrac{1}{2} S w^2$. We have that $\nabla L = Sw$. Substituting into Dozat's formulation gives:
\begin{align}
m_{t+1} &= \beta m_t + (1-\beta) S w_t, \\
w_{t+1} &= w_t - \eta \bigl[\beta^2 m_t + (1-\beta^2) S w_t\bigr].
\end{align}
This is a linear system, which we can express in matrix form as:
\begin{equation}
\begin{pmatrix} w_{t+1} \\ m_{t+1} \end{pmatrix}
= M \begin{pmatrix} w_t \\ m_t \end{pmatrix},
\qquad
M = \begin{pmatrix} 1 - \eta(1-\beta^2) S & -\eta \beta^2 \\ (1-\beta) S & \beta \end{pmatrix}.
\end{equation}
Our system is stable when the eigenvalues of $M$ lie within the unit circle. The characteristic polynomial is determined by the trace and determinant:
\begin{align}
\Tr(M) &= 1 + \beta - \eta(1-\beta^2) S, \\
\Det(M) &= \beta - \eta \beta (1-\beta) S.
\end{align}
This gives the characteristic polynomial:
\begin{equation}
\lambda^2 - \bigl[1 + \beta - \eta(1-\beta^2) S\bigr] \lambda + \bigl[\beta - \eta \beta (1-\beta) S\bigr] = 0.
\end{equation}
We can find the sharpness threshold $S^\ast$ by plugging in $\lambda = -1$, which corresponds to the period-2 oscillations:
\begin{equation}
1 + \bigl[1 + \beta - \eta(1-\beta^2) S^\ast\bigr] + \bigl[\beta - \eta \beta (1-\beta) S^\ast\bigr] = 0.
\end{equation}
Collecting terms:
\begin{equation}
2(1 + \beta) - \eta (1-\beta) S^\ast \bigl[1+ 2 \beta\bigr] = 0,
\end{equation}
Solving for $S^\ast$ yields:
\begin{equation}
S^\ast = \frac{2(1+\beta)}{\eta (1-\beta)(1 + 2\beta)}.
\end{equation}
This matches the sharpness threshold for Nesterov momentum on a quadratic.

However, in the presence of bias correction, Dozat's formulation of NAdam does not coincide with a simple application of Nesterov momentum to Adam.

\subsection{Theoretical Analysis}

\subsubsection{Linear Loss}
 As was the case with RMSProp, for the theoretical analysis section, we will drop the regularization parameter $\varepsilon$.
 
Consider the linear loss $L(w) = b \cdot w$. The gradient is constant everywhere: $\nabla L(w) = b$. Since $\nabla L_+ = \nabla L_- = b$, the average gradient is $\bar{g} = b$. The rod flow equations become:
\begin{align}
    \frac{d\bar{\nu}}{dt} &= (1-\beta_2)(b^2 - \bar{\nu})\,, \\
    \frac{d\bar{m}}{dt} &= (1-\beta_1)(b - \bar{m})\,, \\
    \frac{d\bar{w}}{dt} &= -\frac{\eta}{\sqrt{\bar{\nu}}}\,\bigl[\beta_1 \bar{m} + (1-\beta_1)b\bigr]\,, \\
    \frac{d\Sigma}{dt} &= \tfrac{1}{2} \Phi \otimes \Phi - 2\Sigma\,,
\end{align}
where we have that:
$$\Phi = \begin{pmatrix} -\frac{\eta}{\sqrt{\bar{\nu}}}[\beta_1 \bar{m} + (1-\beta_1)b] \\ (1-\beta_1)(b - \bar{m}) \end{pmatrix}$$
Setting $\frac{d\bar{\nu}}{dt} = 0$ and $\frac{d\bar{m}}{dt} = 0$ gives the steady-state momentum and second moment:
\begin{align}
\bar{m}^* &= b\,,\\
    \bar{\nu}^* &= b^2\,,
\end{align}
Substituting our expressions for $\bar{\nu}^*$ and $\bar{m}^*$ into the $\bar{w}$ equation yields the steady-state equation of motion for the midpoint of the position:
\begin{equation}
    \frac{d\bar{w}}{dt} =- \eta \operatorname{sign}(b)\,.
\end{equation}
Substituting the steady-state values into the phase-space update yields: 
$$\Phi^* = \begin{pmatrix} - \eta \operatorname{sign}(b) \\ 0 \end{pmatrix}.$$ Setting $\frac{d\Sigma}{dt} = 0$ gives the steady-state extent for the position coordinate:
\begin{equation}
    \Sigma_{\delta \delta}^* = \frac{\eta^2}{4}\,.
\end{equation}
Along flat directions, Adam behaves similarly to RMSProp. Because the gradient does not change, the exponential moving average of the gradient is indistinguishable from the instantaneous gradient. Thus, Adam recovers the behavior of RMSProp, acting as Sign-GD with step size $\eta$.

\subsubsection{Quadratic Loss}

Consider the quadratic $L(w) = \frac{1}{2}Sw^2$ with $S > 0$. We will work in one dimension, meaning the phase-space half-displacement is a 2D vector: $$\Delta = \begin{pmatrix} \delta \\ \gamma \end{pmatrix}$$ 
where $\delta$ represents the position half-displacement and $\gamma$ represents the momentum half-displacement. The phase-space extent is $\Sigma = \Delta \otimes \Delta$.

Evaluating the gradients at the endpoints of the rod yields:
\begin{align}
    \nabla L_+ &= S(\bar{w} + \delta) \\
    \nabla L_- &= S(\bar{w} - \delta)
\end{align}
Summing these terms and summing their squares gives:
\begin{align}
    \nabla L_+ + \nabla L_- &= 2S\bar{w} \\
    \nabla L_+^2 + \nabla L_-^2 &= S^2(\bar{w} + \delta)^2 + S^2(\bar{w} - \delta)^2 = 2S^2(\bar{w}^2 + \delta^2)
\end{align}
Plugging these into our rod flow equations for Adam:
\begin{align}
    \frac{d\bar{w}}{dt} &= -\frac{\eta}{\sqrt{\bar{\nu}}}\bigl[\beta_1\, \bar{m} + (1-\beta_1)S\bar{w}\bigr] \\
    \frac{d\bar{m}}{dt} &= (1-\beta_1)\bigl[S\bar{w} - \bar{m}\bigr] \\
    \frac{d\bar{\nu}}{dt} &= (1-\beta_2)\left(S^2(\bar{w}^2 + \delta^2) - \bar{\nu}\right) \\
    \frac{d\Sigma}{dt} &= \tfrac{1}{4}\bigl[\Phi_+ \otimes \Phi_+ + \Phi_- \otimes \Phi_-\bigr] - 2\Sigma
\end{align}
Consider the $(\bar{w}, \bar{m})$ subsystem. It is a linear system:
\begin{equation}
    \frac{d}{dt} \begin{pmatrix} \bar{w} \\ \bar{m} \end{pmatrix} = A \begin{pmatrix} \bar{w} \\ \bar{m} \end{pmatrix}\,, \quad A = \begin{pmatrix} -\frac{\eta(1-\beta_1)S}{\sqrt{\bar{\nu}}} & -\frac{\eta \beta_1}{\sqrt{\bar{\nu}}} \\ (1-\beta_1)S & -(1-\beta_1) \end{pmatrix}\,.
\end{equation}
To determine whether the system decays to the origin, we need to evaluate the trace and the determinant of $A$. 
\begin{align*}
    \Tr(A) &= -(1-\beta_1)\left(\frac{\eta S}{\sqrt{\bar{\nu}}} + 1\right) \\
    \Det(A) 
    &= \frac{\eta S (1-\beta_1)}{\sqrt{\bar{\nu}}}
\end{align*}
As $\Tr(A) < 0$ and $\Det(A) > 0$, we have that our system decays to the origin.

We will now analyze the coupled dynamics of the extent and the second moment at steady-state.
The rod flow equation for the second moment becomes:
\begin{equation}
    \frac{d\bar{\nu}}{dt} = (1-\beta_2)\bigl(S^2\delta^2 - \bar{\nu}\bigr).
\end{equation}
Setting $\frac{d\bar{\nu}}{dt} = 0$ yields:
\begin{equation}
    \bar{\nu}^* = S^2\delta^2
\end{equation}
We will now examine the phase-space extent. At steady-state edge of stability, $\Phi_+ = -2\Delta^\ast$. Writing out the two components of this equation explicitly:
\begin{align}
    -2\delta^\ast &= -\frac{\eta}{\sqrt{\bar{\nu}^*}} \bigl[\beta_1 \gamma^\ast + (1-\beta_1)S\delta^\ast\bigr]\,, \label{eq:adam_quad_delta} \\
    -2\gamma^\ast &= (1-\beta_1)(S\delta^\ast - \gamma^\ast)\,. \label{eq:adam_quad_gamma}
\end{align}
We can solve this system to find the fixed point. First, rearrange the momentum equation (\ref{eq:adam_quad_gamma}) to find the relationship between the momentum half-difference $\gamma$ and the position half-difference $\delta$:
\begin{align}
    \gamma^\ast &= -\left(\frac{1-\beta_1}{1+\beta_1}\right) S\delta\,.
\end{align}
Substituting the expression for $\gamma^\ast$ back into the position equation \eqref{eq:adam_quad_delta} and using the steady-state second-moment relation $\sqrt{\bar{\nu}^\ast} = S\delta^\ast$, we obtain a single equation for $\delta^\ast$:
\begin{align}
    -2\delta^\ast &= -\frac{\eta}{S\delta^\ast} \left[ -\beta_1 \left(\frac{1-\beta_1}{1+\beta_1}\right) S\delta^\ast + (1-\beta_1)S\delta^\ast \right] \nonumber \\
    &= -\frac{\eta}{S\delta^\ast}\,(1-\beta_1)\,S\delta^\ast \left[ 1 - \frac{\beta_1}{1+\beta_1} \right] \nonumber \\
    &= -\eta \left(\frac{1-\beta_1}{1+\beta_1}\right).
\end{align}
The factors of $S\delta^\ast$ cancel, leaving an explicit expression for the steady-state half-displacement:
\begin{equation}
    \delta^\ast = \frac{\eta}{2} \left(\frac{1-\beta_1}{1+\beta_1}\right).
\end{equation}
It is instructive to compare these results with those of RMSProp. 

Broadly, an optimizer wants to maximize progress in flat directions while minimizing the bouncing amplitude in the sharp directions: large oscillations in the sharp directions correspond to less stable gradients.

Adam behaves similarly to RMSProp in flat directions, where both act like Sign-GD with step size $\eta$. Where their behavior differs is in the sharp directions: for the same step size, Adam bounces far less. By incorporating heavy-ball momentum, Adam strictly reduces the amplitude of edge-of-stability bouncing, suppressing the steady-state half-displacement by the factor $(1 - \beta_1)/(1 + \beta_1)$. For a typical value $\beta_1 = 0.9$, the bouncing amplitude is roughly $5\%$ of what it would be under RMSProp.

This comparison also suggests that Adam is more naturally viewed as RMSProp with momentum rather than momentum with an adaptive step size: its behavior in flat directions is essentially that of RMSProp, with momentum playing a more prominent role in its behavior in the sharp directions.

\section{Backward Error Analysis}
\label{app:bea}

\subsection{Overview}

Backward error analysis (BEA) is a technique for understanding the continuous dynamics underlying a discrete update rule. Rather than viewing a discrete-time system as an approximation to some continuous flow, BEA adopts a converse perspective: we ask which continuous dynamics the discrete iterates solve \emph{exactly}. Given a discrete update rule:
\begin{equation}
x_{t+1} - x_t = D(x_t),
\end{equation}
we seek a modified vector field $\tilde{V}(x)$ such that the solution to
\begin{equation}
\frac{dx}{dt} = \tilde{V}(x)
\end{equation}
exactly interpolates the discrete iterates: if $x(t) = x_t$, then $x(t+1) = x_{t+1}$.

Assuming $x(t)$ is smooth, we can express the discrete forward step using a Taylor series:
\begin{equation}
x_{t+1} = x_t + \sum_{n=1}^\infty \frac{1}{n!} \frac{d^n x}{dt^n}.
\end{equation}
Because $x(t)$ is governed by the time-independent ODE $\dot{x} = \tilde{V}(x)$, all higher-order time derivatives can be expressed purely in terms of spatial derivatives via the chain rule. This is elegantly captured by the Lie derivative associated with $\tilde{V}$, which we denote $\tilde{\mathcal{L}} = \tilde{V} \cdot \nabla$.

Applying this operator repeatedly gives the time derivatives:
\begin{align*}
\dot{x} &= \tilde{\mathcal{L}}x = \tilde{V}, \\
\ddot{x} &= \tilde{\mathcal{L}}^2 x = \tilde{\mathcal{L}}\tilde{V} = \nabla \tilde{V} \cdot \tilde{V}, \\
\dddot{x} &= \tilde{\mathcal{L}}^3 x = \tilde{\mathcal{L}}(\nabla \tilde{V} \cdot \tilde{V}) = \nabla(\nabla \tilde{V} \cdot \tilde{V}) \cdot \tilde{V}.
\end{align*}
Using this operator, the Taylor series becomes an operator exponential:
\begin{equation}
x_{t+1} = e^{\tilde{\mathcal{L}}} x_t.
\end{equation}
Equating this to our discrete update $x_{t+1} = x_t + D(x_t)$, we obtain the fundamental operator equation for the modified vector field:
\begin{equation}
D(x) = (e^{\tilde{\mathcal{L}}} - I)x = \sum_{n=1}^\infty \frac{1}{n!} \tilde{\mathcal{L}}^n x.
\end{equation}
Expanding this explicitly in terms of $\tilde{V}$:
\begin{equation}
D = \tilde{V} + \tfrac{1}{2} \nabla \tilde{V} \cdot \tilde{V} + \tfrac{1}{6} \nabla(\nabla \tilde{V} \cdot \tilde{V}) \cdot \tilde{V} + \mathcal{O}(\tilde{V}^4).
\end{equation}

To solve for $\tilde{V}$ in terms of the known discrete displacement $D$, we perform a perturbative inversion. We assume $D \sim \mathcal{O}(\eta)$ and expand the modified vector field as an infinite series graded by powers of $\eta$:
\begin{equation}
\tilde{V} = \tilde{V}_1 + \tilde{V}_2 + \tilde{V}_3 + \cdots,
\end{equation}
where $\tilde{V}_k \sim \mathcal{O}(\eta^k)$. We substitute this series into our expanded fundamental equation:
\begin{align}
D &= \left( \tilde{V}_1 + \tilde{V}_2 + \tilde{V}_3 + \cdots \right) \nonumber \\
  &\quad + \tfrac{1}{2} \nabla \left( \tilde{V}_1 + \tilde{V}_2 + \cdots \right) \cdot \left( \tilde{V}_1 + \tilde{V}_2 + \cdots \right) \nonumber \\
  &\quad + \tfrac{1}{6} \nabla \Big( \nabla \left( \tilde{V}_1 + \cdots \right) \cdot \left( \tilde{V}_1 + \cdots \right) \Big) \cdot \left( \tilde{V}_1 + \cdots \right) + \cdots
\end{align}
We now match terms of the same order in $\eta$ to systematically build up the modified vector field.

Retaining only the $O(\eta)$ terms, we immediately find:
\begin{equation}
\tilde{V}_1 = D.
\end{equation}

Next, we collect all $\mathcal{O}(\eta^2)$ terms. There are two such contributions: the second-order field $\tilde{V}_2$ itself, and the leading term of the $\tfrac{1}{2}\tilde{\mathcal{L}}^2 x$ expansion.
\begin{equation}
D = \tilde{V}_1 + \tilde{V}_2 + \tfrac{1}{2} \nabla \tilde{V}_1 \cdot \tilde{V}_1.
\end{equation}
Substituting $\tilde{V}_1 = D$ and solving for $\tilde{V}_2$:
\begin{equation}
\tilde{V}_2 = -\tfrac{1}{2} \nabla D \cdot D.
\end{equation}

Collecting all terms of order $\eta^3$:
\begin{equation}
D = \tilde{V}_1 + \tilde{V}_2 + \tilde{V}_3 + \tfrac{1}{2} \Big( \nabla \tilde{V}_1 \cdot \tilde{V}_2 + \nabla \tilde{V}_2 \cdot \tilde{V}_1 \Big) + \tfrac{1}{6} \nabla(\nabla \tilde{V}_1 \cdot \tilde{V}_1) \cdot \tilde{V}_1.
\end{equation}
Note that the final term can be rewritten using $\tilde{V}_2 = -\tfrac{1}{2}\nabla \tilde{V}_1 \cdot \tilde{V}_1$. Substituting this back and using $\tilde{V}_1 = D$ yields:
\begin{equation}
\tilde{V}_3 = -\tfrac{1}{2} \nabla D \cdot \tilde{V}_2 - \tfrac{1}{6} \nabla \tilde{V}_2 \cdot D.
\end{equation}

Note that $\tilde{V}_3$ can be computed using Jacobian-vector products (JVPs) of previously-computed lower-order vector fields, allowing for efficient algorithmic implementation. This generalizes to arbitrary $\tilde{V}_k$.

\begin{figure}
  \centering
  \textbf{\large Adam}\par\vspace{0.2cm}
  \includegraphics[width= \linewidth]{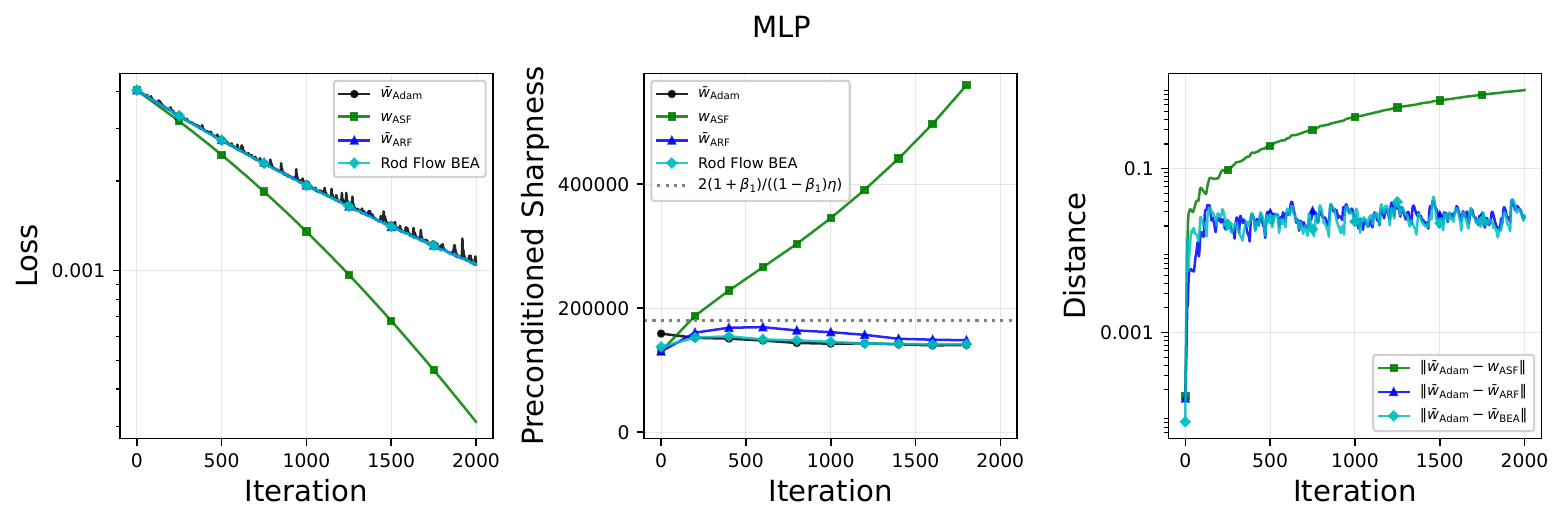}

\caption{\textbf{Backward Error Analysis for Adam.} Comparison of Adam against stable flow, rod flow, and rod flow with the BEA correction, evaluated on the loss, preconditioned sharpness, and distance between iterates. The BEA correction yields no noticeable improvement.}
  \label{fig:summary_adam_bea}
\end{figure}

\subsection{Backward Error Analysis for Rod Flow}

We apply the second-order backward error analysis (BEA) correction to the Adam rod flow center dynamics. The modified vector field $\tilde{V}$ is approximated by:

\begin{equation}
\tilde{V} \approx D - \frac{1}{2} \nabla D \cdot D
\end{equation}

where $D$ represents the discrete difference updates. The discrete displacement for the rod centers $\bar{z} = (\bar{w}, \bar{m})$ is given by:
\begin{align}
D_{\bar{w}} &= -\eta \, P^{-1}(\bar{\nu})\bigl[\beta_1 \bar{m} + (1-\beta_1)\bar{g}\bigr], \\
D_{\bar{m}} &= (1-\beta_1)(\bar{g} - \bar{m}),
\end{align}
We restrict the correction to the center variables due to timescale separation: the extent $\Sigma$ equilibrates on an $\mathcal{O}(1)$ timescale and the second-moment estimate $\bar{\nu}$ changes slowly relative to the center dynamics.

The blocks of the Jacobian of $D$ with respect to $\bar{z}$ are:
\begin{align}
\frac{\partial D_{\bar{w}}}{\partial \bar{w}} &= -\eta \, P^{-1}(\bar{\nu})(1-\beta_1) \bar{H}, &
\frac{\partial D_{\bar{w}}}{\partial \bar{m}} &= -\eta \, P^{-1}(\bar{\nu})\beta_1 I_d, \\
\frac{\partial D_{\bar{m}}}{\partial \bar{w}} &= (1-\beta_1) \bar{H}, &
\frac{\partial D_{\bar{m}}}{\partial \bar{m}} &= -(1-\beta_1) I_d,
\end{align}
where $\bar{H} = \frac{1}{2}(\nabla^2 L_+ + \nabla^2 L_-)$ is the average Hessian at the rod endpoints and $I_d$ is the $d \times d$ identity matrix.

Computing $\nabla D \cdot D$ blockwise gives:
\begin{align}
(\nabla D \cdot D)_{\bar{w}} &= -\eta \, P^{-1}(\bar{\nu})(1-\beta_1) \bar{H} D_{\bar{w}} -\eta \, P^{-1}(\bar{\nu})\beta_1 D_{\bar{m}}, \\
(\nabla D \cdot D)_{\bar{m}} &= (1-\beta_1) \bar{H} D_{\bar{w}} - (1-\beta_1) D_{\bar{m}}.
\end{align}

Subtracting half of each from the discrete step $D$ yields the BEA-corrected center dynamics for the continuous variables:
\begin{align}
\frac{d\bar{w}}{dt} &= D_{\bar{w}} + \frac{\eta}{2} \, P^{-1}(\bar{\nu})(1-\beta_1) \bar{H} D_{\bar{w}} + \frac{\eta}{2} \, P^{-1}(\bar{\nu})\beta_1 D_{\bar{m}}, \\
\frac{d\bar{m}}{dt} &=  D_{\bar{m}} + \frac{(1-\beta_1)}{2} D_{\bar{m}} - \frac{(1-\beta_1)}{2} \bar{H} D_{\bar{w}}.
\end{align}

\cref{fig:summary_adam_bea} compares the BEA-corrected rod flow to the uncorrected version. The two are essentially indistinguishable, which shows that the second-order truncation error between the discrete map and its modified equation is not what drives the discrepancy between rod flow and the discrete iterates. So what is causing it? Two likely candidates are (1) violations of the period-2 oscillation assumption and (2) the presence of multiple sharp directions in the Hessian. In any case, since the BEA correction does not improve accuracy in the deep learning setting, the simpler uncorrected form of rod flow is to be preferred.

\section{Computational Implementation}
\label{app:computation}

Implementing rod flow on realistically-sized neural networks introduces two challenges. First, the extent tensor $\Sigma$ is too large to store explicitly: it is $d \times d$ for non-momentum optimizers and $2d \times 2d$ for momentum-based optimizers. We address this with a low-rank representation of $\Sigma$. Second, Adam's bias-corrected second-moment estimate introduces an explicit dependence on the iterate counter that the continuous flow must track. We address this by incorporating the bias correction directly into the flow.

\subsection{Low-Rank Representation of \texorpdfstring{$\Sigma$}{Sigma}}\label{app:low-rank-phase-sigma}

Consider the rod flow ODE for $\Sigma$:
\[
\frac{d\Sigma}{dt} = \tfrac{1}{4} (\Phi_+ \otimes \Phi_+ + \Phi_- \otimes \Phi_-) - 2 \Sigma.
\]
Because of the separation of time scales between the center dynamics and the extent dynamics, $\Sigma$ is always in quasi-equilibrium:
\begin{equation}
\Sigma \approx \tfrac{1}{8}\bigl[\Phi_+(\Delta) \otimes \Phi_+(\Delta) + \Phi_-(\Delta) \otimes \Phi_-(\Delta)\bigr].
\end{equation}
This expression is at most rank two. In practice, the phase-space displacements at the rod endpoints point in nearly identical directions which causes $\Sigma$ to be effectively rank one. This justifies a low-rank representation of $\Sigma$: rather than storing the full matrix, we work in the low-dimensional subspace spanned by the current direction of $\Sigma$. We therefore represent:
\begin{equation}
    \Sigma = V \Lambda V^\top, \qquad V \in \mathbb{R}^{2d \times r}, \qquad \Lambda \in \mathbb{R}^{r \times r}.
\end{equation}
Here $V$ is an orthonormal matrix whose columns represent the directions along which the rod extends in phase space, and $\Lambda$ is a symmetric matrix whose entries give the squared magnitudes of the half-displacements along those directions. We set $r = 3$ in all experiments, which reduces the storage cost from $\mathcal{O}(d^2)$ to $\mathcal{O}(dr)$.

The update of $(V, \Lambda)$ proceeds in five steps:

\paragraph{Step 1: Decay.} Apply the exponential decay from the $-2\Sigma$ term in the ODE:
\begin{equation}
    \Lambda \leftarrow (1 - 2\,dt)\,\Lambda.
\end{equation}
Since $\Lambda$ is diagonal at the start of each step, this is simply element-wise multiplication.

\paragraph{Step 2: Project.} Decompose the phase-space displacements at the rod's endpoints, $\Phi_+$ and $\Phi_-$, into components parallel and perpendicular to the current eigenspace:
\begin{align}
    \Phi_+^{\parallel} &= V^\top \Phi_+, \qquad \Phi_+^{\perp} = \Phi_+ - V \Phi_+^{\parallel}, \\
    \Phi_-^{\parallel} &= V^\top \Phi_-, \qquad \Phi_-^{\perp} = \Phi_- - V \Phi_-^{\parallel}.
\end{align}
The parallel components are $r$-dimensional vectors, while the perpendicular components live in $\mathbb{R}^{2d}$ but are orthogonal to the current basis.

\paragraph{Step 3: Augment basis.} If the perpendicular components have significant norm (above a threshold $\epsilon_{\rm tol} = 10^{-10}$), add them as new basis directions:
\begin{equation}
    V_{\text{aug}} = \begin{bmatrix} V & \dfrac{\Phi_+^{\perp}}{\|\Phi_+^{\perp}\|} & \dfrac{\Phi_-^{\perp}}{\|\Phi_-^{\perp}\|} \end{bmatrix} \in \mathbb{R}^{2d \times (r+k)},
\end{equation}
where $k \in \{0, 1, 2\}$ counts how many perpendicular components exceed the threshold. We orthogonalize the new directions against each other via Gram--Schmidt.

\paragraph{Step 4: Add outer products.} Extend $\Lambda$ to the augmented basis by padding with zeros for the new directions, then add the rank-$2$ outer product contributions:
\begin{equation}
    \Lambda_{\text{aug}} \leftarrow \Lambda_{\text{aug}} + \tfrac{dt}{4} \left[(V_{\text{aug}}^\top \Phi_+)(V_{\text{aug}}^\top \Phi_+)^\top + (V_{\text{aug}}^\top \Phi_-)(V_{\text{aug}}^\top \Phi_-)^\top\right].
\end{equation}
This produces a symmetric matrix $\Lambda_{\text{aug}} \in \mathbb{R}^{(r+k) \times (r+k)}$ that is generally no longer diagonal.

\paragraph{Step 5: Truncate and reorthogonalize.} Compute the eigendecomposition $\Lambda_{\text{aug}} = U \tilde{\Lambda} U^\top$, where the columns of $U$ are eigenvectors and $\tilde{\Lambda}$ is diagonal with eigenvalues in decreasing order. Retaining only the top $r$ eigenpairs, we rotate the basis into the new eigenvector coordinates:
\begin{equation}
    V \leftarrow V_{\text{aug}} U_{:,\,:r}, \qquad \Lambda \leftarrow \mathrm{diag}(\tilde{\lambda}_1, \ldots, \tilde{\lambda}_r),
\end{equation}
where $U_{:,\,:r}$ denotes the first $r$ columns of $U$. The new $\Lambda$ is diagonal by construction. We finish with a QR re-orthogonalization of $V$ to correct for numerical drift.

\subsection{Bias Correction}
\label{app:bias-correction}

Adam utilizes bias-corrected moment estimates to counteract the fact that its exponential moving averages (EMAs) are initialized at zero. 

To understand why this is necessary, consider the linear loss $L = bw$. The gradient is constant for all time: $g_t = b$. 
If we initialize the moments at zero ($m_0 = 0$ and $\nu_0 = 0$), we have that the raw EMAs evolve as:
\begin{equation}
m_t = \beta_1 m_{t-1} + (1-\beta_1) g_t = (1 - \beta_1^t) b
\end{equation}
\begin{equation}
\nu_t = \beta_2 \nu_{t-1} + (1-\beta_2) g_t^2 = (1 - \beta_2^t) b^2
\end{equation}
Early in training, these estimates are drastically biased toward zero. With the standard $\beta_2 = 0.999$, the raw second moment $\nu_t$ requires $\mathcal{O}(10^3)$ iterations to properly equilibrate.

To resolve this, Adam defines the bias-corrected moments:
\begin{equation}
\hat{m}_t = \frac{m_t}{1 - \beta_1^t}
\end{equation}
\begin{equation}
\hat{\nu}_t = \frac{\nu_t}{1 - \beta_2^t}
\end{equation}
which, in our linear loss example, perfectly recover the true moments $b$ and $b^2$ from the very first step.

Our continuous-time rod flow implements this by inheriting a synthetic step counter from the outer discrete training loop. At each substep of the rod flow, we pass the current discrete iteration index $t$ and compute the bias-correction factors:
\begin{equation}
\mathrm{bc}_1 = 1 - \beta_1^{\,t+1}, \qquad \mathrm{bc}_2 = 1 - \beta_2^{\,t+1}.
\end{equation}
Note that the index is $t+1$ rather than $t$. Although the bias-correction factor is computed at time $t$, it is $m_{t+1}$ that actually drives the position update---and this is the moment estimate to which the correction must apply.

In full, the bias-corrected rod flow ODEs for Adam are:
\begin{align}
\frac{d\bar{w}}{dt} &= -\eta \, P^{-1}(\,\widehat{\bar{\nu}}\,)\,\bigl[\beta_1\, \widehat{\bar{m}} + (1-\beta_1)\, \widehat{\bar{g}}\bigr], \\[4pt]
\frac{d\bar{m}}{dt} &= (1-\beta_1)\bigl[\bar{g} - \bar{m}\bigr], \\[4pt]
\frac{d\bar{\nu}}{dt} &= (1-\beta_2)\,\biggl(\frac{\nabla L_+^{\odot 2} + \nabla L_-^{\odot 2}}{2} - \bar{\nu}\biggr),  \\[4pt]
\frac{d\Sigma}{dt} &= \tfrac{1}{4}\bigl[\Phi_+ \otimes \Phi_+ + \Phi_- \otimes \Phi_-\bigr] - 2\,\Sigma,
\end{align}
where the bias-corrected moments and gradient are
\begin{equation}
\widehat{\bar{m}} = \frac{\bar{m}}{\mathrm{bc}_1}, \qquad \widehat{\bar{\nu}} = \frac{\bar{\nu}}{\mathrm{bc}_2}, \qquad \widehat{\bar{g}} = \frac{\bar{g}}{\mathrm{bc}_1},
\end{equation}
so that the bracketed term in the position update,
\begin{equation}
\beta_1\, \widehat{\bar{m}} + (1-\beta_1)\, \widehat{\bar{g}} \;=\; \frac{\beta_1\, \bar{m} + (1-\beta_1)\, \bar{g}}{\mathrm{bc}_1},
\end{equation}
is the bias-corrected version of $m_{t+1}$.

\section{Experimental Details}
\label{app:experiments}

\subsection{Stable Flows}
\label{app:stable_flows}

The \textit{stable flow} of a discrete-time optimizer is its na\"{\i}ve continuous-time limit. For gradient descent, the stable flow is the familiar gradient flow:
\begin{equation}
\frac{dw}{dt} = -\eta\,\nabla L(w).
\end{equation}

Stable flows track the discrete optimizer well when the sharpness (or preconditioned sharpness) is below the EoS threshold. During EoS, they fail to capture the oscillatory dynamics that the discrete optimizer exhibits about its center trajectory. We use the stable flow as the natural baseline against which to benchmark rod flow.

Below are the stable flows for the eight optimizers that we studied.

\textbf{Heavy Ball}
\begin{align*}
\frac{dw}{dt} &= -\eta\bigl[\beta m + (1-\beta)\nabla L(w)\bigr] \\
\frac{dm}{dt} &= (1-\beta)\bigl[\nabla L(w) - m\bigr]
\end{align*}

\textbf{Nesterov}
\begin{align*}
\frac{dw}{dt} &= -\eta\bigl[\beta m + (1-\beta)\nabla L(w - \eta\beta m)\bigr] \\
\frac{dm}{dt} &= (1-\beta)\bigl[\nabla L(w - \eta\beta m) - m\bigr]
\end{align*}

\textbf{Scalar RMSProp}
\begin{align*}
\frac{dw}{dt} &= -\eta \, P^{-1}(\nu)\nabla L(w) \\
\frac{d\nu}{dt} &= (1-\beta_2)\bigl[\|\nabla L(w)\|^2 - \nu\bigr]
\end{align*}

\textbf{RMSProp}
\begin{align*}
\frac{dw}{dt} &= -\eta \, P^{-1}(\nu)\nabla L(w) \\
\frac{d\nu}{dt} &= (1-\beta_2)\bigl[\nabla L(w)^{\odot 2} - \nu\bigr]
\end{align*}

 \textbf{Scalar Adam}
  \begin{align*}
  \frac{dw}{dt} &= -\eta \, P^{-1}(\hat{\nu})\bigl[\beta_1 \hat{m} + (1-\beta_1)\nabla
  \widehat{L}(w)\bigr] \\
  \frac{dm}{dt} &= (1-\beta_1)\bigl[\nabla L(w) - m\bigr] \\
  \frac{d\nu}{dt} &= (1-\beta_2)\bigl[\|\nabla L(w)\|^2 - \nu\bigr]
  \end{align*}

\textbf{Scalar NAdam}
\begin{align*}
\frac{dw}{dt} &= -\eta \, P^{-1}(\hat{\nu})\bigl(\beta_1^2 \hat{m} + (1-\beta_1^2)\nabla L(w)\bigr) \\
\frac{dm}{dt} &= (1-\beta_1)\bigl[\nabla L(w) - m\bigr] \\
\frac{d\nu}{dt} &= (1-\beta_2)\bigl[\|\nabla L(w)\|^2 - \nu\bigr]
\end{align*}

\textbf{Adam}
\begin{align*}
\frac{dw}{dt} &= -\eta \, P^{-1}(\hat{\nu})\bigl[\beta_1 \hat{m} + (1-\beta_1)\nabla
  \widehat{L}(w)\bigr] \\
\frac{dm}{dt} &= (1-\beta_1)\bigl[\nabla L(w) - m\bigr] \\
\frac{d\nu}{dt} &= (1-\beta_2)\bigl[\nabla L(w)^{\odot 2} - \nu\bigr]
\end{align*}

\textbf{NAdam}
\begin{align*}
\frac{dw}{dt} &= -\eta \, P^{-1}(\hat{\nu})\bigl(\beta_1^2 \hat{m} + (1-\beta_1^2)\nabla L(w)\bigr) \\
\frac{dm}{dt} &= (1-\beta_1)\bigl[\nabla L(w) - m\bigr] \\
\frac{d\nu}{dt} &= (1-\beta_2)\bigl[\nabla L(w)^{\odot 2} - \nu\bigr]
\end{align*}

\subsection{Experimental Procedure}
\label{app:procedure}

Each experiment begins with a warm-start phase in which the discrete optimizer is run on its own. The warm-start length is chosen based on prior testing to be long enough for the iterates to settle into the steady-state edge of stability. At its conclusion, both the stable flow and the rod flow are initialized from the current value of the discrete iterates.

\paragraph{Setup.} For each architecture--optimizer pair, we train on a fixed subset of $5{,}000$ CIFAR-10 examples with MSE loss against one-hot targets. Gradients are computed full-batch over all $5{,}000$ examples. For the ViT, where activation memory dominates, we microbatch the per-example gradient. Inputs are channel-normalized using the standard CIFAR-10 statistics: $\mu = (0.4914, 0.4822, 0.4465)$, $\sigma = (0.2470, 0.2435, 0.2616)$. We use no data augmentation, weight decay, or dropout.

\paragraph{Flow initialization.} At iteration $t = \texttt{warmup\_iterations} - 1$, the discrete state is used to seed both flows. The flow centers are set to the average of the last two iterates:
\begin{align}
\bar{w}_0^{\text{flow}} &= \tfrac{1}{2}(w_t + w_{t+1}), \\
\bar{m}_0^{\text{flow}} &= \tfrac{1}{2}(m_t + m_{t+1}) \quad \text{(momentum-based optimizers)}, \\
\bar{\nu}_0^{\text{flow}} &= \tfrac{1}{2}(\nu_t + \nu_{t+1}) \quad \text{(adaptive optimizers)}.
\end{align}
For rod flow, the extent tensor $\Sigma$ is initialized using the phase-space half-difference:
\begin{align}
\delta_0 &= \tfrac{1}{2}(w_{t+1} - w_t), \\
\gamma_0 &= \tfrac{1}{2}(m_{t+1} - m_t), \\
\Sigma_0 &\propto (\delta_0, \gamma_0)\otimes(\delta_0, \gamma_0).
\end{align}

\paragraph{Lockstep loop.} For $t \geq \texttt{warmup\_iterations}$, each iteration of the outer loop executes the following three updates in sequence:
\begin{enumerate}
    \item \textbf{Discrete optimizer.} Take one step at $w_t$ to produce $w_{t+1}, m_{t+1}, \nu_{t+1}$.
    \item \textbf{Stable flow.} Advance $(\bar{w}_{\text{sf}}, \bar{m}_{\text{sf}}, \bar{\nu}_{\text{sf}})$ by $n_{\text{sf}}$ forward-Euler substeps of size $\Delta t_{\text{sf}}$.
    \item \textbf{Rod flow.} Advance $(\bar{w}_{\text{rod}}, \bar{m}_{\text{rod}}, \bar{\nu}_{\text{rod}}, \Sigma_{\text{rod}})$ by $n_{\text{rod}}$ forward-Euler substeps of size $\Delta t_{\text{rod}}$.
\end{enumerate}
Total simulated time per outer-loop iteration is $n_{\text{sf}}\cdot\Delta t_{\text{sf}} = n_{\text{rod}}\cdot\Delta t_{\text{rod}} = 1$, so one continuous-time unit corresponds to one discrete optimizer step. Across all architectures we use $n = 10$ substeps with $\Delta t = 0.1$.

\begin{table}[h]
\centering
\caption{Hyperparameters used in the MLP-on-CIFAR-10 runs.}
\label{tab:hyperparams_mlp}
\begin{tabular}{lcccccc}
\toprule
\textbf{Optimizer} & $\eta$ & $\beta_1$ & $\beta_2$ & $\epsilon$ & \textbf{n\_iter} & \textbf{warmup} \\
\midrule
Heavy Ball       & $0.05$    & $0.4$  & ---     & ---       & 16{,}000 & 14{,}000 \\
Nesterov         & $0.08$    & $0.8$  & ---     & ---       & 12{,}000 & 10{,}000 \\
Scalar RMSProp   & $10^{-4}$ & ---    & $0.99$  & $0$       & 4{,}000  & 2{,}000  \\
RMSProp          & $10^{-4}$ & ---    & $0.99$  & $0$       & 4{,}000  & 2{,}000  \\
Adam             & $10^{-4}$ & $0.8$  & $0.999$ & $10^{-7}$ & 4{,}000  & 2{,}000  \\
Scalar Adam      & $10^{-3}$ & $0.8$  & $0.99$  & $10^{-8}$ & 4{,}000  & 2{,}000  \\
NAdam            & $10^{-4}$ & $0.6$  & $0.999$ & $10^{-7}$ & 4{,}000  & 2{,}000  \\
Scalar NAdam     & $5\!\times\!10^{-4}$ & $0.5$ & $0.99$ & $10^{-8}$ & 3{,}500 & 1{,}500 \\
\bottomrule
\end{tabular}
\end{table}

\paragraph{Eigenvalue computation.} Both the raw and effective Hessian eigenvalues are computed using a warm-started Lanczos solver. Between calls, the solver caches its previous eigenbasis and reuses it as the starting iterate for the next call, exploiting the slow drift of the top eigenspace over the course of training. Eigenvalues are sampled every $200$ discrete steps---a cadence that prevents the eigensolver from dominating wall-clock time while still resolving the dynamics around the EoS threshold.

\subsection{Quantities Tracked}
\label{app:tracked}

We track two main classes of quantities: intrinsic observables (computed independently for the discrete iterates, the stable flow, and the rod flow) and cross-trajectory comparison metrics.

\paragraph{Intrinsic observables.}
\begin{itemize}
\item \textbf{Losses} (\texttt{disc\_loss\_w}, \texttt{disc\_loss\_wbar}, \texttt{sf\_loss}, \texttt{rod\_loss\_wbar}): Evaluated at the raw discrete iterate $w_t$ and the respective trajectory centers $\bar{w}$. 

\item \textbf{Oscillation amplitudes} (\texttt{disc\_delta\_norms}, \texttt{disc\_mu\_norms}, \texttt{delta\_norm\_rod}, \texttt{mu\_norm\_rod}): The norms of the position and momentum half-differences $\delta, \gamma$. For the discrete trajectory, these are explicit finite differences. For rod flow, they are extracted from the principal eigencomponents of $\Sigma$.

\item \textbf{Sharpness} (\texttt{disc\_pre\_sharpness\_wbar}, \texttt{sf\_pre\_sharpness}, \texttt{rod\_pre\_sharpness\_wbar}): The top eigenvalue of the preconditioned Hessian at the respective centers.
\item \textbf{Preconditioner state} (\texttt{nu\_norm\_disc}, \texttt{nu\_norm\_sf}, \texttt{nu\_norm\_rod}, \texttt{disc\_mean\_ess}): For adaptive optimizers, we track the norm of the second-moment vector $\nu$ across all trajectories.
\end{itemize}

\paragraph{Cross-trajectory comparison.}
\begin{itemize}
\item \textbf{Center distances} (\texttt{dist\_wbar\_disc\_to\_sf}, \texttt{dist\_wbar\_disc\_to\_rod}): Euclidean distances between the discrete center $\bar{w}_t$ and each continuous flow center.
\item \textbf{Directional alignment} (\texttt{cos\_delta\_alignment}): Cosine similarity between $\delta_{\text{disc}}$ and $\delta_{\text{rod}}$.
\end{itemize}

All scalar quantities are recorded every step. Eigenvalue-based metrics are recorded every $200$ steps.

\subsection{Architectures}
\label{app:architectures}

\paragraph{MLP.} A three-layer fully-connected network with hidden width $200$ and $\tanh$ activations. CIFAR-10 inputs are flattened to $\mathbb{R}^{3072}$, passed through two hidden layers of width $200$, and projected to a $10$-dimensional output. All linear layers include biases, giving a total of $656{,}810$ parameters.

\paragraph{CNN.} A small two-block convolutional network. Each block applies a $3\times 3$ convolution with padding $1$ (with bias), followed by $\tanh$ and $2\times 2$ average pooling; the first block has $3 \to 32$ channels and the second $32 \to 32$. After two pooling stages, the $32 \times 32$ inputs are reduced to $8 \times 8$ feature maps, which are flattened and passed through a linear readout to dimension $10$. Total parameter count: $30{,}634$.

\paragraph{ViT.} A small Vision Transformer based on the \texttt{SimpleViT} variant from \texttt{vit\_pytorch}. We use patch size $4$ (so each $32\times 32$ image yields $64$ patches), embedding dimension $64$, $3$ transformer blocks, $8$ attention heads, and per-block MLP hidden size $256$. Position information is supplied by a fixed 2-D sin-cos positional encoding rather than learned embeddings, following the SimpleViT design. The linear classification head is initialized at half its default magnitude (\texttt{init\_scale}=$0.5$) to reduce the initial sharpness, which would otherwise place the iterates immediately in a divergent regime. To ensure that the second and third derivatives required by the Lanczos eigensolvers are well-defined along the trajectory, we replace every \texttt{LayerNorm} module with a \texttt{TrainableLayerNorm} (\texttt{src/architectures/utils.py}) that matches the standard forward behavior and learnable scale/shift parameters but computes its mean and variance with smooth operations rather than cached running statistics. Total parameter count is approximately $165{,}000$, dominated by the three transformer blocks.

\section{Experimental Results}\label{app:results}

Across all eight optimizers, the rod flow tracks the discrete-time optimizers significantly more accurately than the stable flow does.

Because of the finite step size, the discrete-time optimizers cannot follow the direction of steepest descent \textit{exactly}---so the loss decreases more slowly for the discrete-time optimizer than for the stable flow. Rod flow matches this slower loss trajectory.

Across the different optimizers, rod flow hovers at the correct preconditioned sharpness threshold. Stable flow, by contrast, is meant to proxy the step size going to zero. It has no preconditioned sharpness limit and continues to sharpen well past the stability threshold.

While there is some discrepancy between the center of rod flow and those of the discrete iterates, it is much smaller than the corresponding discrepancy for stable flow. And a consistent feature is that the rod flow discrepancy levels off: after an initial phase of minor divergence, the discrepancy quickly plateaus. Rod flow also accurately tracks the magnitudes of the position half-displacement and the momentum half-displacement of the discrete iterates.

For preconditioned methods, there is strong agreement between the second moment $\nu$ of rod flow and the second moment of the discrete-time optimizer. An advantage of rod flow is that it not only tracks the center of the discrete iterates, but also models where the iterates actually visit---the endpoints of the rod. Because it has access to gradient evaluations at these endpoints, $\nu$ is tracked accurately.

One interesting observation is the behavior of the $\delta$ alignment. A consistent feature across optimizers is that, rather than hovering near 1, the $\delta$ alignment fluctuates over the course of training. This could be due to the existence of multiple sharp directions in the Hessian.

(Note that in the experimental figures below, the momentum oscillation is mistakenly denoted as $\mu$ instead of $\gamma$.)

\clearpage

% Heavy Ball
\begin{figure}
  \centering
  \textbf{\large Heavy Ball}\par\vspace{0.2cm}
  \includegraphics[width=0.33\linewidth]{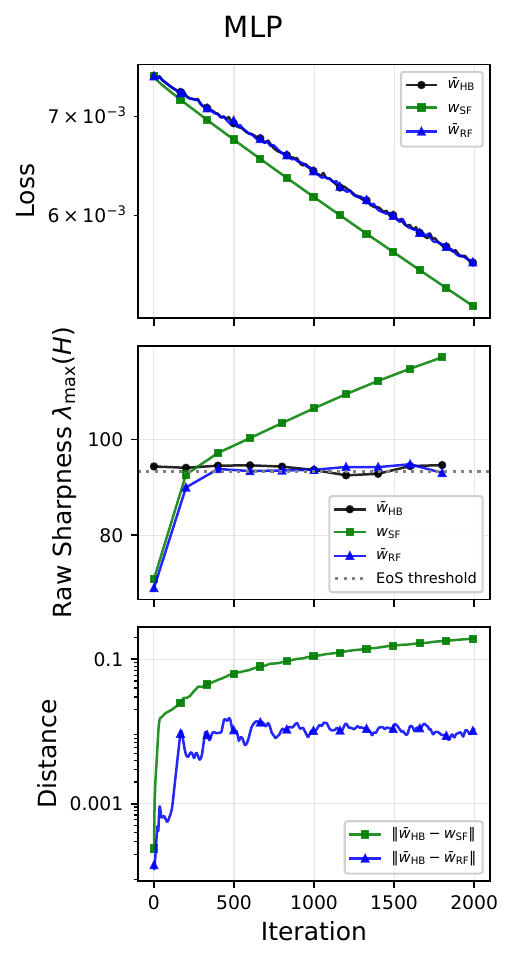}%
  \hfill
  \includegraphics[width=0.33\linewidth]{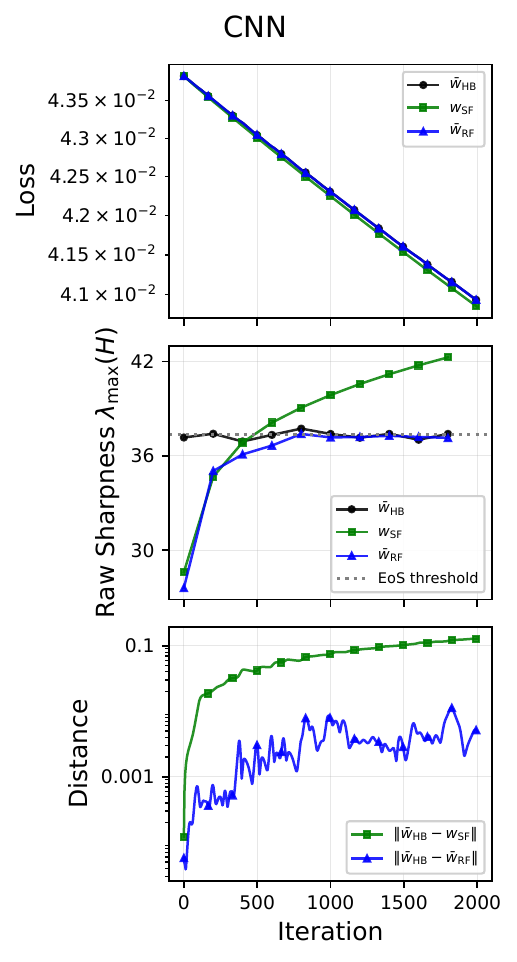}%
  \hfill
  \includegraphics[width=0.33\linewidth]{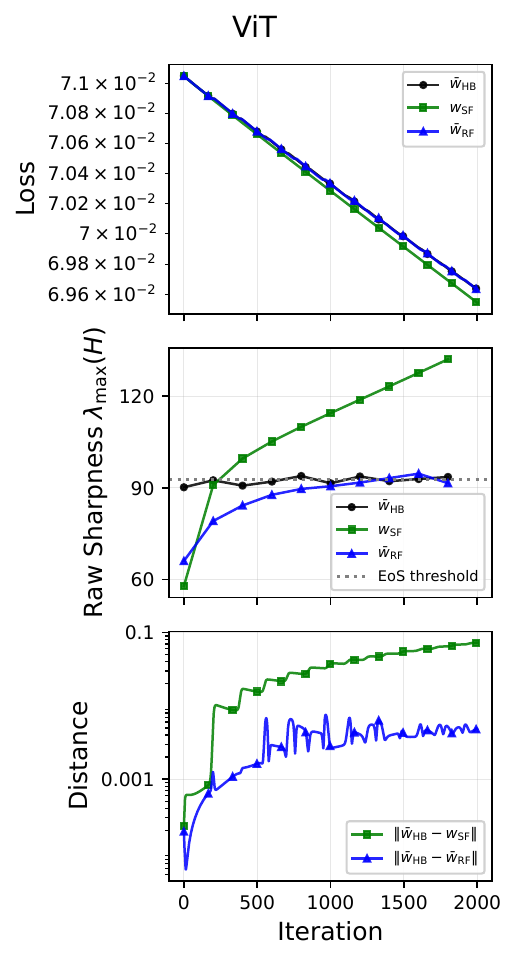}%
  \caption{\textbf{Experimental Results for Heavy Ball Momentum.} \textbf{MLP:} $\eta = 0.05$, $\beta = 0.4$. \textbf{CNN:} $\eta = 0.125$, $\beta = 0.4$. \textbf{ViT:} $\eta = 0.04$, $\beta = 0.3$.}
  \label{fig:centers_hb}
\end{figure}

% Nesterov
\begin{figure}
  \centering
  \textbf{\large Nesterov}\par\vspace{0.2cm}
  \includegraphics[width=0.33\linewidth]{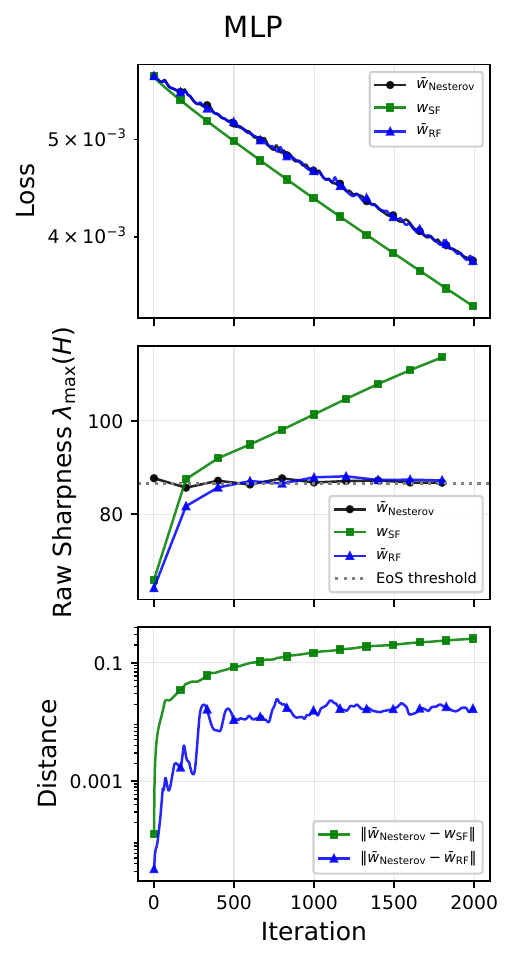}%
  \hfill
  \includegraphics[width=0.33\linewidth]{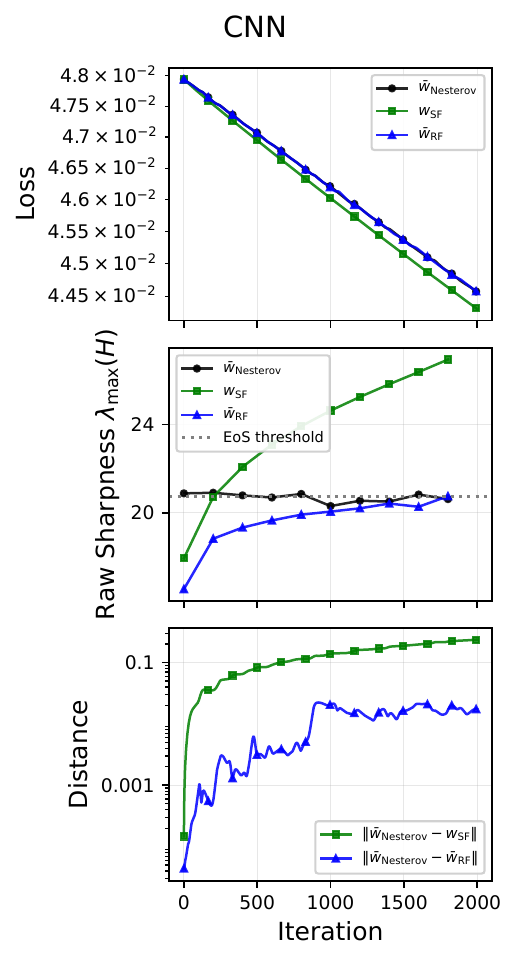}%
  \hfill
  \includegraphics[width=0.33\linewidth]{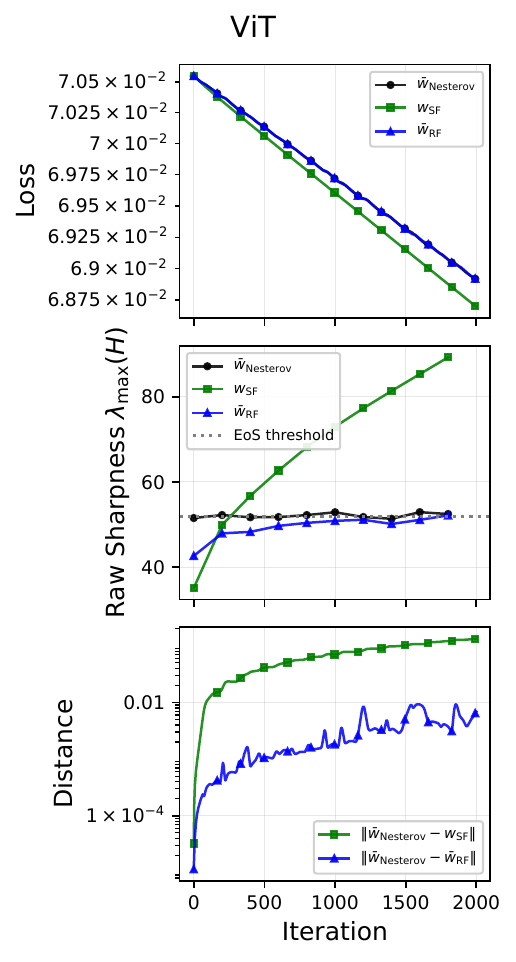}%
  \caption{\textbf{Experimental Results for Nesterov Momentum.} \textbf{MLP:} $\eta = 0.08$, $\beta = 0.8$. \textbf{CNN:} $\eta = 0.125$, $\beta = 0.4$. \textbf{ViT:} $\eta = 0.02$, $\beta = 0.4$.}
  \label{fig:centers_nesterov}
\end{figure}

% SRMSProp
\begin{figure}
  \centering
  \textbf{\large SRMSProp}\par\vspace{0.2cm}
  \includegraphics[width=0.33\linewidth]{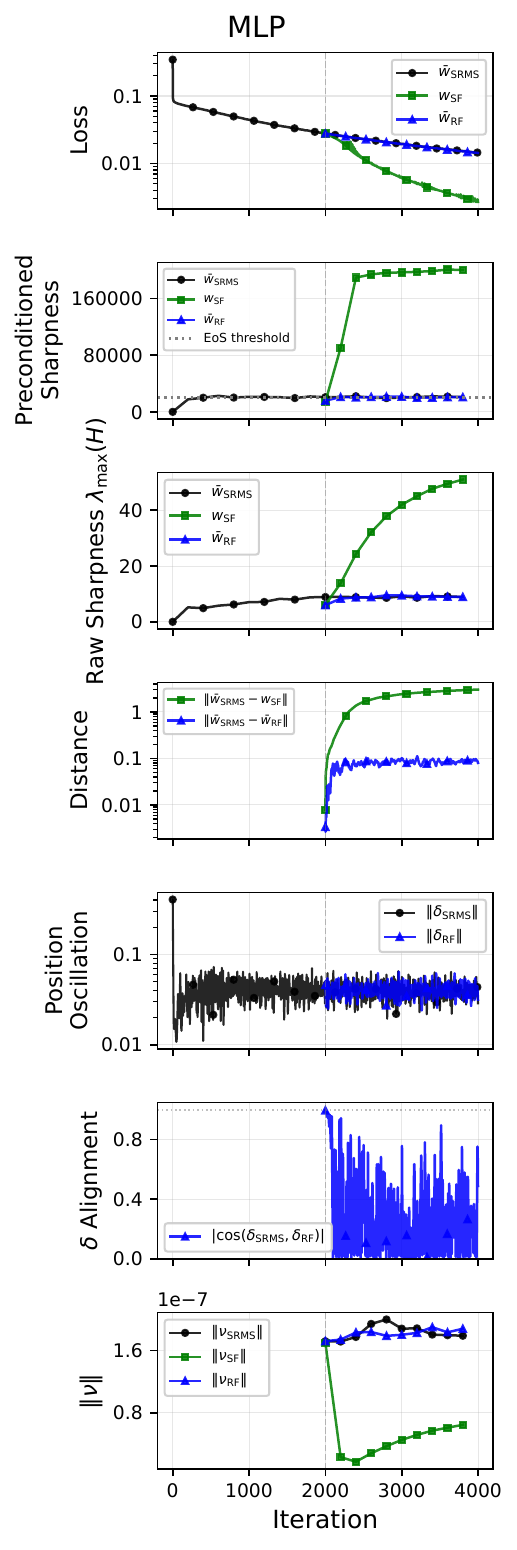}%
  \hfill
  \includegraphics[width=0.33\linewidth]{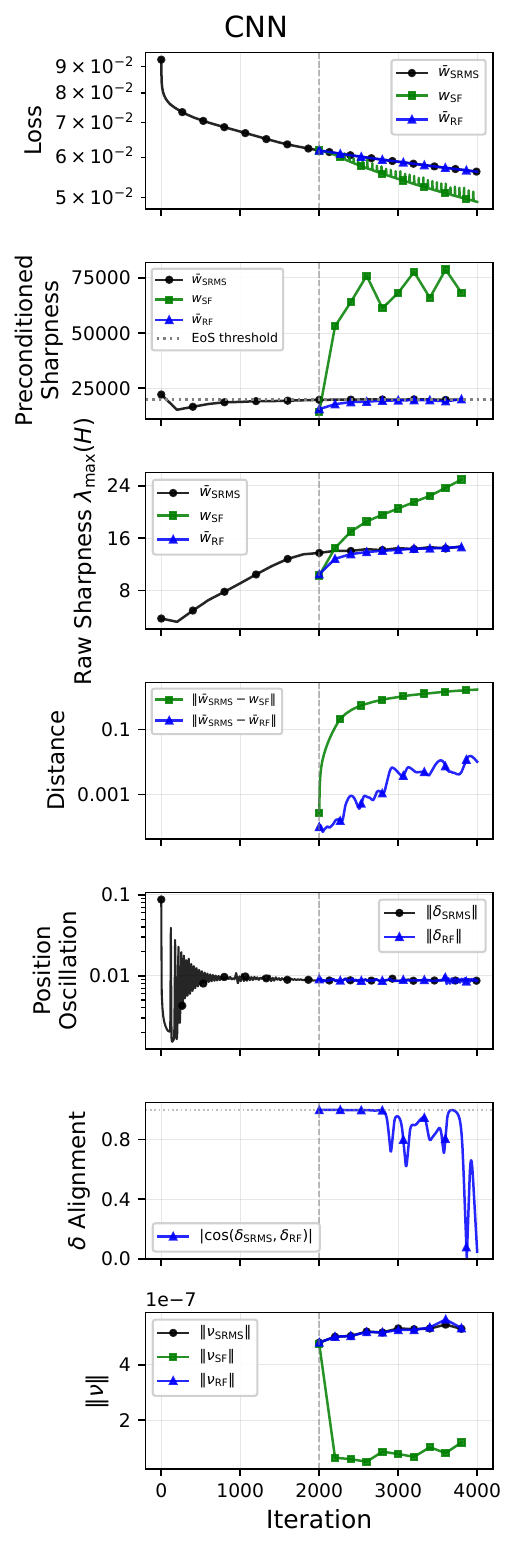}%
  \hfill
  \includegraphics[width=0.33\linewidth]{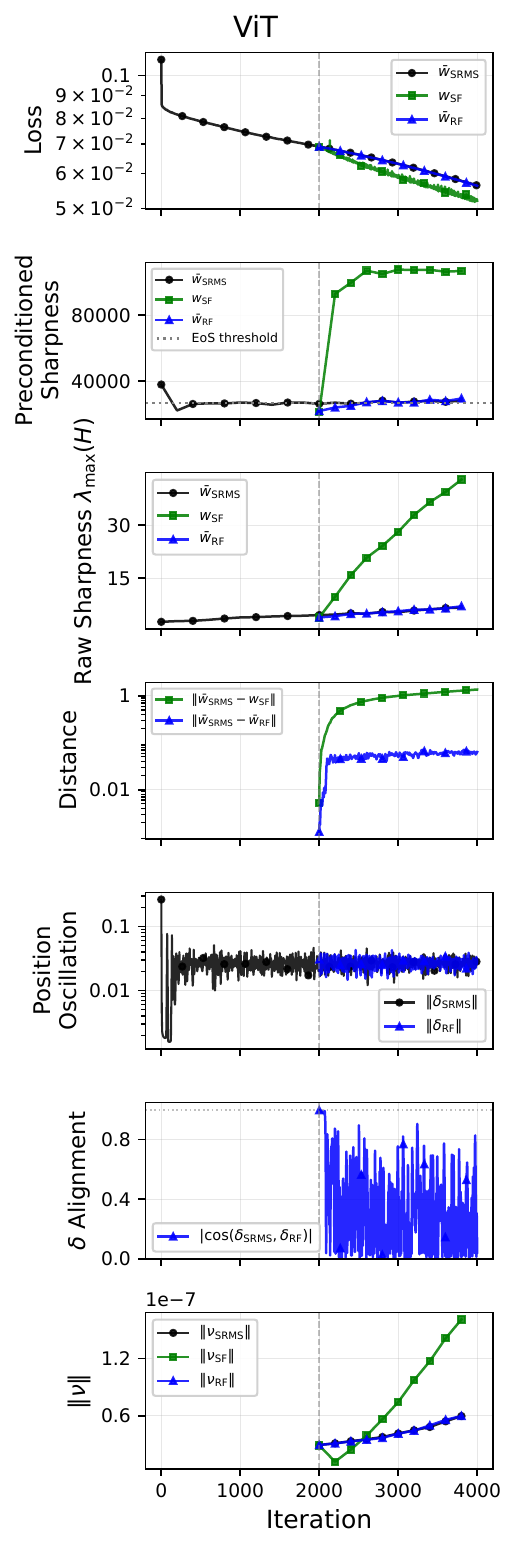}
    \caption{\textbf{Experimental Results for Scalar RMSProp} (no bias correction). \textbf{MLP:} $\eta = 10^{-4}$, $\beta_2 = 0.99$. \textbf{CNN:} $\eta = 10^{-4}$, $\beta_2 = 0.99$. \textbf{ViT:} $\eta = 7.5 \times 10^{-5}$, $\beta_2 = 0.99$.}
  \label{fig:summary_srmsprop}
\end{figure}

% RMSProp
\begin{figure}
  \centering
  \textbf{\large RMSProp}\par\vspace{0.2cm}
  \includegraphics[width=0.33\linewidth]{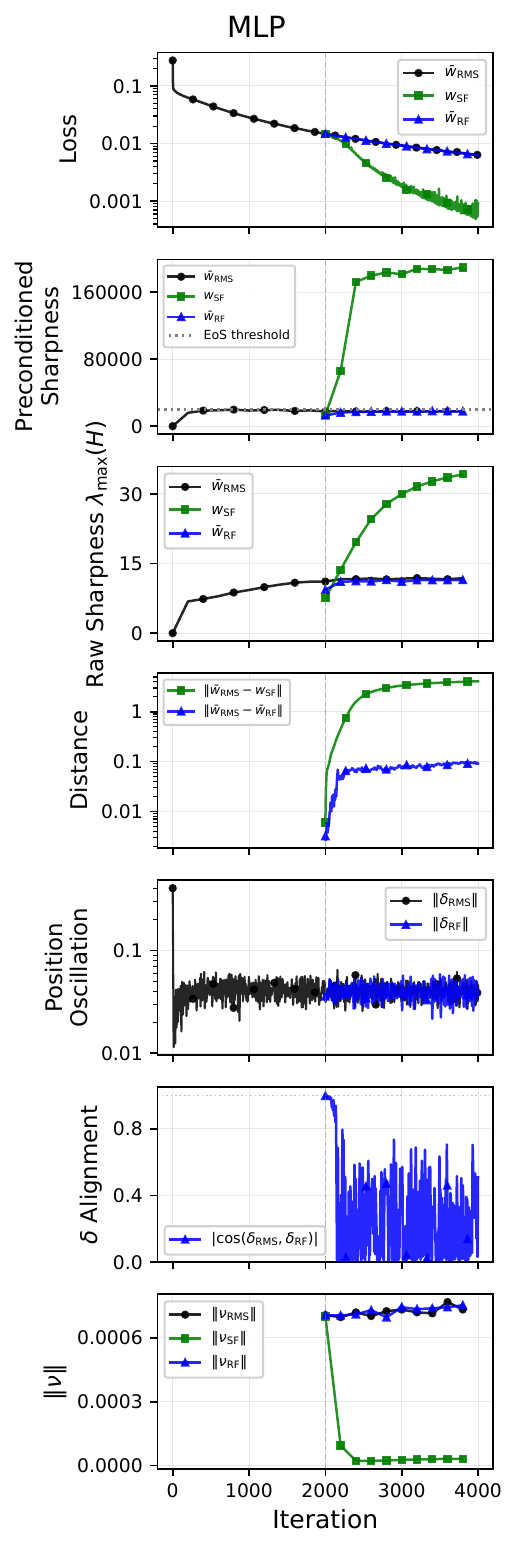}%
  \hfill
  \includegraphics[width=0.33\linewidth]{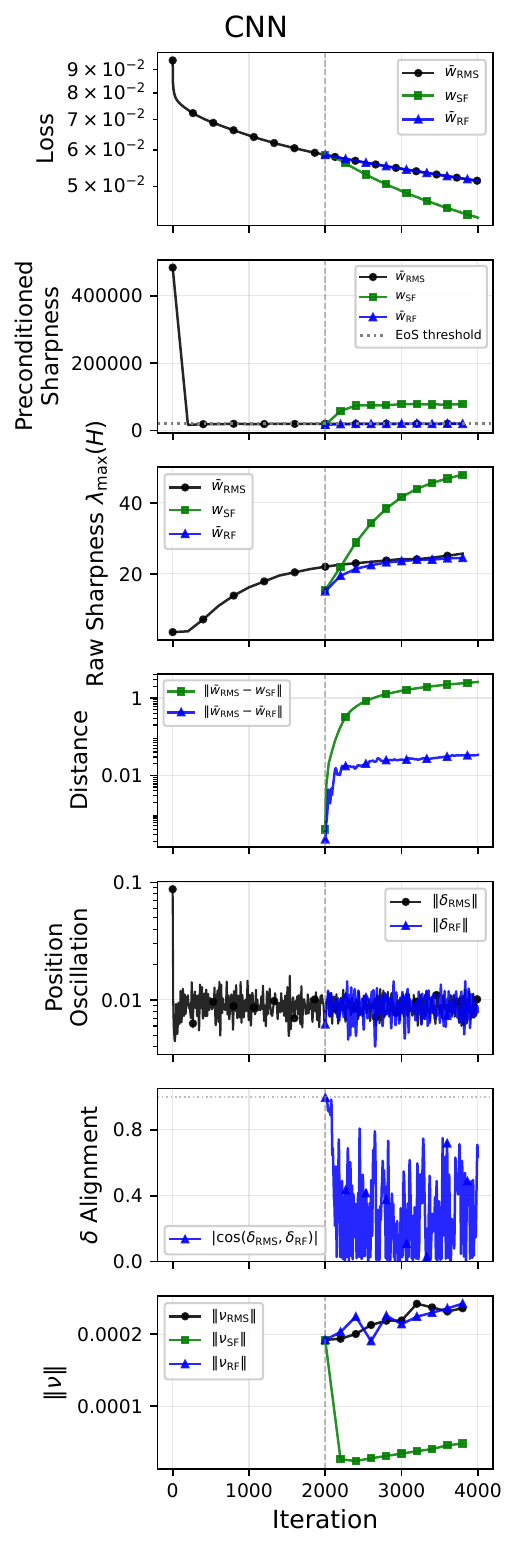}%
  \hfill
  \includegraphics[width=0.33\linewidth]{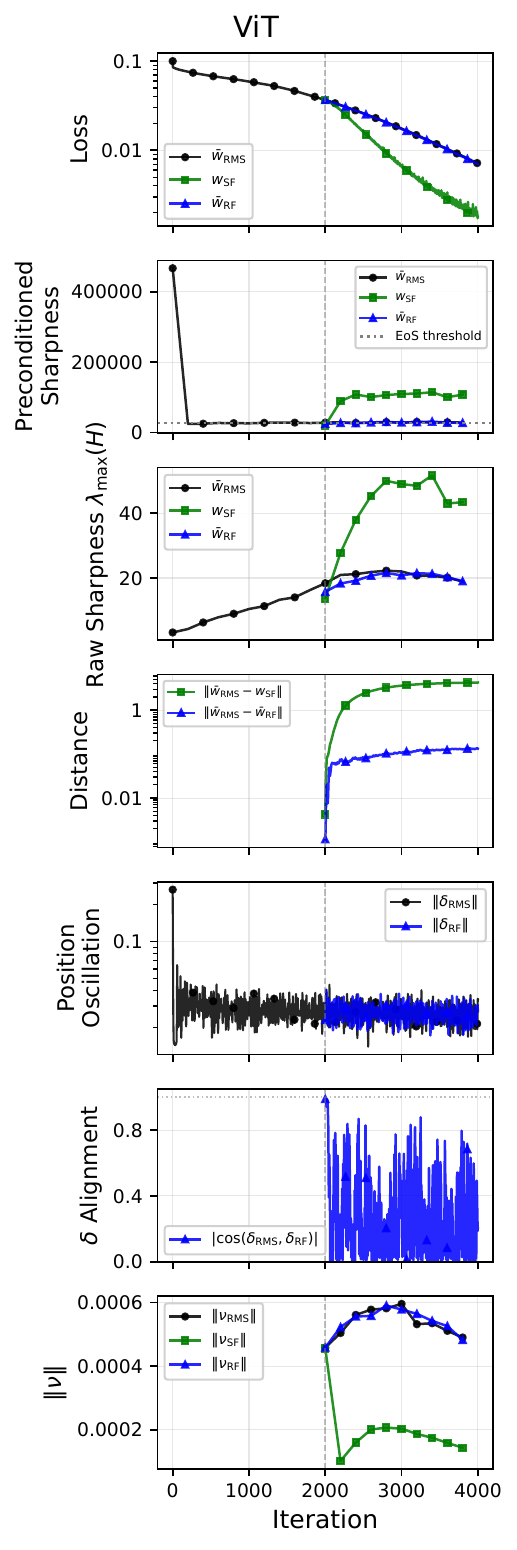}
  \caption{\textbf{Experimental Results for RMSProp} (no bias correction). \textbf{MLP:} $\eta = 10^{-4}$, $\beta_2 = 0.99$. \textbf{CNN:} $\eta = 10^{-4}$, $\beta_2 = 0.99$. \textbf{ViT:} $\eta = 7.5 \times 10^{-5}$, $\beta_2 = 0.99$.}
  \label{fig:summary_rmsprop}
\end{figure}

% SAdam
\begin{figure}
  \centering
  \textbf{\large SAdam}\par\vspace{0.2cm}
  \includegraphics[width=0.33\linewidth]{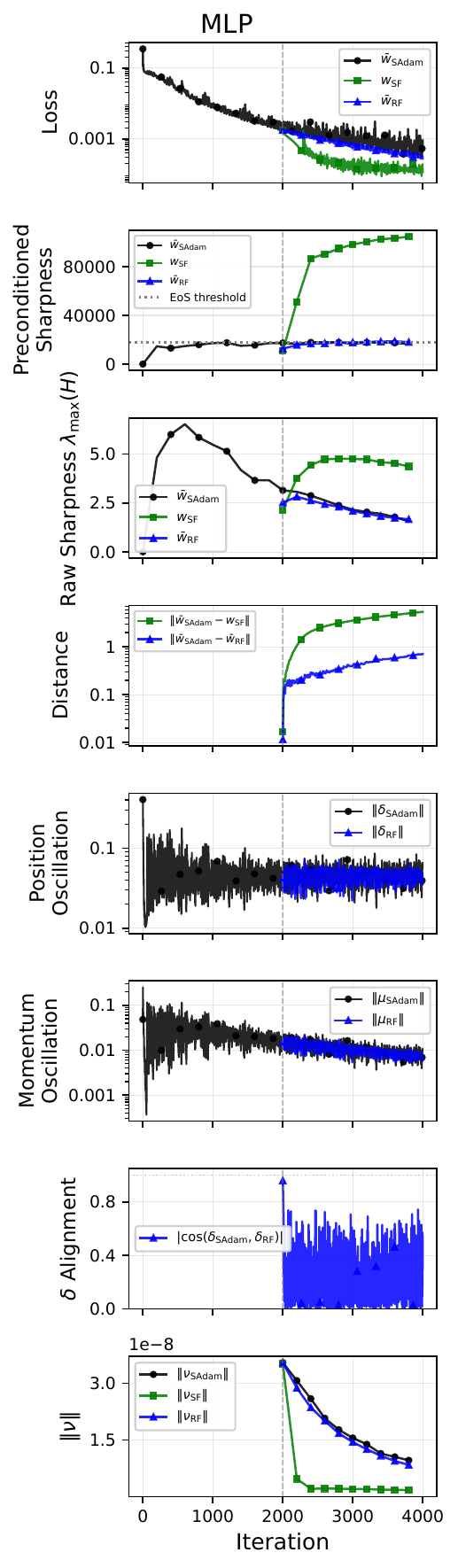}%
  \hfill
  \includegraphics[width=0.33\linewidth]{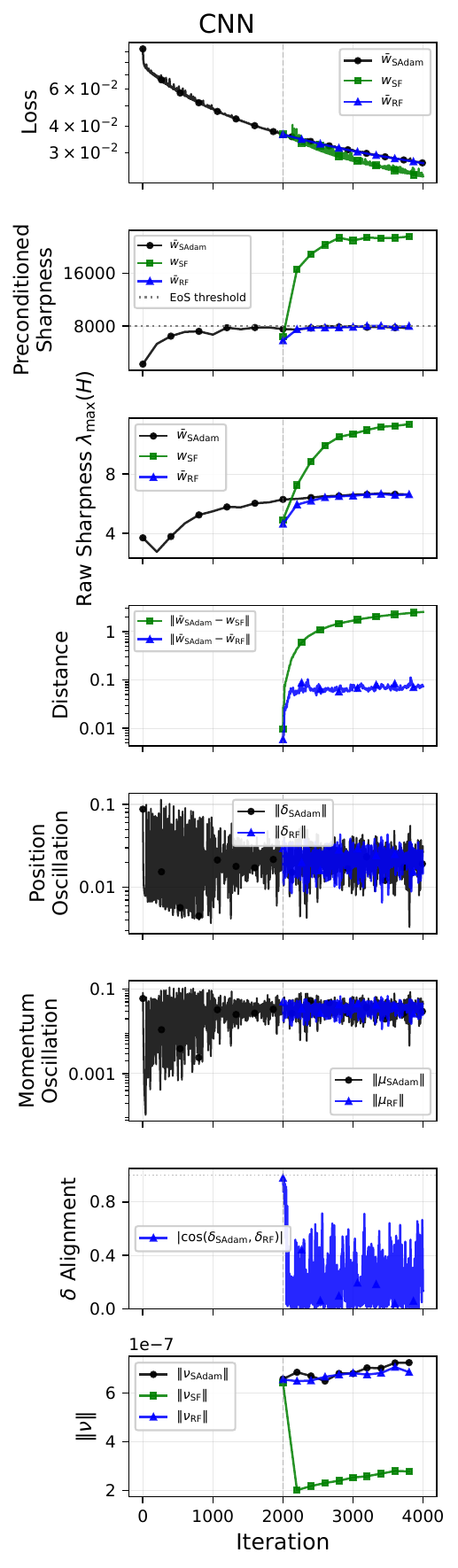}%
  \hfill
  \includegraphics[width=0.33\linewidth]{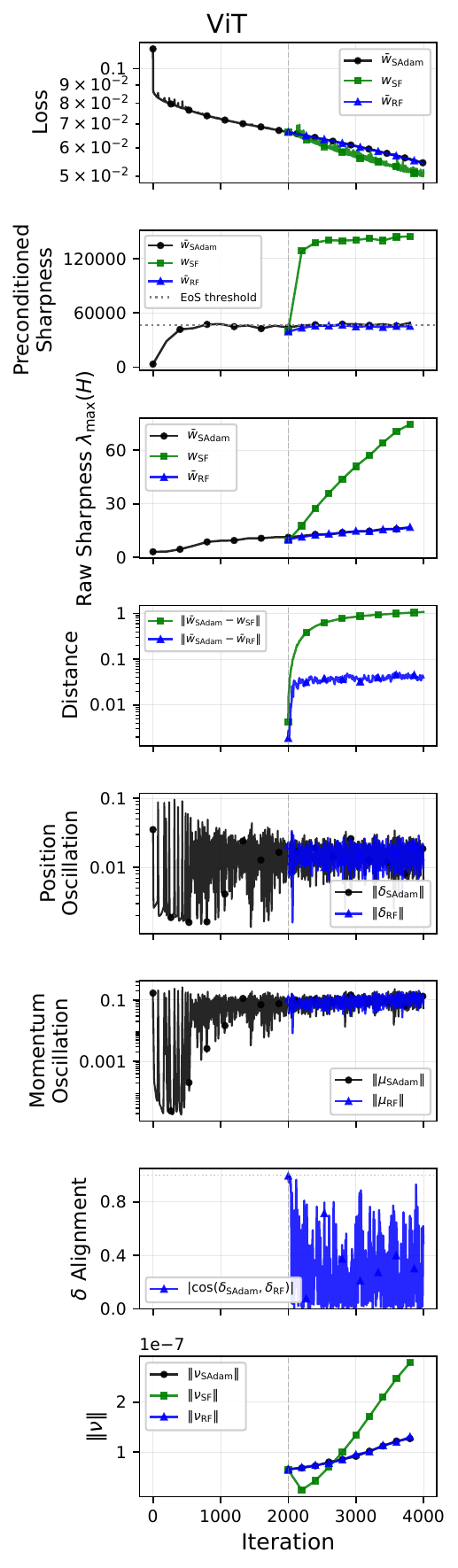}
  \caption{\textbf{Experimental Results for Scalar Adam} (with bias correction). \textbf{MLP:} $\eta = 10^{-3}$, $\beta_1 = 0.8$, $\beta_2 = 0.99$. \textbf{CNN:} $\eta = 10^{-3}$, $\beta_1 = 0.6$, $\beta_2 = 0.99$. \textbf{ViT:} $\eta = 10^{-4}$, $\beta_1 = 0.4$, $\beta_2 = 0.99$.}
  \label{fig:summary_sadam}
\end{figure}

% SNAdam
\begin{figure}
  \centering
  \textbf{\large SNAdam}\par\vspace{0.2cm}
  \includegraphics[width=0.33\linewidth]{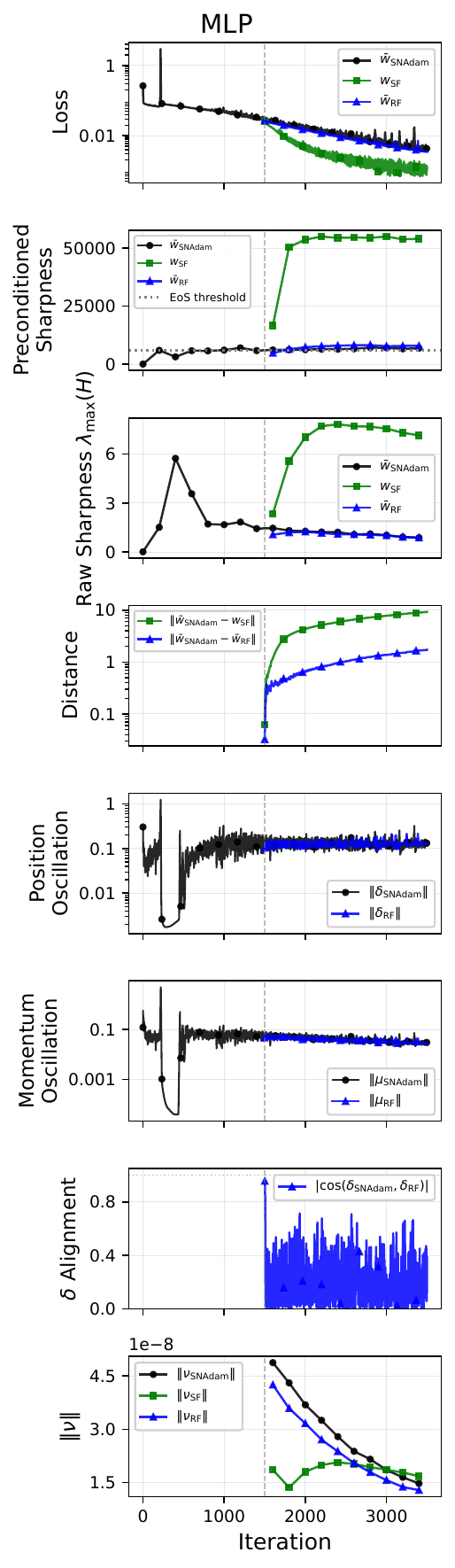}%
  \hfill
  \includegraphics[width=0.33\linewidth]{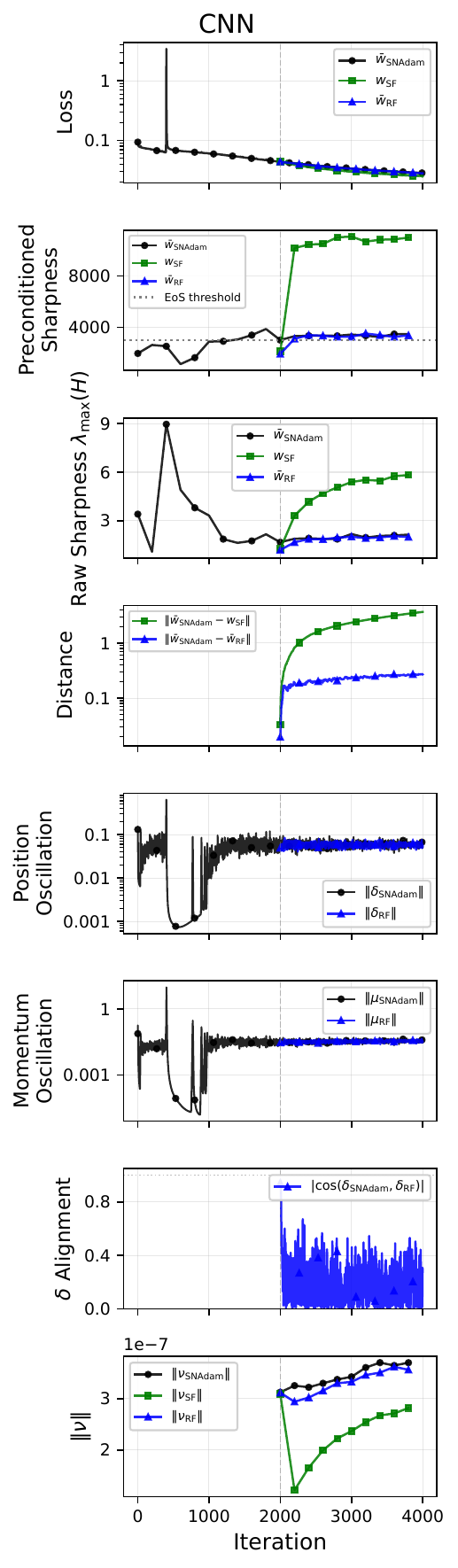}%
  \hfill
  \includegraphics[width=0.33\linewidth]{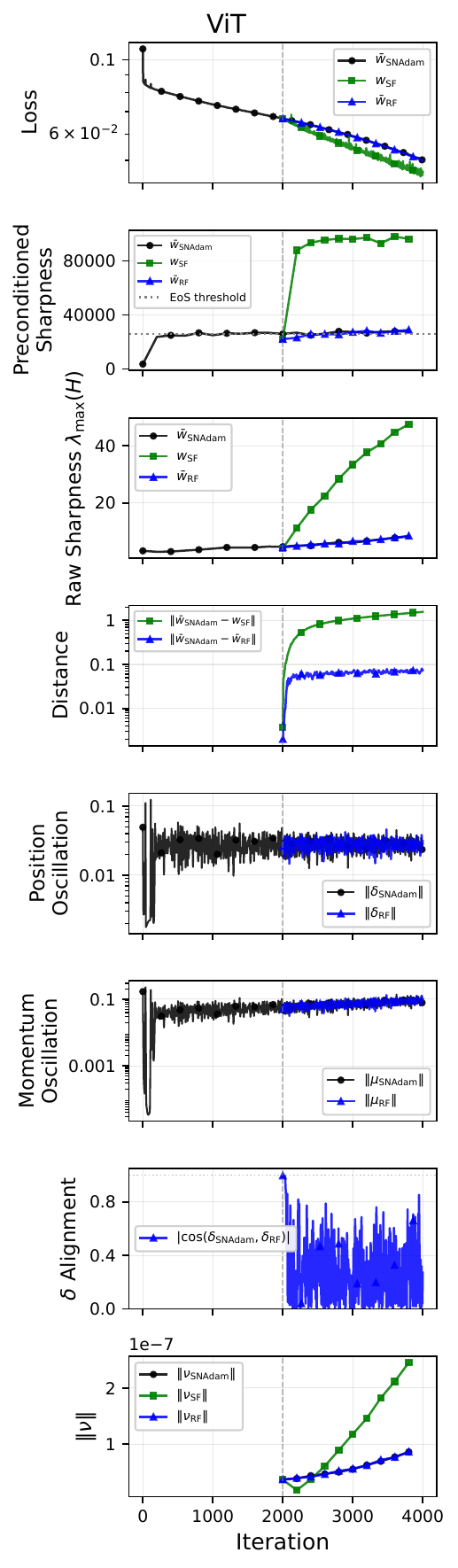}
  \caption{\textbf{Experimental Results for Scalar NAdam} (with bias correction). \textbf{MLP:} $\eta = 5 \times 10^{-4}$, $\beta_1 = 0.5$, $\beta_2 = 0.99$. \textbf{CNN:} $\eta = 10^{-3}$, $\beta_1 = 0.5$, $\beta_2 = 0.99$. \textbf{ViT:} $\eta = 10^{-4}$, $\beta_1 = 0.4$, $\beta_2 = 0.99$.}
  \label{fig:summary_snadam}
\end{figure}

% Adam
\begin{figure}
  \centering
  \textbf{\large Adam}\par\vspace{0.2cm}
  \includegraphics[width=0.33\linewidth]{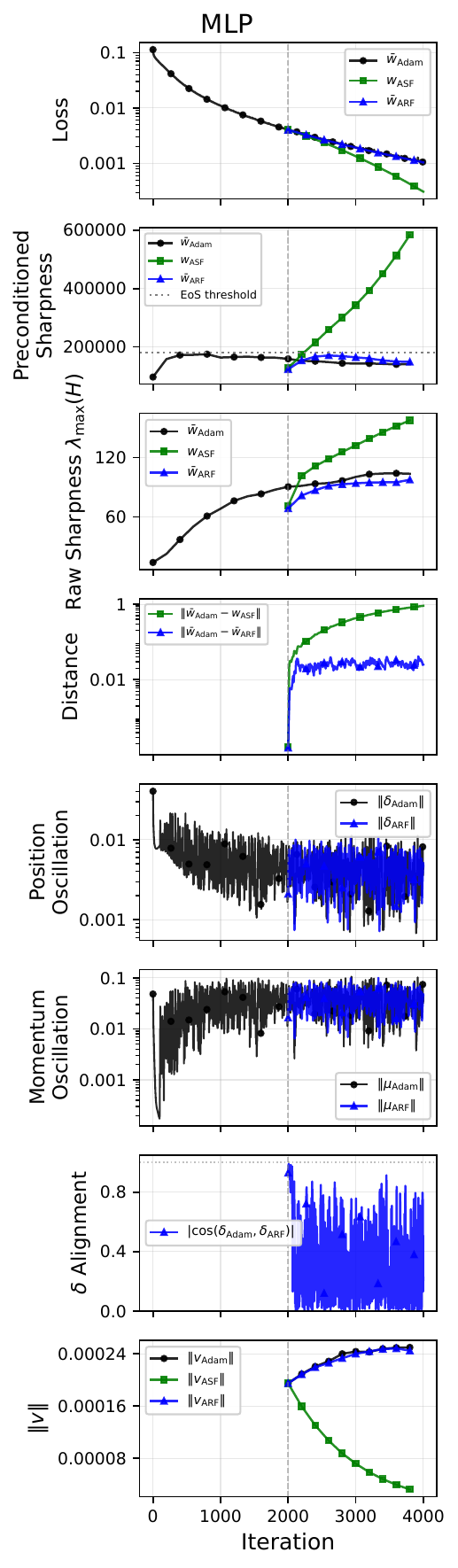}%
  \hfill
  \includegraphics[width=0.33\linewidth]{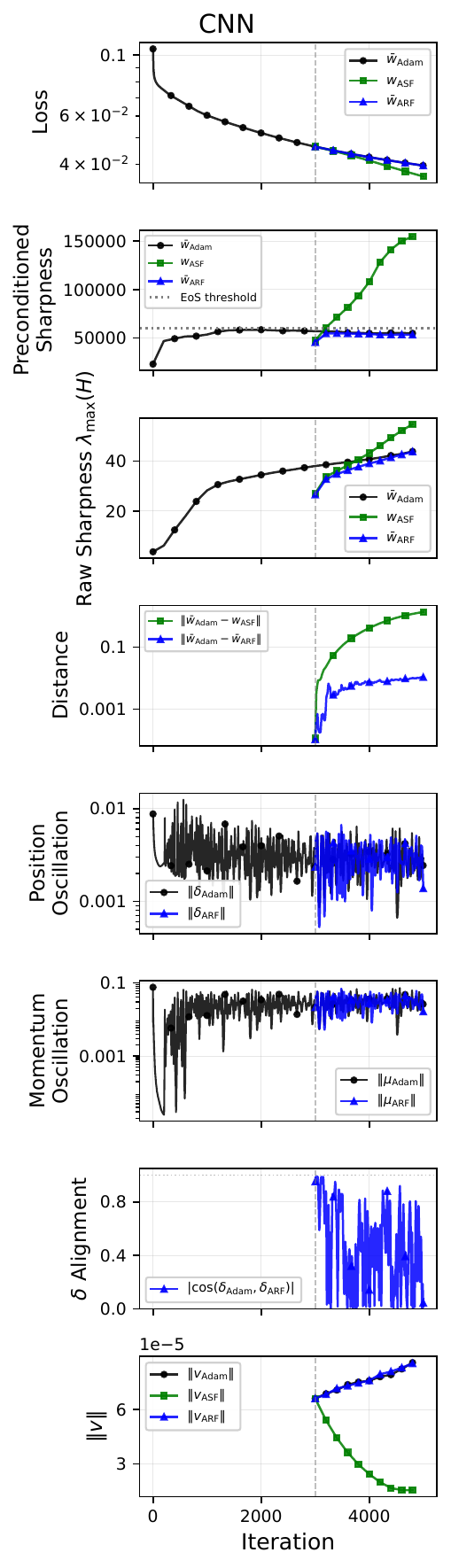}%
  \hfill
  \includegraphics[width=0.33\linewidth]{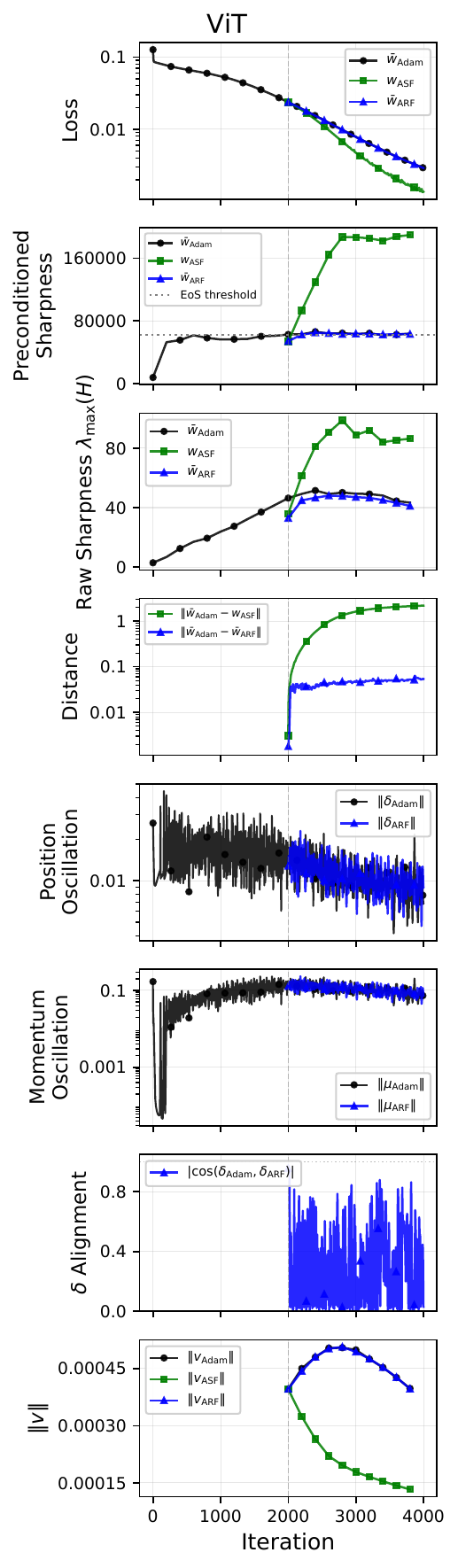}
  \caption{\textbf{Experimental Results for Adam} (with bias correction). \textbf{MLP:} $\eta = 10^{-4}$, $\beta_1 = 0.8$, $\beta_2 = 0.999$. \textbf{CNN:} $\eta = 10^{-4}$, $\beta_1 = 0.5$, $\beta_2 = 0.999$. \textbf{ViT:} $\eta = 7.5 \times 10^{-5}$, $\beta_1 = 0.4$, $\beta_2 = 0.999$.}
  \label{fig:summary_adam}
\end{figure}

% NAdam
\begin{figure}
  \centering
  \textbf{\large NAdam}\par\vspace{0.2cm}
  \includegraphics[width=0.33\linewidth]{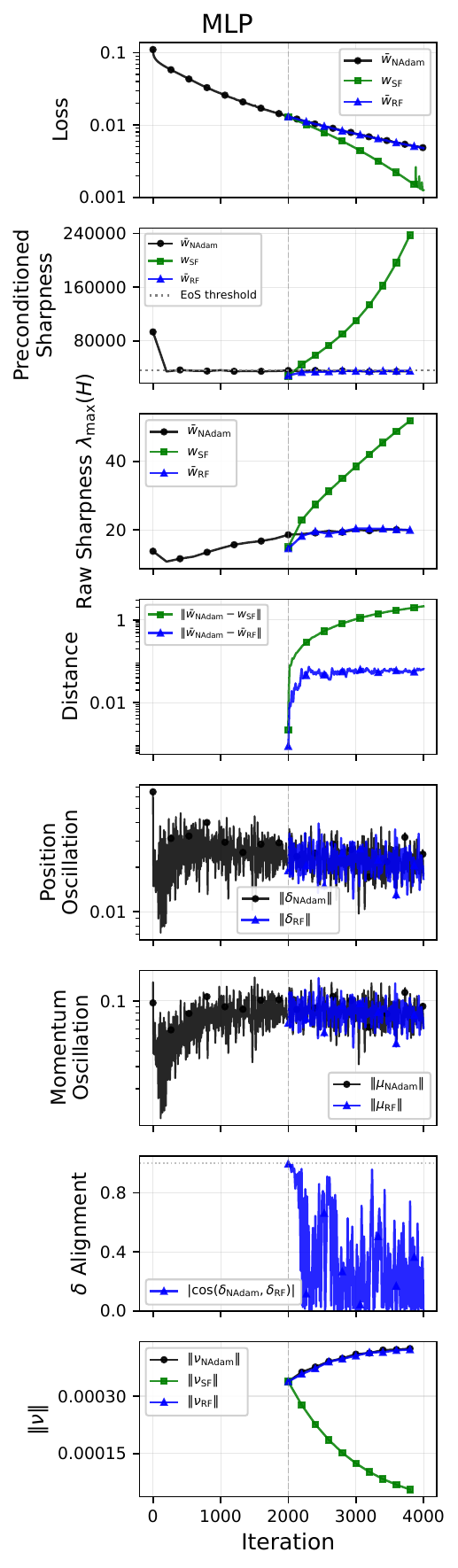}%
  \hfill
  \includegraphics[width=0.33\linewidth]{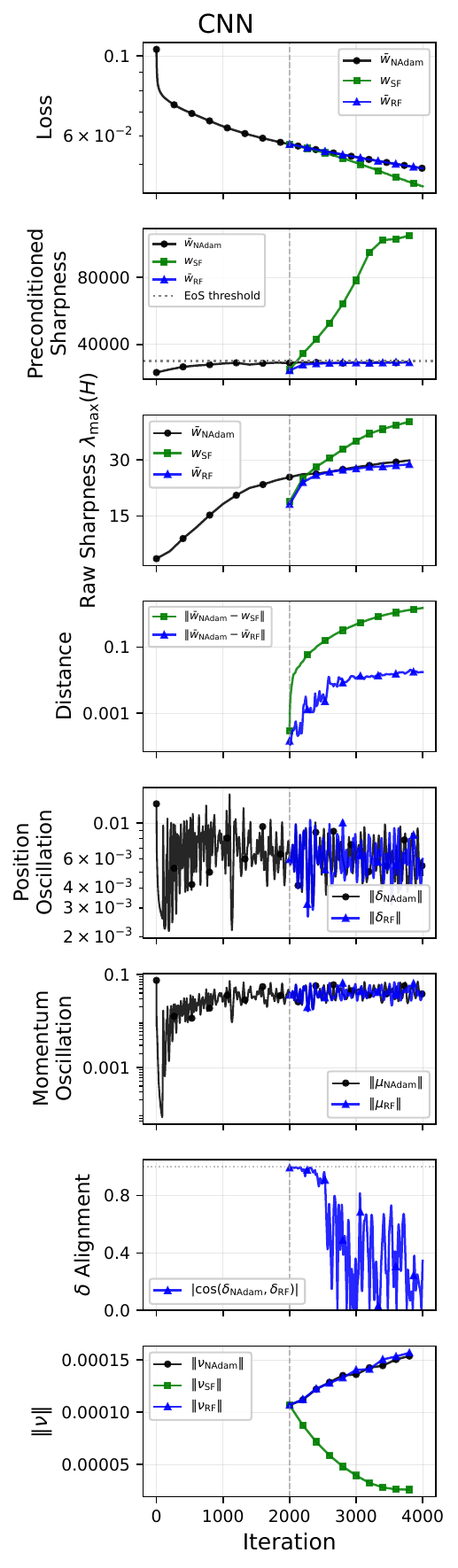}%
  \hfill
  \includegraphics[width=0.33\linewidth]{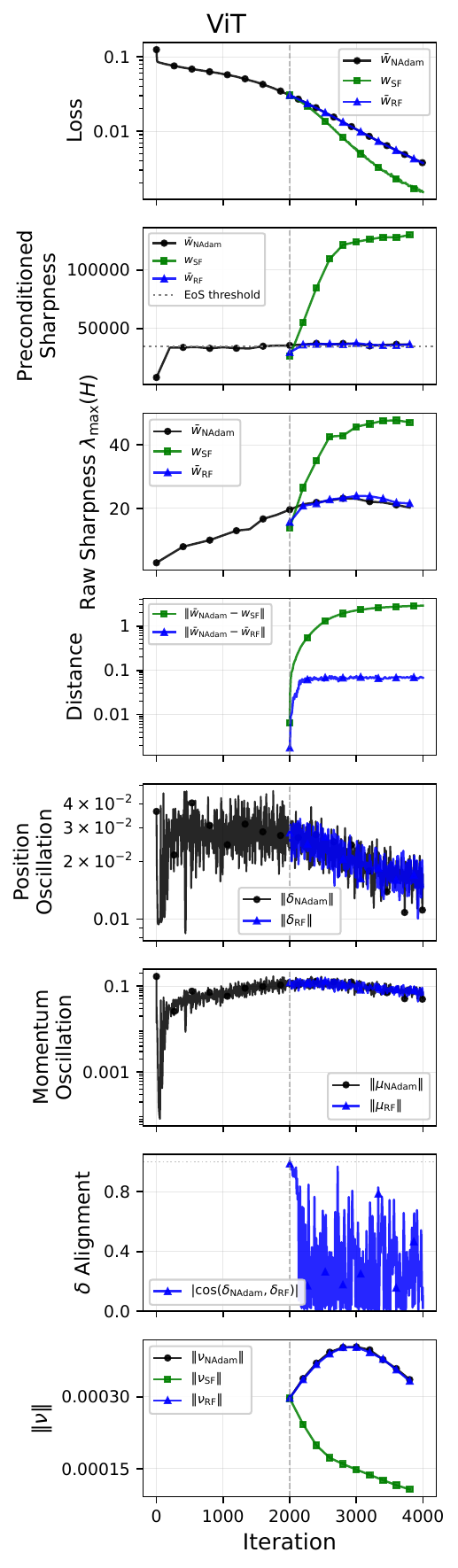}
  \caption{\textbf{Experimental Results for NAdam} (with bias correction). \textbf{MLP:} $\eta = 10^{-4}$, $\beta_1 = 0.6$, $\beta_2 = 0.999$. \textbf{CNN:} $\eta = 10^{-4}$, $\beta_1 = 0.5$, $\beta_2 = 0.999$. \textbf{ViT:} $\eta = 7.5 \times 10^{-5}$, $\beta_1 = 0.4$, $\beta_2 = 0.999$.}
  \label{fig:summary_nadam}
\end{figure}

\end{document}